\documentclass[journal]{IEEEtran}

\ifCLASSINFOpdf
\else
\fi
\usepackage{graphicx}
\usepackage{amsmath}
\usepackage{amsfonts}
\usepackage{multicol}
\usepackage{multirow}
\usepackage{cite}
\usepackage{mathrsfs}
\usepackage{amsmath}
\usepackage{color,xcolor}
\usepackage{color}
\usepackage[pagebackref=true,breaklinks=true,letterpaper=true,colorlinks,bookmarks=false]{hyperref}

\usepackage{times}
\usepackage{epsfig}
\usepackage{graphicx}
\usepackage{amsmath}
\usepackage{amssymb}
\usepackage{graphicx}
\usepackage{cite}
\usepackage{enumerate}
\usepackage{cases}
\usepackage{multirow}
\usepackage{verbatim}
\usepackage{amssymb}
\usepackage{CJK}
\usepackage{algorithm}
\usepackage{algorithmicx}
\usepackage{algpseudocode}

\usepackage{color}
\usepackage{bm}
\usepackage{booktabs}
\usepackage{colortbl}
\usepackage[switch]{lineno}

\newcommand{\etal}{\textit{et al}.}
\newcommand{\ie}{\textit{i}.\textit{e}.}
\newcommand{\eg}{\textit{e}.\textit{g}.}
\newcommand{\etc}{\textit{etc}}

\begin{document}
	%\linenumbers
	
	\title{Underwater Image Enhancement via Medium Transmission-Guided Multi-Color Space Embedding}
	
	\author{Chongyi~Li,
		Saeed~Anwar,~\IEEEmembership{Member,~IEEE,}
		Junhui~Hou,~\IEEEmembership{Senior Member,~IEEE,}\\
		Runmin~Cong,~\IEEEmembership{Member,~IEEE,}
		Chunle~Guo,
		and Wenqi~Ren~\IEEEmembership{Member,~IEEE}\\
		%
%		\thanks{This work was supported in part by the Hong Kong RGC under Grants  CityU 21211518, 11219019, and 11202320; in part by Hong Kong GRF-RGC General Research Fund under Grant 9042958 (CityU 11203820) and 9042816 (CityU 11209819); in part by the Beijing Nova Program under Grant Z201100006820016; in part by the National Natural Science Foundation of China under Grant 62002014; in part by Young Elite Scientist Sponsorship Program by the China Association for Science and Technology under Grant 2020QNRC001. Chunle Guo is sponsored by CAAI-Huawei MindSpore Open Fund.}
		\thanks{Chongyi Li is with the Department of Computer Science, City University of Hong Kong, Kowloon 999077, Hong Kong
			(e-mail: lichongyi25@gmail.com). 
			
			Saeed Anwar is with Data61, CSIRO, and Australian National University (ANU) (e-mail: saeed.anwar@csiro.au).
			
			Junhui Hou is with the Department of Computer Science, City University of Hong Kong, Kowloon 999077, Hong Kong (e-mail: jh.hou@cityu.edu.hk).
			
			Runmin Cong is the Institute of Information Science, Beijing Jiaotong University, Beijing 100044, China, and also with the Beijing Key Laboratory of Advanced Information Science and Network Technology, Beijing Jiaotong University, Beijing 100044, China  (e-mail: rmcong@bjtu.edu.cn).
			
			Chunle Guo is the College of Computer Science, Nankai University, Tianjin, China (e-mail: guochunle@nankai.edu.cn).

			Wenqi Ren is with State Key Laboratory of Information Security, Institute of Information Engineering, Chinese Academy of Sciences, China (e-mail: rwq.renwenqi@gmail.com).
			
			(Corresponding author: Junhui Hou.)
			
This paper has supplementary downloadable material available at http://ieeexplore.ieee.org., provided by the author. The material includes 1) the detailed network structure with hyper-parameters, 2) more
visual results on various benchmarks, and the comparison on a real underwater video. Contact lichongyi25@gmail.com for further questions about this work.

	}}
	\markboth{IEEE TRANSACTIONS ON IMAGE PROCESSING}%
	{Shell \MakeLowercase{\textit{et al.}}: Bare Demo of IEEEtran.cls for Journals}

	\maketitle
	
	\begin{abstract}
		Underwater images suffer from color casts and low contrast due to  wavelength- and distance-dependent attenuation and  scattering. To solve these two degradation issues, we present an underwater image enhancement network via medium transmission-guided multi-color space embedding, called \emph{Ucolor}. Concretely, we first propose a multi-color space encoder network, which enriches the diversity of feature representations by incorporating the characteristics of different color spaces into a unified structure. Coupled with an attention mechanism, the most discriminative features extracted from multiple color spaces are adaptively integrated and highlighted. Inspired by underwater imaging physical models, we design a medium transmission (indicating the percentage of the scene radiance reaching the camera)-guided decoder network to enhance the response of network towards quality-degraded regions. As a result, our network can effectively improve the visual quality of underwater images by exploiting  multiple color spaces embedding and the advantages of both physical model-based and learning-based methods. Extensive experiments demonstrate that our \emph{Ucolor} achieves superior performance against state-of-the-art methods in terms of both visual quality and quantitative metrics. The code is publicly available at: \url{https://li-chongyi.github.io/Proj_Ucolor.html}.
	\end{abstract}
	
	
	\IEEEpeerreviewmaketitle

	\section{Introduction}
	\IEEEPARstart{U}{nderwater} images inevitably suffer from quality degradation issues caused by wavelength- and distance-dependent attenuation and scattering \cite{Akkaynak2017}. Typically, when the light propagates through water, it suffers from selective attenuation that results in various degrees of color deviations. Besides, the light is scattered by suspending particles such as micro phytoplankton and non-algal particulate in water, which causes low contrast. 
	An effective solution to recover underlying clean images is of great significance for improving the visual quality of images captured in water and accurately understanding underwater world.

	The quality degradation degrees of underwater images can be implicitly reflected by the medium transmission that represents the percentage of the scene radiance reaching the camera.
	Hence, physical model-based underwater image enhancement methods \cite{Chiang2012,Drews2016,Li2016,LiICIP2016,Berman2017,Li2017prl,Zhuang2021} mainly focus on the accurate estimation of medium transmission. With the estimated medium transmission and other key underwater imaging parameters such as the homogeneous background light, a clean image can be obtained by reversing an underwater imaging physical model.
	Though physical model-based methods can achieve promising performance in some cases, they tend to produce unstable and sensitive results when facing challenging underwater scenarios. This is because 1) estimating the medium transmission is fundamentally ill-posed, 2) estimating multiple underwater imaging parameters is knotty for traditional methods, and 3) the assumed underwater imaging models do not always hold.
	
	\begin{figure}
		\begin{center}
			\begin{tabular}{c@{ }c@{ }c@{ }c@{ }c}
				\includegraphics[height=3cm]{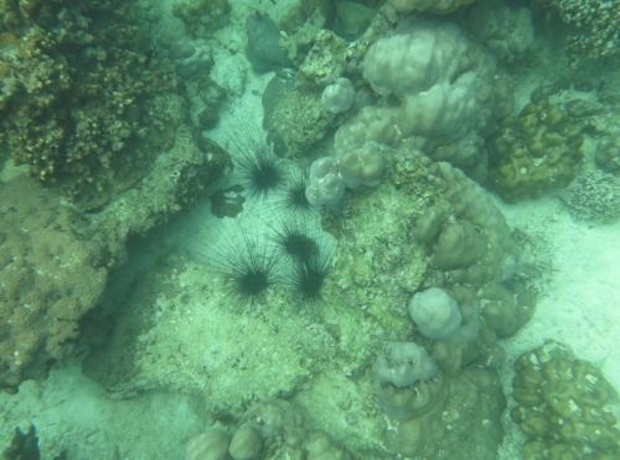}&
				\includegraphics[height=3cm]{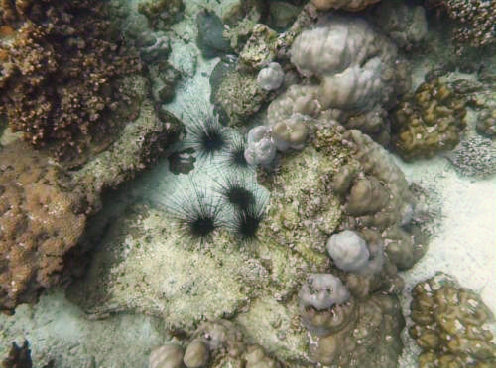}\\
				(a) input & (b)  Ucolor \\
				\includegraphics[height=3cm]{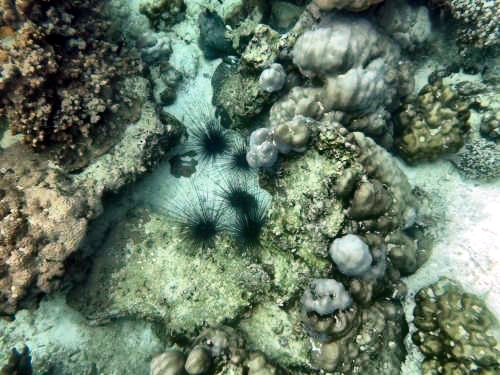}&
				\includegraphics[height=3cm]{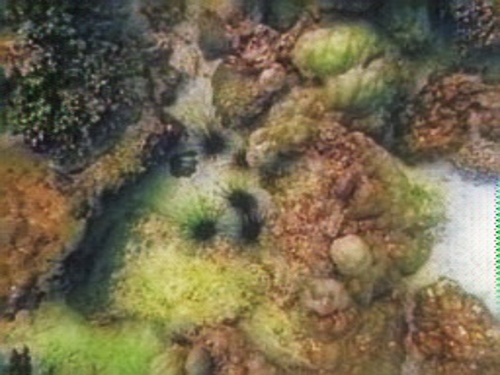}\\
				(c) Ancuti \etal~\cite{Ancuti2012} & (d) UcycleGAN~\cite{UCycleGAN}\\
			\end{tabular}
		\end{center}
		%\vspace{-0.35cm}
		\caption{Visual comparisons on a real underwater image. Our Ucolor removes both the greenish color deviation and the effect of scattering. In contrast, the compared methods either remain the color deviation or introduce extra color artifacts.}
		%	\vspace{-0.5cm}
		\label{fig:im_sample1}
	\end{figure}
	
	Recently, deep learning technology has shown impressive performance on underwater image enhancement~\cite{WaterGAN,UWCNN,UCycleGAN}. These deep learning-based methods often apply the networks that were originally designed for other visual tasks to underwater images. Thus, their performance is still far behind when compared with current deep visual models \cite{ZeroDCE,ZeroDCE++,LiTGRS,ASIFNet}. The main reason is that the design of current deep underwater image enhancement models neglects the domain knowledge of underwater imaging.

	In this work, we propose to solve the issues of color casts and low contrast of underwater images by leveraging rich encoder features and exploiting the advantages of physical model-based and learning-based methods.
	Unlike previous deep models \cite{UCycleGAN,Guo2019,Libenchmark} that only employ the features extracted from RGB color space, we examine the feature representations through a multi-color space encoder network, then highlight the most representative features via an attention mechanism.
	Such a manner effectively improves the generalization capability of deep networks, and also incorporates the characteristics of different color spaces into a unified structure.  
	This is rarely studied in the context of underwater image enhancement.
	Inspired by the conclusion that the quality degradation of underwater images can be reflected by the medium transmission \cite{Revisedmodel}, we propose a medium transmission-guided decoder network to enhance the response of our network towards quality-degraded regions.
	The introduction of medium transmission allows us to incorporate the advantage of physical model-based methods into deep networks, which accelerates network optimization and improves enhancement performance.
	Since our method is purely data-driven, it can tolerate the errors caused by inaccurate medium transmission estimation.

	In Fig. \ref{fig:im_sample1}, we present a representative example by the proposed Ucolor against two underwater image enhancement methods. 
	As shown, both the classical fusion-based method \cite{Ancuti2012} (Fig. \ref{fig:im_sample1}(c)) and the deep learning-based method \cite{UCycleGAN} (Fig. \ref{fig:im_sample1}(d)) fail to cope with the challenging underwater image with greenish tone and low contrast well. 
	In contrast, our Ucolor (Fig. \ref{fig:im_sample1}(b)) achieves the visually pleasing result in terms of color, contrast, and naturalness. The main contributions of this paper are highlighted as follows.
	\begin{itemize}
		\item We propose a  multi-color space encoder network coupled with an attention mechanism for incorporating the characteristics of different color spaces into a unified structure and adaptively selecting the most representative features. 
		\item We propose a medium transmission-guided decoder network to enforce the network to pay more attention to quality-degraded regions. It explores the complementary merits between domain knowledge of underwater imaging and deep neural networks. 
		\item Our Ucolor achieves state-of-the-art performance on several recent benchmarks in terms of both visual quality and quantitative metrics.
	\end{itemize}

The rest of this paper is organized as follows. Section~\ref{Related Work} presents the related works of underwater image enhancement. Section~\ref{Proposed Method} introduces the proposed method. In Section~\ref{Experiments}, the qualitative and quantitative experiments are conducted. Section~\ref{Conclusion} concludes this paper.
	
	\begin{figure*}[htb]
		\centering
		\centerline{\includegraphics[width=18cm,height=7cm]{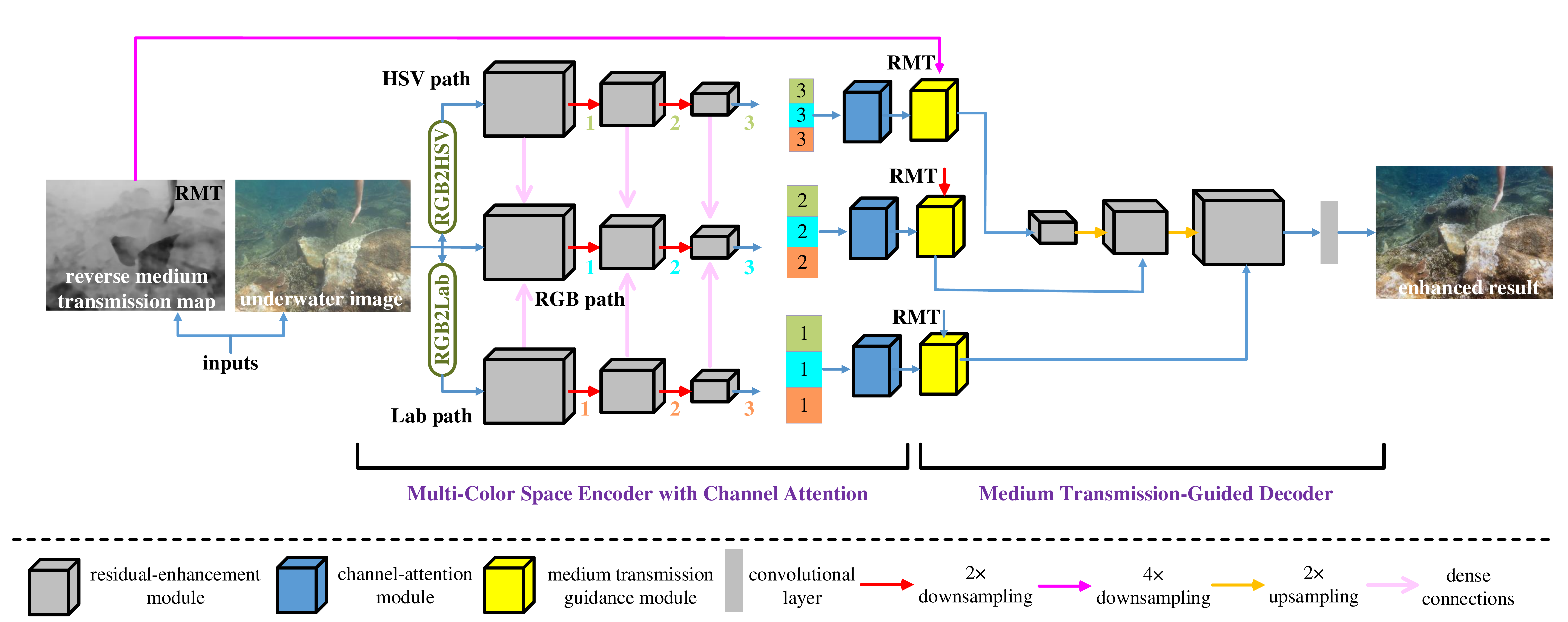}}
		\caption{Overview of the architecture of Ucolor. Our Ucolor consists of a multi-color space encoder network and a medium transmission-guided decoder network. In our method, we normalize the values of the medium transmission map to [0,1] and feed the reverse medium transmission map  (denoted as RMT) to the medium transmission guidance module. `downsampling' is implemented by max pooling, while `upsampling' is implemented  by  bilinear  interpolation. `dense connections' represents the concatenation operation along the channel dimension for each set of features from the corresponding convolutional layer in different color-space encoder paths. `convolutional layer' has the kernel of size 3$\times$3 and stride 1. In the Ucolor, all convolutional layers adopt kernels of size 3$\times$3 and stride 1. A detailed network structure with the hyper-parameters can be found in the Supplementary Material.}
		\label{fig:pipeline}
	\end{figure*}

	\section{Related Work}
	\label{Related Work}
	In addition to extra information~\cite{Bryson2012,Akkaynak2019} and specialized hardware devices \cite{Schechner2004}, underwater image enhancement can be roughly classified into two groups: traditional methods and deep learning-based methods.
	
	\noindent
	\textbf{Traditional Methods.} Early attempts aim to adjust the pixel values for visual quality improvement, such as dynamic pixel range stretching~\cite{Iqbal2010}, pixel distribution adjustment~\cite{Ghani2015}, and image fusion~\cite{Ancuti2012,Ancuti2018,Gao2019}. 
	For example, Ancuti \etal \cite{Ancuti2012}  first obtained the color-corrected and contrast-enhanced versions of an underwater image, then computed the corresponding weight maps, finally combined the advantages of different versions. Ancuti \etal \cite{Ancuti2018} further improved the fusion-based underwater image enhancement strategy and proposed to blend two versions that are derived from a white-balancing algorithm based on a multiscale fusion strategy.  Most recently, based on the observation that the information contained in at least one color channel is close to completely lost under adverse conditions such as hazy nighttime, underwater, and non-uniform artificial illumination, Ancuti \etal \cite{Ancuti2019} proposed a color channel compensation (3C) pre-processing method. As a pre-processing step, the 3C operator can improve traditional restoration methods.
	
Although these physical model-free methods can improve the visual quality to some extent, they omit the underwater imaging mechanism and thus tend to produce either over-/under-enhanced results or introduce artificial colors. For example, the color correction algorithm in \cite{Ancuti2012} is not always reliable when encountering diverse and challenging underwater scenes. Compared with these methods, our method takes underwater image formation models into account and employs the powerful learning capability of deep networks, making the enhanced images look more natural and visually pleasing.
	
	The widely used underwater image enhancement methods are physical model-based, which estimate the parameters of underwater imaging models based on prior information. These priors include
	red channel prior~\cite{Galdran2015}, underwater dark channel prior~\cite{Drews2016}, minimum information prior~\cite{Li2016}, blurriness prior~\cite{Peng2017}, general dark channel prior~\cite{Peng2018}, \etc. 
	For example, Peng and Cosman \cite{Peng2017} proposed an underwater image depth estimation algorithm based on image blurriness and light absorption. With the estimated depth, the clear underwater image can be restored based on an underwater imaging model.
	Peng \etal \cite{Peng2018} further proposed a generalization of the dark channel prior to deal with diverse images captured under severe weather. A new underwater image formation model was proposed in \cite{Revisedmodel}. Based on this model, an underwater image color correction method was presented using underwater RGB-D images \cite{Akkaynak2019}.
	
	These physical model-based methods are either time-consuming or sensitive to the types of underwater images~\cite{Liubenchmark}. Moreover, the accurate estimation of complex underwater imaging parameters challenges current physical model-based methods \cite{Galdran2015,Drews2016,Li2016,Peng2017,Peng2018}. For example, the blurriness prior used in  \cite{Peng2017} does not always hold, especially for clear underwater images. In contrast, our method can more accurately restore underwater images by exploiting the advantages of both physical model-based and data driven-based methods.

	\noindent
	\textbf{Deep Learning Models.} 
	The emergence of deep learning has led to considerable improvements in low-level visual tasks \cite{SRCNN,Pan18,CycleGAN,GuoTIP18,Ren2019,LiZero2020}. 
	There are several attempts made to improve the performance of underwater image enhancement through deep learning strategy \cite{USurvey}. As a pioneering work, Li \etal~\cite{WaterGAN} employed a Generative Adversarial Network (GAN) and an image formation model to synthesize degraded/clean image pairs for supervised learning.
	To avoid the requirement of paired training data, a weakly supervised underwater color correction network (UCycleGAN) was proposed in~\cite{UCycleGAN}. Furthermore, Guo \etal~\cite{Guo2019} introduced a multi-scale dense GAN for robust underwater image enhancement. Li \etal ~\cite{UWCNN} proposed to simulate the realistic underwater images according to different water types and an underwater imaging physical model. With ten types of synthesized underwater images, ten underwater image enhancement (UWCNN) models were trained, in which each UWCNN model was used to enhance the corresponding type of underwater images.
	Recently, Li \etal ~\cite{Libenchmark} collected a real paired underwater image dataset for training deep networks and proposed a gated fusion network to enhance underwater images. This proposed gated deep model requires three preprocessing images including a Gamma correction image, a contrast improved image, and a white-balancing image as the inputs of the gated network.  A wavelet corrected transformation was proposed for underwater image enhancement in \cite{Adarsh2019}. Yang \etal~\cite{Yang2020} proposed a conditional generative adversarial network to improve the perceptual quality of underwater images.
	
	These underwater image enhancement models usually  apply the existing deep network structures for general purposes to underwater images and neglect the unique characteristics of underwater imaging. For example, \cite{UCycleGAN} directly uses the CycleGAN \cite{CycleGAN} network structure, and \cite{Libenchmark} adopts a simple multi-scale convolutional network. For unsupervised models \cite{UCycleGAN,Guo2019}, they still inherit the disadvantage of GAN-based models, which produces unstable enhancement results. In \cite{UWCNN}, facing an input underwater image, how to select the corresponding UWCNN model is challenging.
	Consequently, the robustness and generalization capability of current deep learning-based underwater image enhancement models are limited and unsatisfactory.

In contrast to existing deep learning-based underwater image enhancement methods, our method has the following unique characteristics: 
			1) the multi-color space encoder network coupled with an attention mechanism that enables the diverse feature representations  from multi-color space and adaptively selects the most representative  information;
			2) the medium transmission-guided decoder network that incorporations the domain knowledge of underwater imaging into deep structures by tailoring the attention mechanism for emphasizing the quality-degraded regions;
			3) our method does not require any pre-processing steps and adopts supervised learning, thus producing more stable results;
			4) our method adopts end-to-end training and is able to handle most underwater scenes in a unified structure;
			and 5) our method achieves outstanding performance on various underwater image datasets.
	These innovations provide new ideas for exploring the complementary merits between domain knowledge of underwater imaging and deep learning strategy and the advantages of multi-color space encoder.

	\begin{figure*}
		\begin{center}
			\begin{tabular}{c@{ }c@{ }c@{ }c@{ }c@{ }c@{ }c@{ }c}
				\includegraphics[height=1.9cm,width=2.35cm]{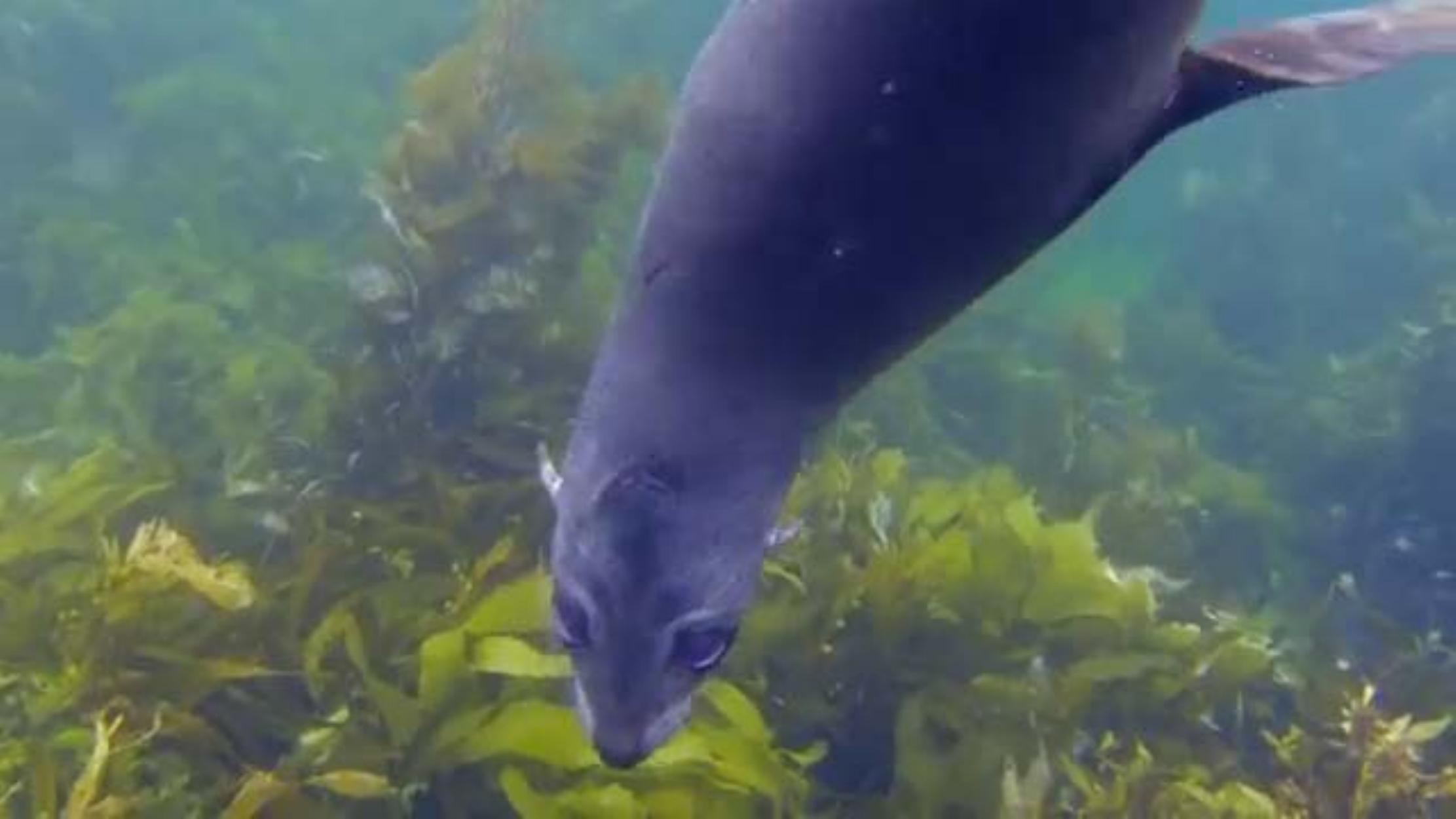} &
				\includegraphics[height=1.9cm,width=2.35cm]{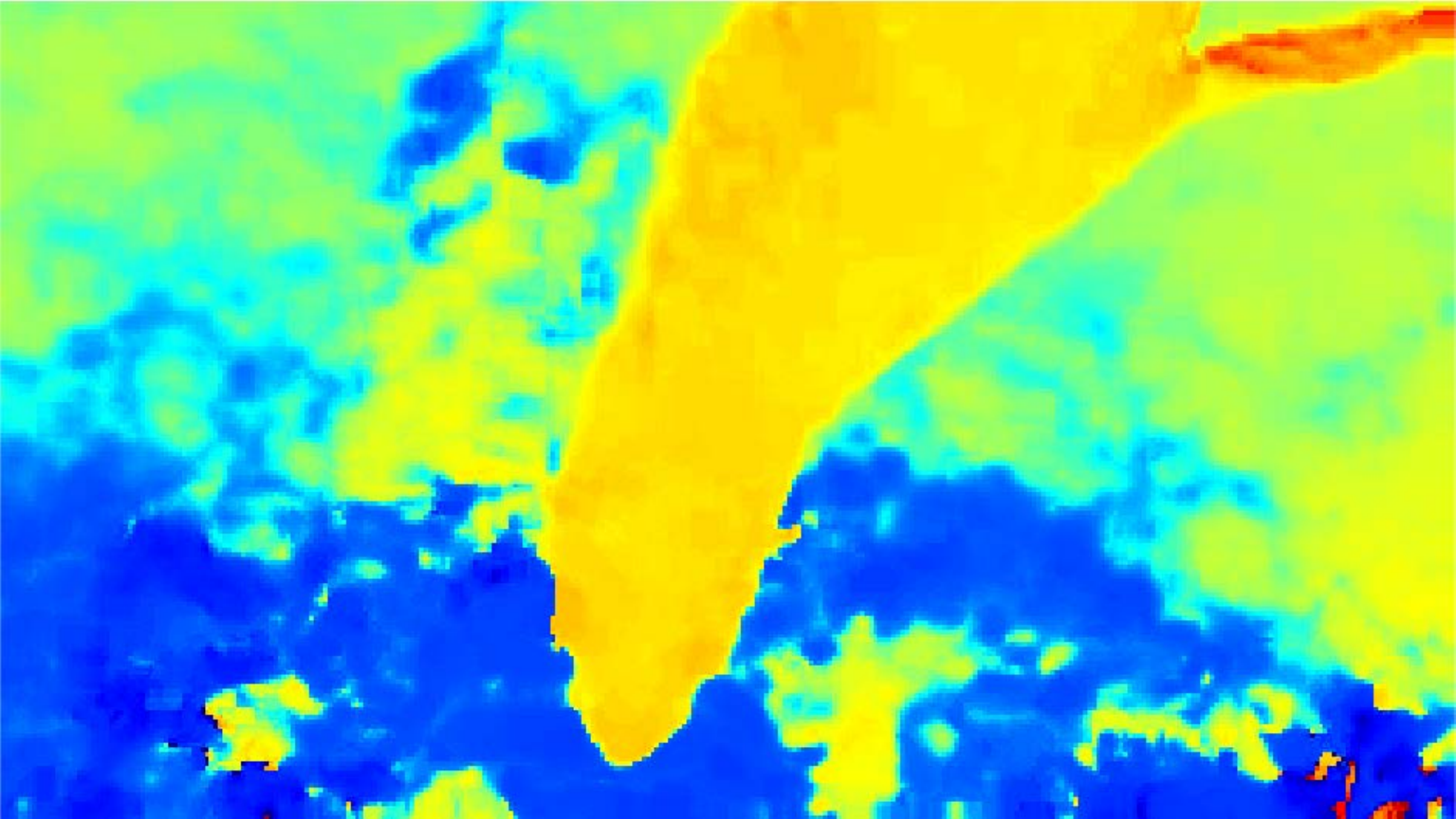} &
				\includegraphics[height=1.9cm,width=2.35cm]{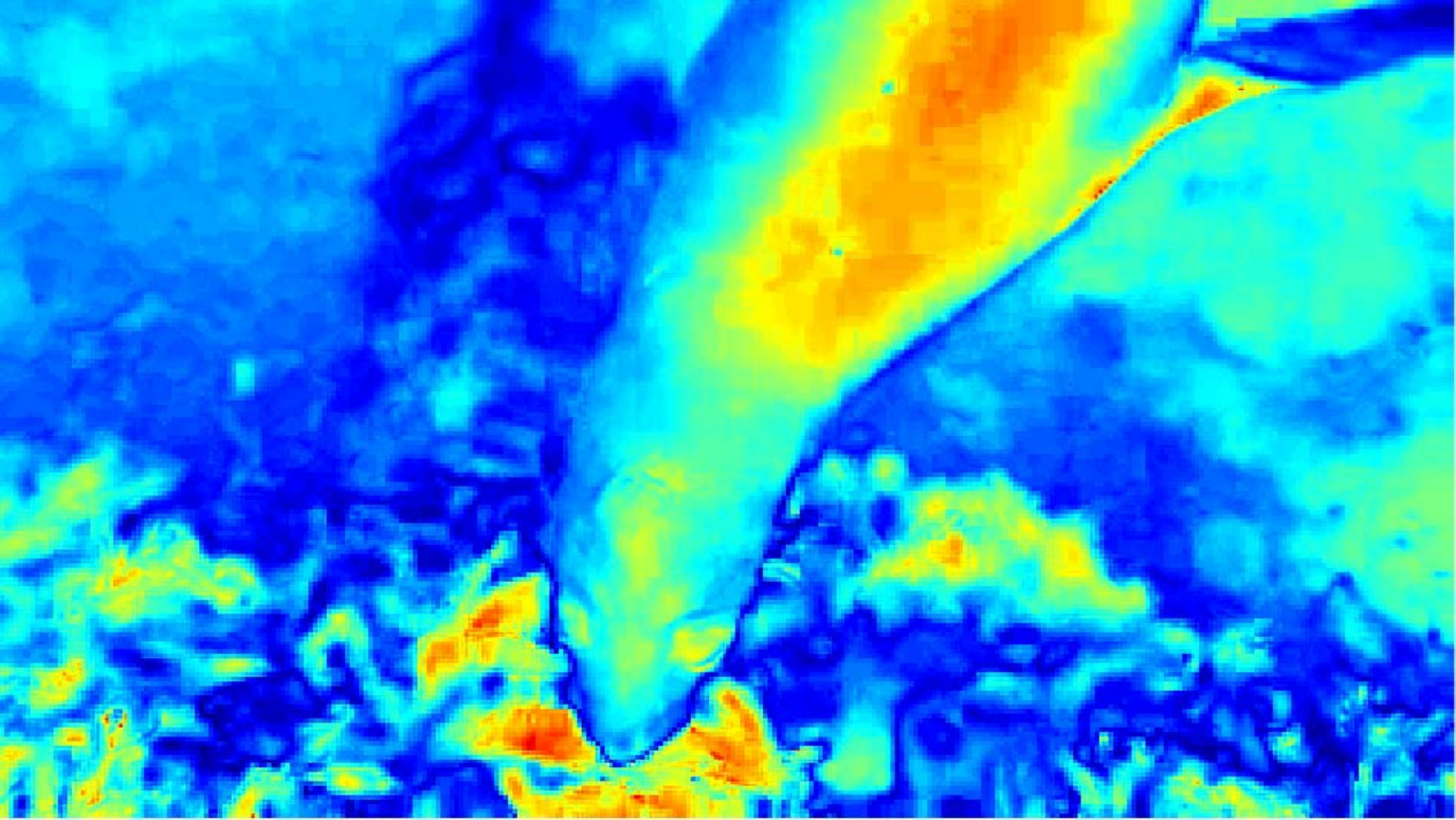} &
				\includegraphics[height=1.9cm,width=2.35cm]{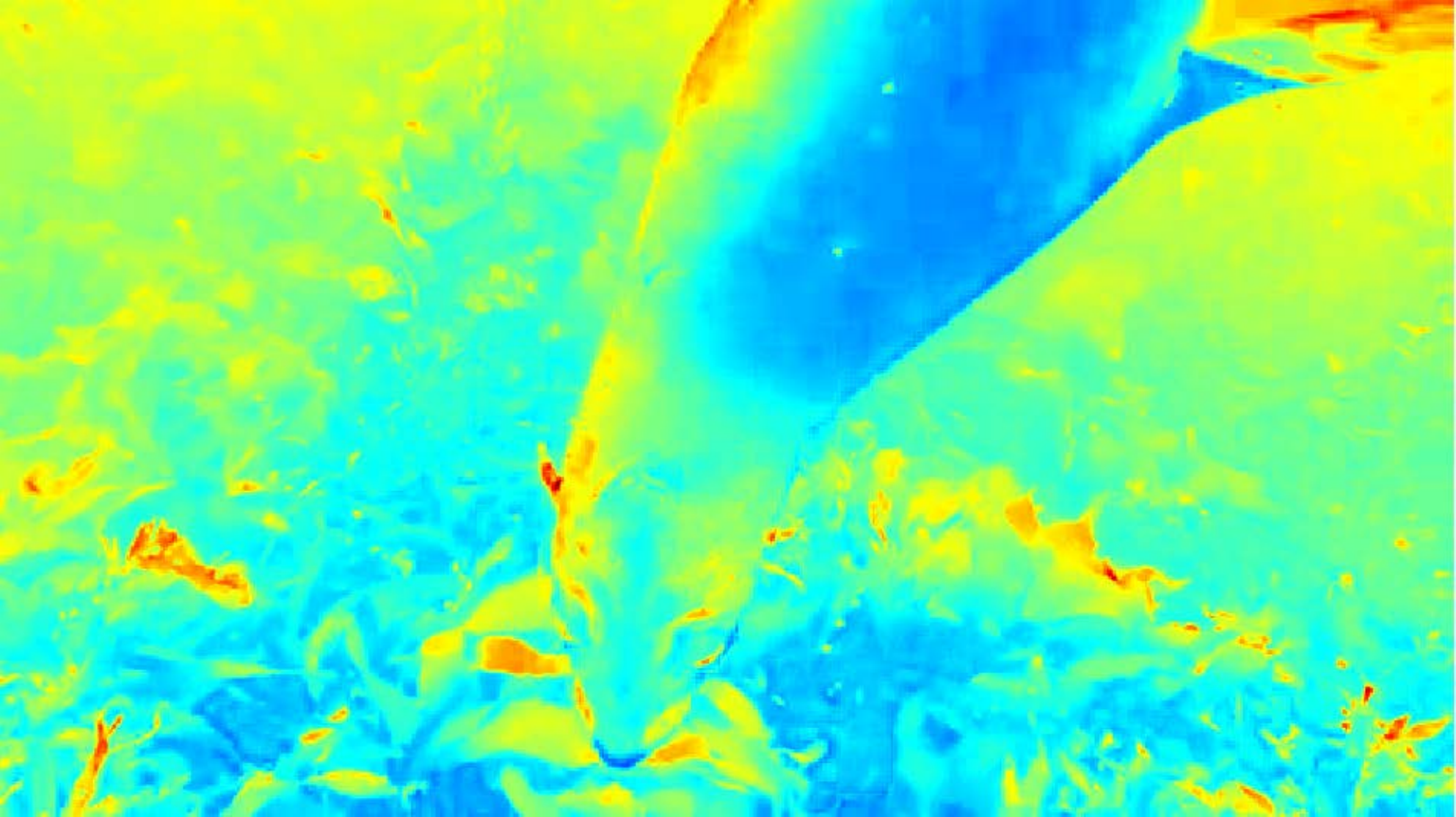} &
				\includegraphics[height=1.9cm,width=2.35cm]{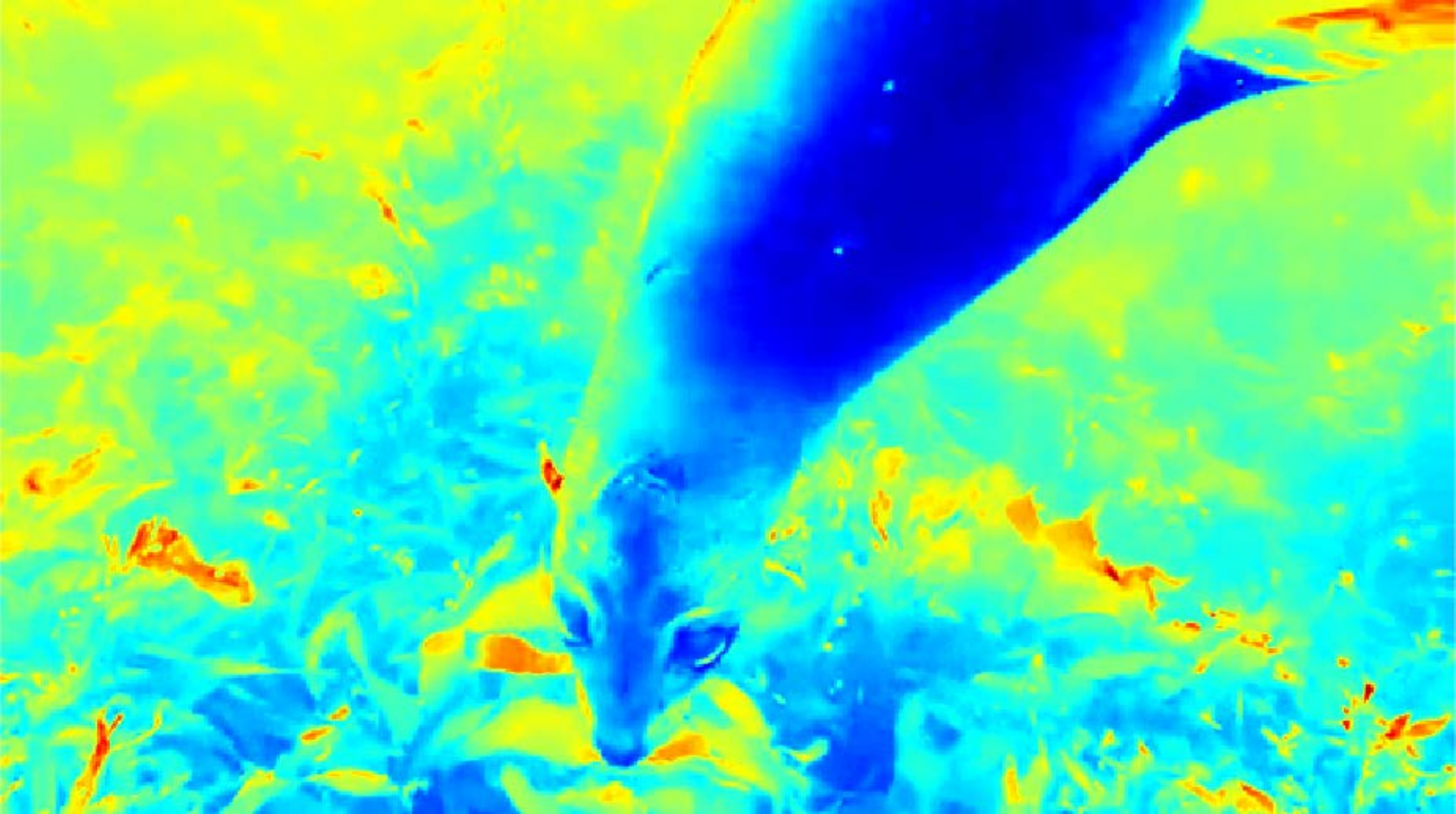} &
				\includegraphics[height=1.9cm,width=2.35cm]{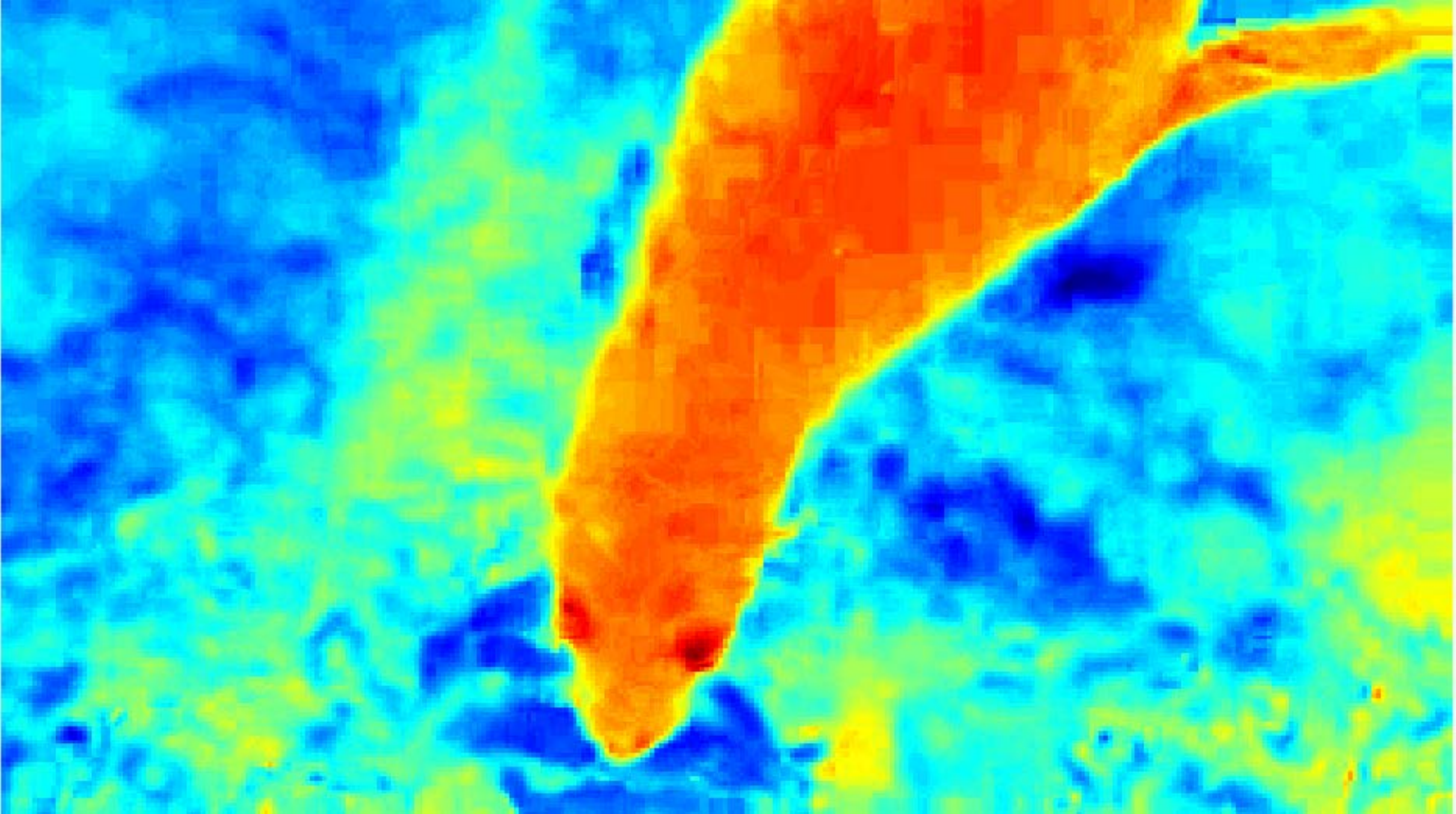} &
				\includegraphics[height=1.9cm,width=2.35cm]{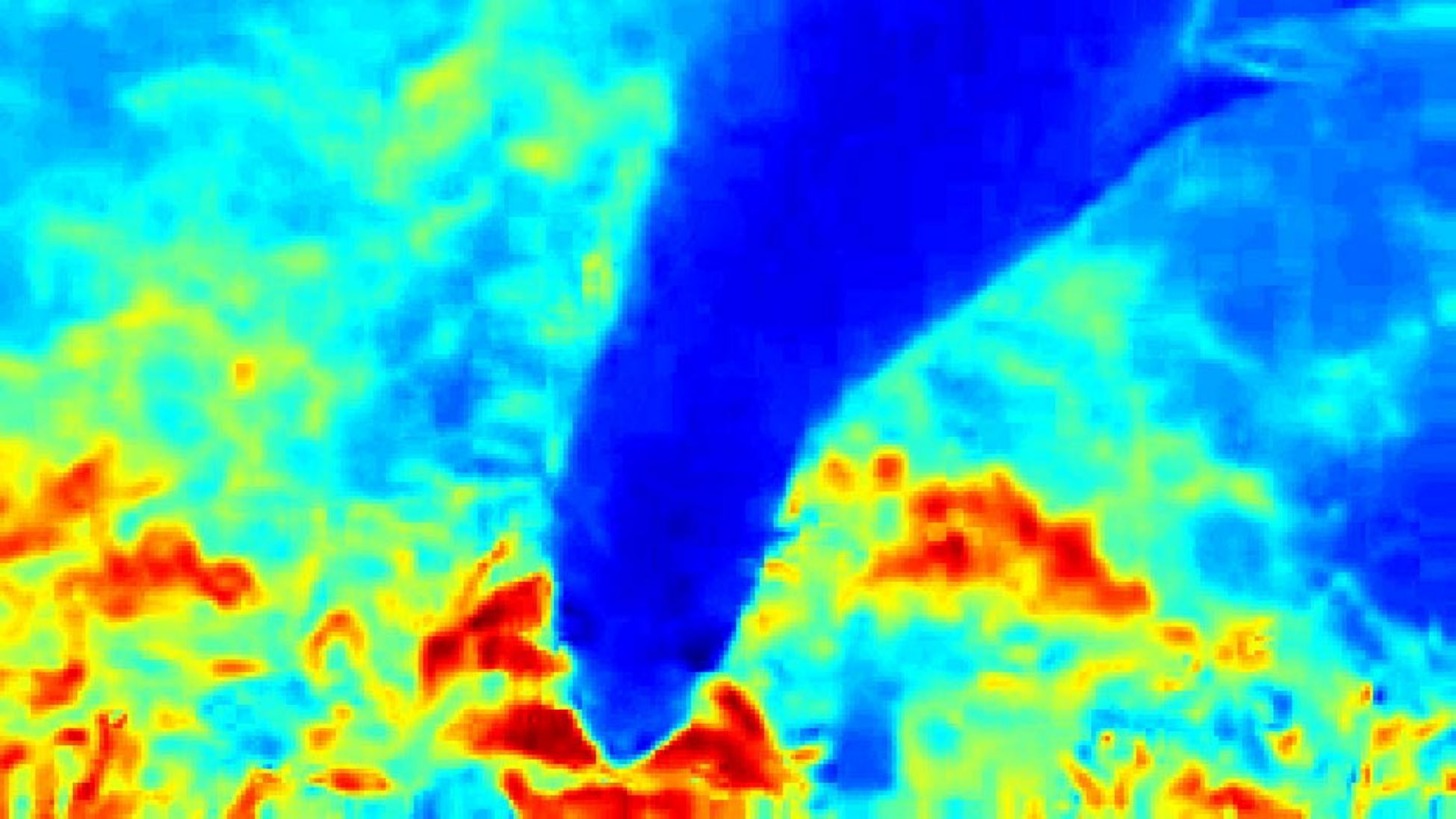} \\
				(a) RGB & (b) Hue & (c) Saturation  & (d) Value & (e) Lightness & (f) a component & (g) b component\\
			\end{tabular}
		\end{center}
		\caption{A visual example of the channels extracted from different color spaces. For visualization, we normalize the values to the range of [0,1]. The channels in (b)-(g) are represented by heatmaps, where the color ranging from blue to red represents the value from small to large.}
		\label{fig:color_map}
	\end{figure*}

	\section{Proposed Method}
	\label{Proposed Method}
	We present the overview architecture of Ucolor in Fig.~\ref{fig:pipeline}. 
	\textbf{In the multi-color space encoder network}, an underwater image first goes through color space transformation. Three encoder paths named HSV path, RGB path, and Lab path are formed. 
	In each path, the input is forwarded to three  serial redisual-enhancement modules, thus obtaining three levels of feature representations using a 2$\times$ downsampling operation  (noted Fig.~\ref{fig:pipeline} as 1, 2, and 3). 
	Simultaneously, we enhance the RGB path by densely connecting the features of the RGB path with the corresponding features of the HSV path and the Lab path. 
	We then concatenate the same level features of these three parallel paths to form three sets of multi-color space encoder features. 
	At last, we separately feed these three sets of features to the corresponding channel-attention module that serves as a tool to spotlight the most representative and informative features. 
	\textbf{In the medium transmission-guided decoder network}, the selected encoder features by channel-attention modules and the same sizes of reverse medium transmission (RMT) map are forwarded to the medium transmission guidance module for emphasizing quality-degraded regions. Here, we employ the max pooling operation to achieve different sizes of RMT maps. Then, the outputs of the medium transmission guidance modules are fed to the corresponding residual-enhancement module. After three serial residual-enhancement modules and two 2$\times$ upsampling operations, the decoder features are forwarded to a convolution layer for reconstructing the result.

In what follows, we detail the key components of our method,  including the multi-color space encoder (Sec. \ref{Multi-Color Space Encoder}), the residual-enhancement module (Sec. \ref{Residual-Enhancement Module}), the channel-attention module (Sec. \ref{Channel-Attention Module}),  the medium transmission guidance module (Sec. \ref{Medium Transmission Guidance Module}), and the loss function (Sec. \ref{Loss Function}).

	\subsection{Multi-Color Space Encoder}
	\label{Multi-Color Space Encoder}
	
	Compared with terrestrial scene images, the color deviations of underwater images cover more comprehensive ranges, differing from the bluish or greenish tone to a yellowish one. The diversity in color casts severely limits the traditional network architectures~\cite{Unet,SRnet}.
	Inspired by the traditional enhancement algorithms that operate in various color spaces \cite{Iqbal2010,Naik2003},  we extract features in three color spaces (RGB, HSV, and Lab) where the same image has different visual representations in various color systems as shown in Fig.~\ref{fig:color_map}.

	Concretely, the image is easy to store and display in RGB color space because of its strong color physical meaning. However, the three components (R, G, and B) are highly correlated, which are easy to be affected by the changes of luminance, occlusion, shadow, and other factors. By contrast, HSV color space can intuitively reflect the hue, saturation, brightness, and contrast of the image. Lab color space makes the colors better distributed, which is able to express all the colors that the human eye can perceive.

	These color spaces have obvious differences and advantages. To combine their properties for underwater image enhancement,  we incorporate the characteristics of different color spaces into a unified deep structure, where all the image degradation related components (color, hue, saturation, intensity, and luminance) can be taken into account. Moreover, the color difference of two points with a small distance in one color space may be large in other color spaces. Thus, the multiple color spaces embedding can facilitate the measurement of 
	color deviations of underwater images.
	Additionally, the multi-color space encoder brings more nonlinear operations during color space transformation. It is known that the nonlinear transformation generally improves the performance of deep models \cite{Simonyan2015}. In the ablation study, we will analyze the contribution of each color space.

	\begin{figure}[!t]
		\centering
		\centerline{\includegraphics[width=1\linewidth]{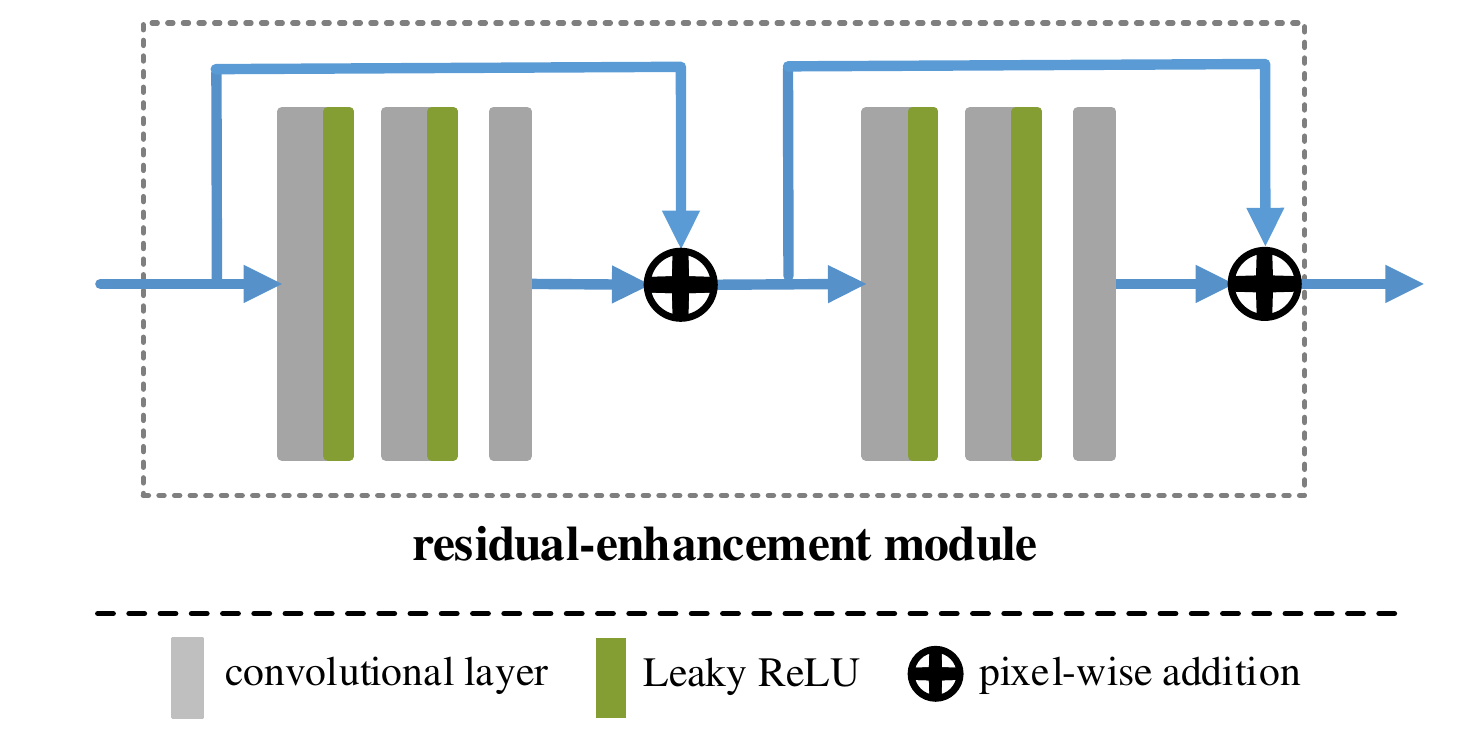}}
		\caption{The schematic illustration of the residual-enhancement module. Each residual-enhancement module is composed of two residual blocks, where each block is built by three stacked convolutions followed by the Leaky ReLU activation function, except for the last one. After each residual block, a pixel-wise addition is used as an identity connection.}
		\label{fig:redisual}
	\end{figure}

	\subsection{Residual-Enhancement Module}
	\label{Residual-Enhancement Module}
	
	Fig.~\ref{fig:redisual} presents the details of the residual-enhancement module. This residual-enhancement module aims to preserve the data fidelity and address gradient vanishing~\cite{RedisualNet}. In each residual-enhancement module, the convolutional layers have an identical number of filters. The numbers of filters are progressively increased from 128 to 512 by a factor 2 in the encoder network while they are decreased from 512 to 128 by a factor 2 in the decoder network.  All the convolutional layers have the same kernel sizes of 3$\times$3 and stride 1.

	\begin{figure*}[!tb]
		\centering
		\centerline{\includegraphics[width=0.8\linewidth]{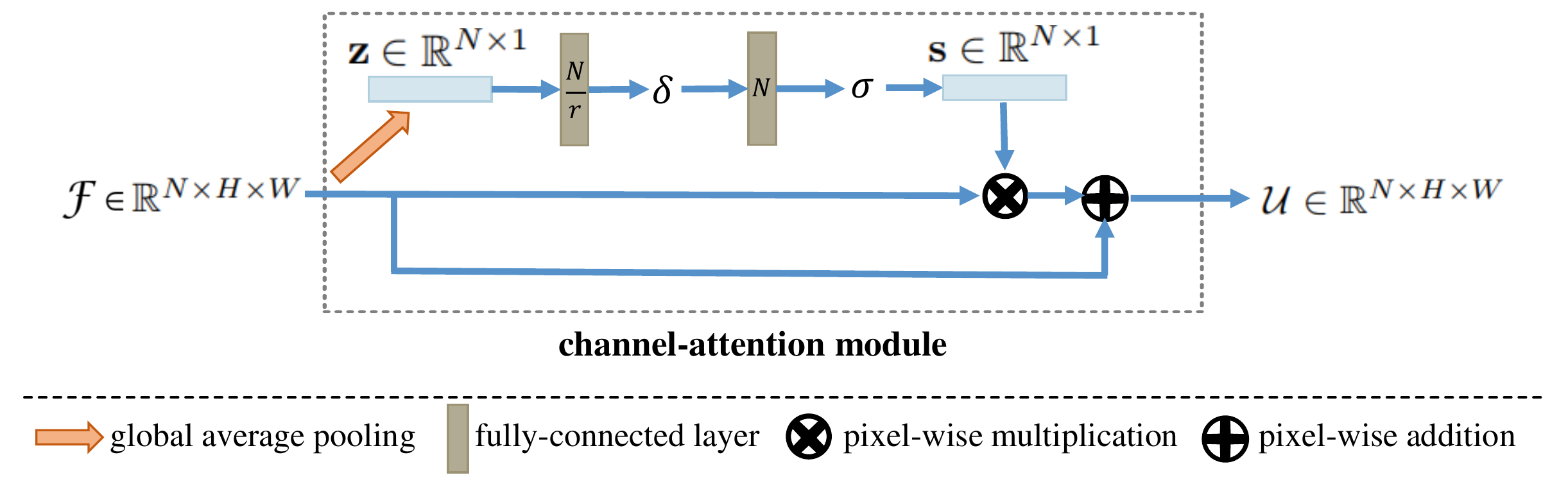}}
		\caption{The schematic illustration of the channel-attention module. The channel-attention module performs feature recalibration using global information. After going through global average pooling and fully-connected layers, the informative features  are emphasized and the less usefull features are suppressed in the input features $\mathcal{F}$, thus obtaining rescaled features  $\mathcal{U}$.}
		\label{fig:channel}
	\end{figure*}
	
	\subsection{Channel-Attention Module}
	\label{Channel-Attention Module}
	In view of the specific definition of each color space, these features extracted from three color spaces should have different contributions. Therefore, we employ a channel-attention module to explicitly exploit the interdependencies between the channel features extracted from different color spaces. The details of the channel-attention module are depicted in Fig. \ref{fig:channel}.
	
	Assume the input features $\mathcal{F}=\textsf{Cat}(\mathbf{F}_{1},\mathbf{F}_{2},\cdots, \mathbf{F}_{N})$ $\in$ $\mathbb{R}^{N\times H\times W}$, where $\mathbf{F}$ is a feature map from one path at a specific level (the level is denoted as 1, 2, and 3 in Fig.~\ref{fig:pipeline}), $N$ is the number of feature maps, $\textsf{Cat}$ represents the feature concatenation; and $H$ and $W$ are the height and width of input image, respectively. We first perform the global average pooling on input features $\mathcal{F}$, leading to a channel descriptor $\mathbf{z}\in\mathbb{R}^{N\times 1}$,  which is an embedded global distribution of channel-wise feature responses. The $k$-th entry of $\mathbf{z}$ can be expressed as:
	\begin{equation}
		\label{equ_1}
		z_{k} =\frac{1}{H\times W} \sum_{i}^{H} \sum_{j}^{W} \mathbf{F}_{k}(i,j),
	\end{equation}
	where $k\in[1,N]$. To fully capture channel-wise dependencies,  a self-gating mechanism  \cite{ChannelAtt} is used to produce a collection of per-channel modulation weights $\mathbf{s}\in\mathbb{R}^{N\times 1}$:
	\begin{equation}
		\label{equ_2}
		\mathbf{s}=\mathscr{\sigma}(\mathbf{W}_{2}\ast(\mathscr{\delta}(\mathbf{W}_{1}\ast\mathbf{z}))),
	\end{equation}
	where $\mathscr{\sigma}(\cdot)$ represents the Sigmoid activation function,  $\mathscr{\delta}(\cdot)$ represents the ReLU activation function, $\ast$ denotes the convolution operation, and $\mathbf{W}_{1}$ and $\mathbf{W}_{2}$ are the weights of two fully-connected layers with the numbers of their output channels equal to  $\frac{N}{r}$ and $N$, respectively, where $r$ is set to 16 for reducing the computational costs. At last, these weights are applied to input features $\mathcal{F}$ to generate rescaled features $\mathcal{U}$ $\in$ $\mathbb{R}^{N\times H\times W}$. Moreover, to avoid gradient vanishing problem and keep good properties of original features, we treat the channel-attention weights in an identical mapping fashion:
	\begin{equation}
		\label{equ_3}
		\mathcal{U} =\mathcal{F}\oplus \mathcal{F}\otimes \mathbf{s},
	\end{equation}
	where $\oplus$ and $\otimes$ denote the pixel-wise addition and pixel-wise multiplication, respectively.

	\subsection{Medium Transmission Guidance Module}
	\label{Medium Transmission Guidance Module}
	
	\begin{figure}[!tb]
		\centering
		\centerline{\includegraphics[width=1\linewidth]{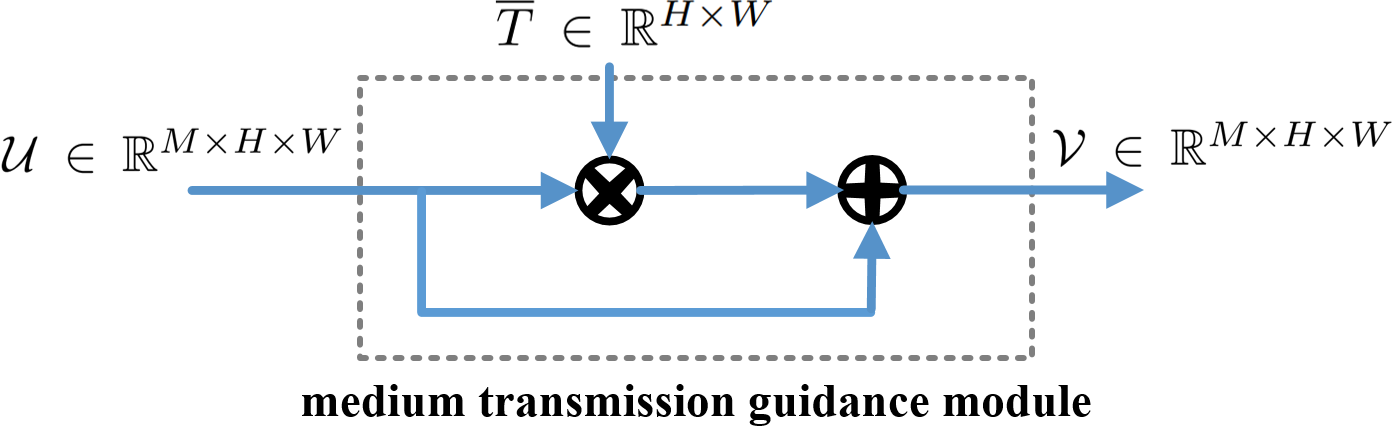}}
		\caption{The schematic illustration of the medium transmission guidance module. The RMT map $\overline{T}$ as a  feature selector is used to weight the importance of different spatial positions of the input features $\mathcal{U}$, thus obtaining highlighted output features $\mathcal{V}$.}
		\label{fig:depth}
	\end{figure}

	According to the image formation model in bad weather \cite{IFM1,IFM2}, which is widely used in image dehazing and underwater image restoration algorithms \cite{Chiang2012,Peng2017,Peng2018}, the quality-degraded image can be expressed as: 
	\begin{equation}
		I^c(x)=J^c(x)\otimes T(x)\oplus A^c(x)\otimes(1-T(x)), c\in\{r, g, b\}, 
		\label{eq:1}
	\end{equation}
	where $x$ indicates the pixel index, $I$ is the observed image,  $J$ is the clear image, $A$ is the homogeneous background light, and $T$ is the medium transmission that represents the percentage of scene radiance reaching the camera after reflecting in the medium, indicating the degrees of quality degradation in different regions.
	
	We incorporate the medium transmission map into the decoder network via the proposed medium transmission guidance module. Specifically, we use the reverse medium transmission (RMT) map (denoted as $\overline{T}\in\mathbb{R}^{H\times W}$) as the pixel-wise attention map. The RMT map $\overline{T}$ is obtained by $\mathbf{1}$-$T$ ($T\in\mathbb{R}^{H\times W}$ is the medium transmission map in the range of [0,1], and $\mathbf{1}\in\mathbb{R}^{H\times W}$ is the matrix with all elements equal to 1), which indicates that the higher quality degradation pixels should be assigned larger attention weights.
	
	Since the corresponding ground truth medium transmission map of an input underwater image is not available in practice, it is difficult to train a deep neural network for the estimation of medium transmission map. To solve this issue, we employ prior-based estimation algorithms to obtain the medium transmission map.  Inspired by the robust general dark channel prior \cite{Peng2018}, we estimate the medium transmission map as:
	\begin{equation}
		\tilde{T}(x)=\max\limits_{c,y\in\Omega(x)}(\frac{A^{c}-I^{c}(y)}{\max(A^{c},1-A^{c})}),
		\label{eq:55}
	\end{equation}
	where $\tilde{T}$ is the estimated medium transmission map,  $\Omega(x)$ represents a local patch of size 15$\times$15 centered at $x$, and $c$ denotes the color channel. As shown, the medium transmission estimation is related to the homogeneous background light $A$. In \cite{Peng2018}, the homogeneous background light is estimated based on the depth-dependent color change. Due to the limited space, we refer the readers to  \cite{Peng2018} for more details. We will compare and analyze the effects of the medium transmission maps estimated by different algorithms in the ablation study.

	With the RMT map, the schematic illustration of the proposed medium transmission guidance module is shown in Fig.~\ref{fig:depth}. As shown, we utilize the RMT map as a  feature selector to weight the importance of different spatial positions of the features.  The high-quality degradation pixels (the pixels with larger RMT values) are assigned higher weights, which can be expressed as:
	\begin{equation}
		\label{equ_3}
		\mathcal{V} =\mathcal{U}\oplus \mathcal{U}\otimes \overline{T},
	\end{equation}
	where $\mathcal{V}$ $\in$ $\mathbb{R}^{M \times H\times W}$ and $\mathcal{U}$ $\in$ $\mathbb{R}^{M \times H\times W}$ respectively represent the output features after the medium transmission guidance module and the input feature. We treat the RMT weights as an identity connection to avoid gradient vanishing and tolerate the errors caused by inaccurate medium transmission estimation. Besides, our purely data-driven framework also tolerates the inaccuracy of medium transmission maps. 
	
	\subsection{Loss Function}
	\label{Loss Function}
%Commonly, the deep models trained using the $\ell_{1}$ loss achieve high quantitative scores but produce artifacts, because the $\ell_{1}$ loss tends to penalize the large errors and tolerate small errors. Moreover, the perceptual loss $L_{per}$ measures the differences between the results and ground truth images at the semantic level, thus leading to visually pleasing results, while slightly compromising the quantitative scores. 
Following previous works \cite{Perceptual,LedigCVPR17}, to achieve a good balance between visual quality and quantitative scores, we use the linear combination of the $\ell_{2}$ loss $L_{\ell_{2}}$ and the perceptual loss $L_{per}$, and the  final loss $L_{f}$ for training our network is experessed as:
	\begin{equation}
		\label{equ_6}
		L_{f}=L_{\ell_{2}}+\lambda L_{per},
	\end{equation}
where $\lambda$ is empirically set to 0.01 for balancing the scales of different losses. Specifically, the $\ell_{2}$ loss measures the difference between the reconstructed result $\hat{J}$ and corresponding ground truth $J$ as:
	\begin{equation}
		\label{equ_2}
		L_{\ell_{2}}=\sum_{m=1}^{H}\sum_{n=1}^{W} (\hat{J}(m,n)-J(m,n))^2.
	\end{equation}
	The perceptual loss is computed based on the VGG-19 network $\phi$ \cite{Simonyan2015} pre-trained on the ImageNet dataset \cite{Deng2009}. Let $\phi_{j}(\cdot)$ be the $j$th convolutional layer. We measure the distance between the feature representations of the reconstructed result $\hat{J}$ and ground truth image $J$ as:
	\begin{equation}
		\label{equ_1}
		L_{per}=\sum_{m=1}^{H}\sum_{n=1}^{W} |\phi_{j}(\hat{J})(m,n)-\phi_{j}(J)(m,n)|.
	\end{equation}
	We compute the perceptual loss at layer relu5\_4 of the VGG-19 network. An ablation study towards the loss function will be presented.

	\section{Experiments}
	\label{Experiments}
	In this section, we first describe the implementation details, then introduce the experiment settings. We compare our method with representative methods and provide a series of ablation studies to verify each component of Ucolor. We show the failure case of our method at the end of this section.  Due to the limited space, more experimental results can be found in the supplementary material.
	
	\subsection{Implementation Details}
	To train the Ucolor, we randomly selected 800 pairs of underwater images from UIEB \cite{Libenchmark} underwater image enhancement dataset. The UIEB dataset includes 890 real underwater images with corresponding reference images. Each reference image was selected by 50 volunteers from 12 enhanced results. It covers diverse underwater scenes, different characteristics of quality degradation, and a broad range of image content, but the number of underwater images is inadequate to train our network. Thus, we incorporated 1,250 synthetic underwater images selected from a synthesized underwater image dataset \cite{UWCNN}, which includes 10 subsets denoted by ten types of water (I, IA, IB, II, and III for open ocean water and 1, 3, 5, 7, and 9 for coastal water).  To augment the training data, we randomly cropped image patches of size 128$\times$128.

	We implemented the Ucolor using the MindSpore Lite tool \cite{Minidspore}. A batch-mode learning method with a batch size of 16 was applied. The filter weights of each layer were initialized by Gaussian distribution, and the bias was initially set as a constant. We used ADAM for network optimization and fixed the learning rate to $1e^{-4}$. 
	
	\subsection{Experiment Settings}
	\noindent
	\textbf{Benchmarks.} 
	For testing, we used the rest 90 pairs of real data of the UIEB dataset, denoted as \textbf{Test-R90}, while 100 pairs of synthetic data from each subset of  \cite{UWCNN} forming a total of 1k pairs, denoted as \textbf{Test-S1000}.  We also conducted comprehensive experiments on three more benchmarks, \ie,~\textbf{Test-C60} \cite{Libenchmark}, \textbf{SQUID} \cite{UnderwaterPami2020}, and  \textbf{Color-Check7} \cite{Ancuti2018}. Test-C60 contains 60 real underwater images without reference images provided in the UIEB dataset \cite{Libenchmark}. Different from Test-R90, the images in  Test-C60 are more challenging, which fail current methods. The SQUID \cite{UnderwaterPami2020} dataset contains 57 underwater stereo pairs taken from four different dive sites in Israel. We used the 16 representative examples presented in the project page of SQUID\footnote{\url{http://csms.haifa.ac.il/profiles/tTreibitz/datasets/ambient_forwardlooking/index.html}} for testing. Specifically, for four dive sites (Katzaa, Michmoret, Nachsholim, Satil), four representative samples were selected from each dive site. Each image has a resolution of 1827$\times$2737.
	Color-Check7 contains 7 underwater Color Checker images taken with different cameras provided in \cite{Ancuti2018}, which are employed to evaluate the robustness and accuracy of underwater color correction. The cameras used to take the Color Checker pictures are Canon D10, Fuji Z33, Olympus Tough 6000, Olympus Tough 8000, Pentax W60, Pentax W80, and Panasonic TS1, denoted as Can D10, Fuj Z33, Oly T6000, Oly T8000, Pen W60, Pen W80, and Pan TS1 in this paper.

	\noindent
	\textbf{Compared Methods.} We compared our Ucolor with ten methods, including one physical model-free method (Ancuti \etal~\cite{Ancuti2012}), three physical model-based methods (Li \etal~\cite{Li2016}, Peng \etal~\cite{Peng2017}, GDCP~\cite{Peng2018}), four deep learning-based methods (UcycleGAN~\cite{UCycleGAN}, Guo \etal~\cite{Guo2019}, Water-Net~\cite{Libenchmark}, UWCNN~\cite{UWCNN}), and two  baseline deep models (denoted as Unet-U~\cite{Unet} and Unet-RMT~\cite{Unet}) that are trained using the same training data and loss functions as our Ucolor. Different from Ucolor, Unet-U and Unet-RMT employ the structure of Unet~\cite{Unet}. In addition, the inputs of Unet-U and  Unet-RMT are an underwater image and the concatenation of an underwater image and its RTM map that is estimated using the same algorithm as our Ucolor, respectively. The comparisons with the two baseline deep models aim at demonstrating the advantages of our network architecture and supplementing the compared deep learning-based methods.
	
	Since the source code of Ancuti \etal~\cite{Ancuti2012} is not publicly available, we used the code\footnote{\url{https://github.com/bilityniu/underimage-fusion-enhancement}} implemented by other researchers to realize code of \cite{Ancuti2012}. 
	For Li \etal~\cite{Li2016}, Peng \etal~ \cite{Peng2017}, GDCP~\cite{Peng2018}, UcycleGAN~\cite{UCycleGAN}, and Water-Net~\cite{Libenchmark}, we used the released codes to produce their results. The results of  Guo \etal~\cite{Guo2019} were provided by the authors.  Note that UcycleGAN~\cite{UCycleGAN} is an unsupervised methods, \ie, training with unpaired data, and thus there is no need to retrain it with our training data. Same as our Ucolor, Water-Net~\cite{Libenchmark} randomly selected the same number of training data from the UIEB dataset for training. For UWCNN~\cite{UWCNN}, we used the original UWCNN models, in which each UWCNN model was trained using the underwater images synthesized by one type of water. We discarded the UWCNN$\_$typeIA and UWCNN$\_$typeIB models because their results are similar to those of UWCNN$\_$typeI.
	Besides, we also retrained the UWCNN model (denoted as UWCNN$\_$retrain) using the same training data as our Ucolor.

		\begin{figure*}
		\begin{center}
			\begin{tabular}{c@{ }c@{ }c@{ }c@{ }}
				\includegraphics[height=3cm,width=4cm]{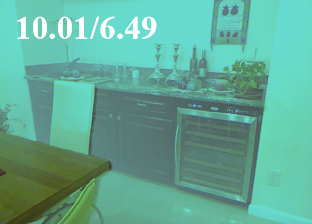} &
				\includegraphics[height=3cm,width=4cm]{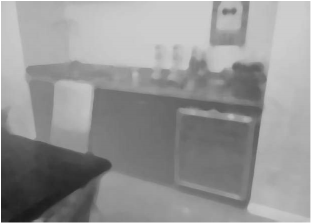} &
				\includegraphics[height=3cm,width=4cm]{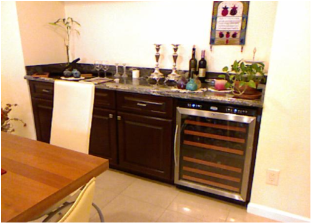}&
				\includegraphics[height=3cm,width=4cm]{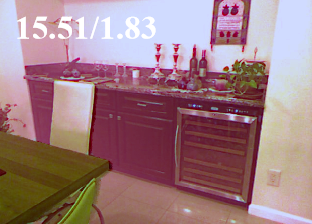}\\
				(a) input & (b) RMT map &(c) GT & (d)  GDCP \cite{Peng2018} \\
				\includegraphics[height=3cm,width=4cm]{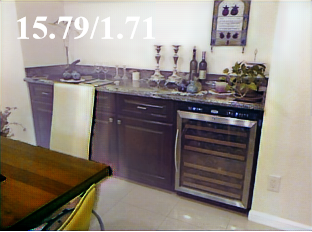}&
				\includegraphics[height=3cm,width=4cm]{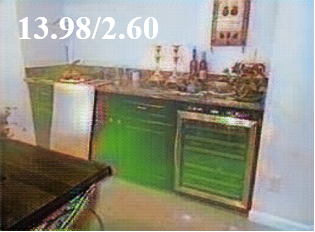}&
				\includegraphics[height=3cm,width=4cm]{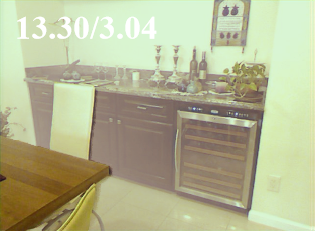}&
				\includegraphics[height=3cm,width=4cm]{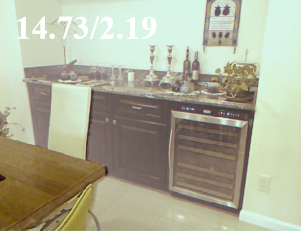}\\		
				(e) Guo \etal~\cite{Guo2019} & (f) UcycleGAN \cite{UCycleGAN} &  (g) UWCNN\_typeII \cite{UWCNN} & (h) UWCNN\_retrain  \cite{UWCNN} \\
				\includegraphics[height=3cm,width=4cm]{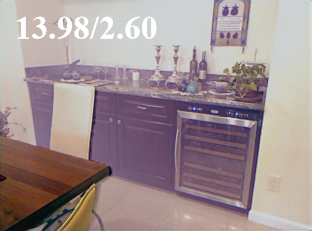}&
				\includegraphics[height=3cm,width=4cm]{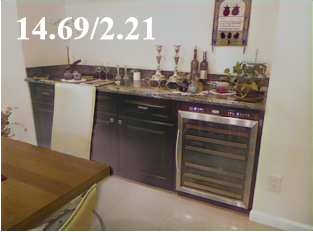}&
				\includegraphics[height=3cm,width=4cm]{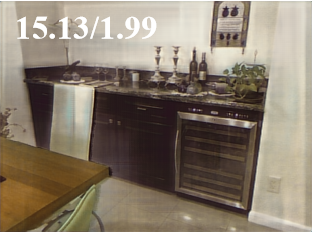}&
				\includegraphics[height=3cm,width=4cm]{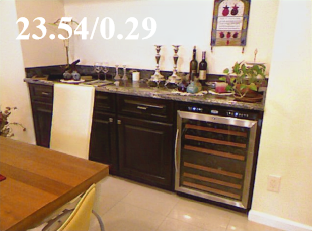}\\
				(i) Water-Net \cite{Libenchmark} & (j) Unet-U \cite{Unet} & (k) Unet-RMT \cite{Unet} & (l) Ucolor\\
			\end{tabular}
		\end{center}
		\caption{Visual comparisons on a synthetic underwater image sampled from \textbf{Test-S1000}. The numbers on the top-left corner of each image refer to its PSNR (dB)/MES ($\times 10^3$).}
		\label{fig:sythetic2}
	\end{figure*}

	\begin{figure*}
		\begin{center}
			\begin{tabular}{c@{ }c@{ }c@{ }c@{ }}
				\includegraphics[height=3cm,width=4cm]{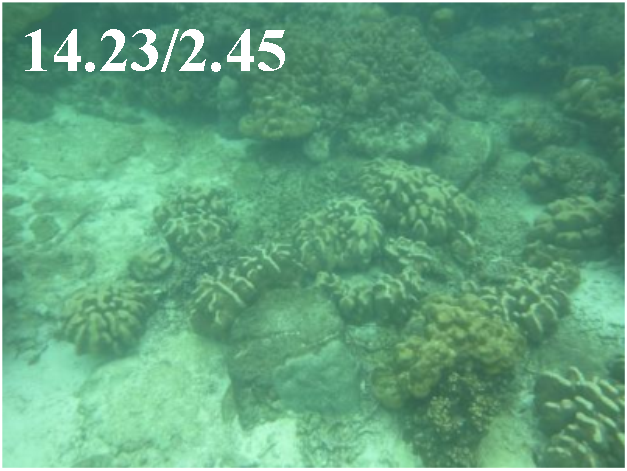} &
				\includegraphics[height=3cm,width=4cm]{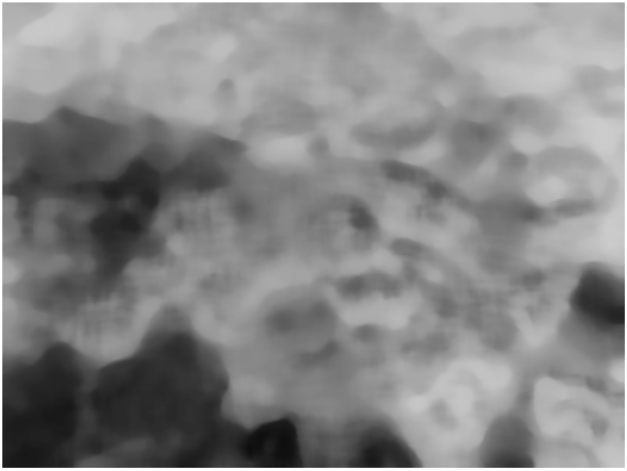} &
				\includegraphics[height=3cm,width=4cm]{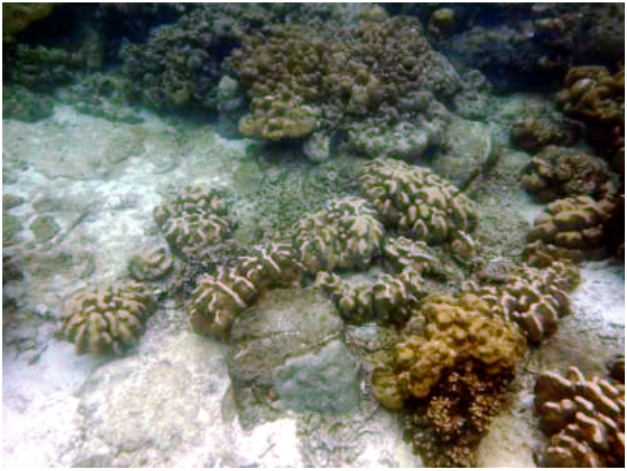}&
				\includegraphics[height=3cm,width=4cm]{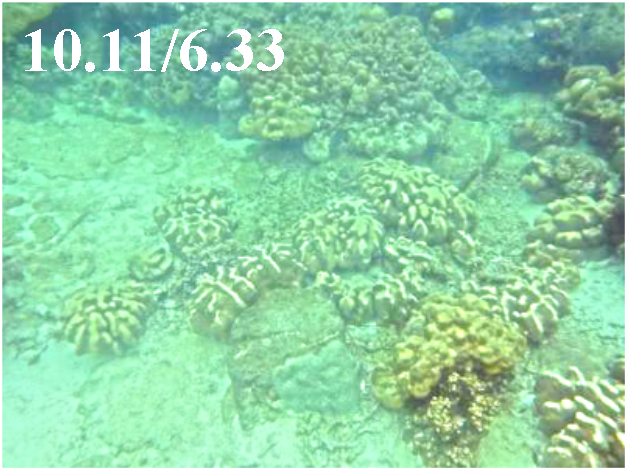}\\
				(a) input & (b) RMT map &(c) reference & (d)  GDCP \cite{Peng2018} \\
				\includegraphics[height=3cm,width=4cm]{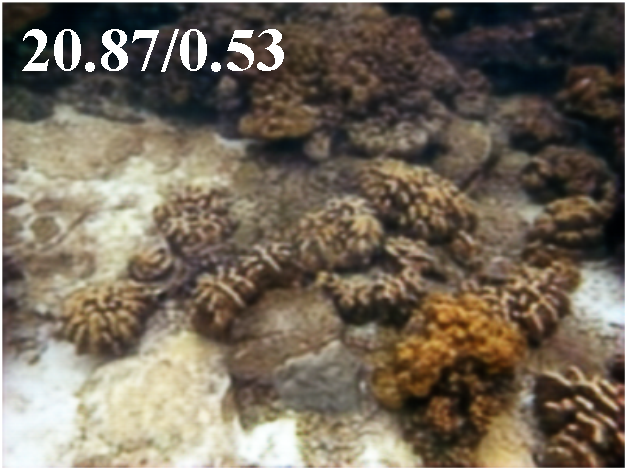}&
				\includegraphics[height=3cm,width=4cm]{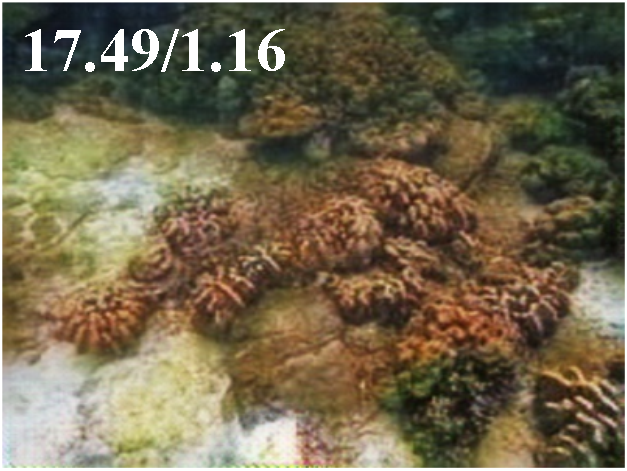}&
				\includegraphics[height=3cm,width=4cm]{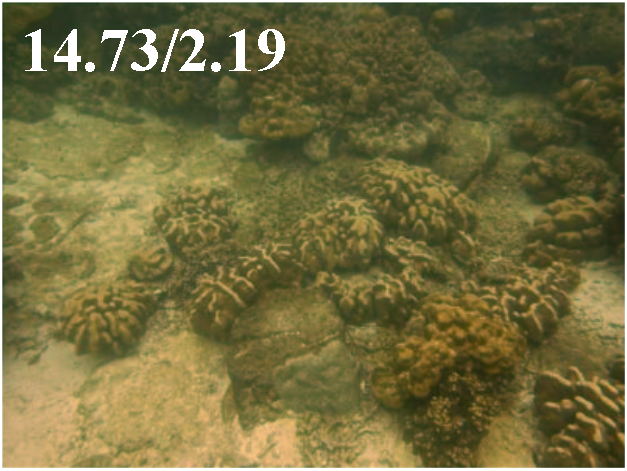}&
				\includegraphics[height=3cm,width=4cm]{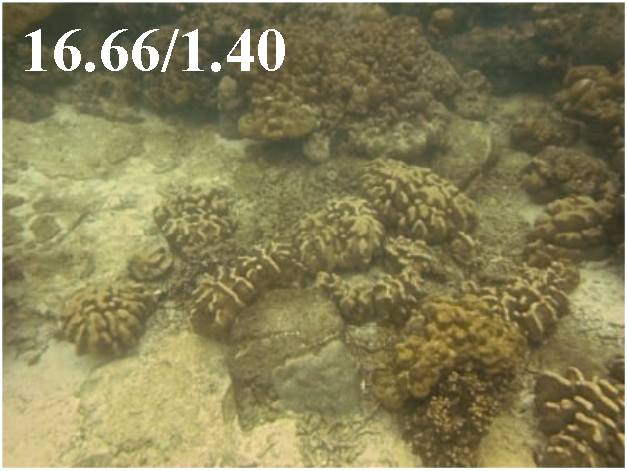}\\		
				(e) Guo \etal~\cite{Guo2019} & (f) UcycleGAN \cite{UCycleGAN} &  (g) UWCNN\_typeII \cite{UWCNN} & (h) UWCNN\_retrain  \cite{UWCNN} \\
				\includegraphics[height=3cm,width=4cm]{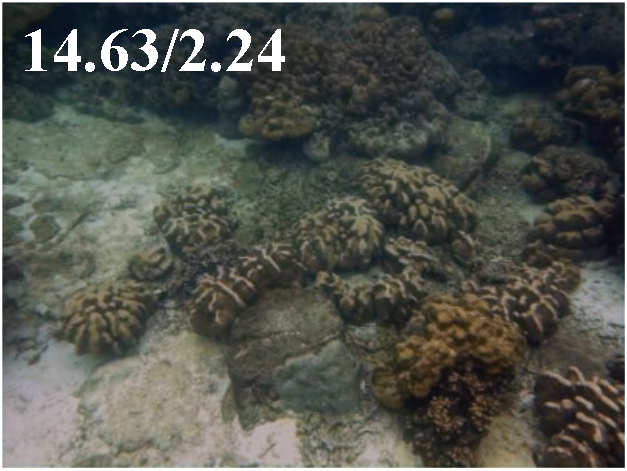}&
				\includegraphics[height=3cm,width=4cm]{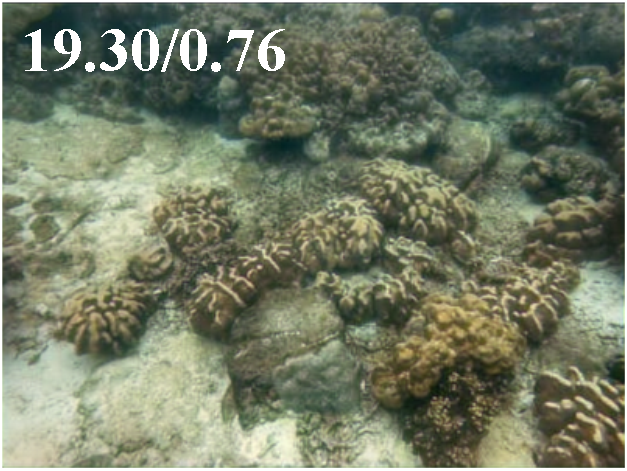}&
				\includegraphics[height=3cm,width=4cm]{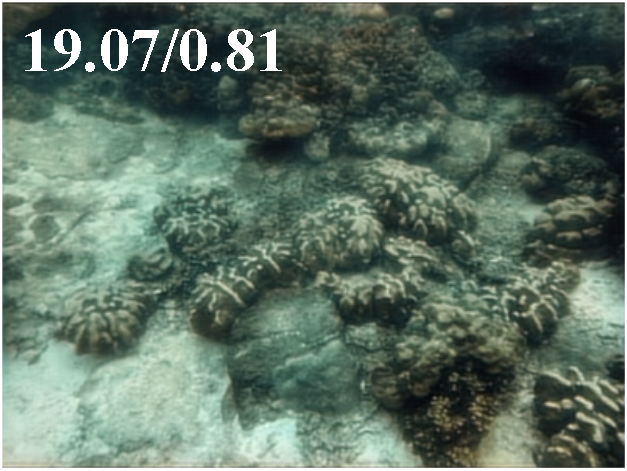}&
				\includegraphics[height=3cm,width=4cm]{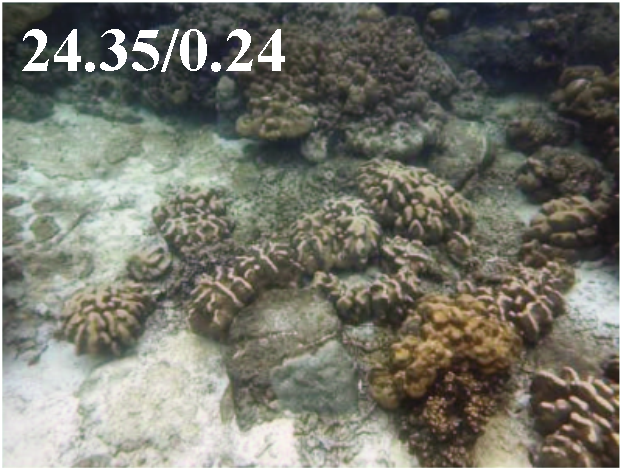}\\
				(i) Water-Net \cite{Libenchmark} & (j) Unet-U \cite{Unet} & (k) Unet-RMT \cite{Unet} & (l) Ucolor\\
			\end{tabular}
		\end{center}
		\caption{Visual comparisons on a typical real underwater image with obvious greenish color deviation and low-contrast sampled from \textbf{Test-R90}. The numbers on the top-left corner of each image refer to its PSNR (dB)/MES ($\times 10^3$).}
		\label{fig:real_data}
	\end{figure*}

	\begin{figure*}
		\begin{center}
			\begin{tabular}{c@{ }}
				\includegraphics[width=0.9\linewidth]{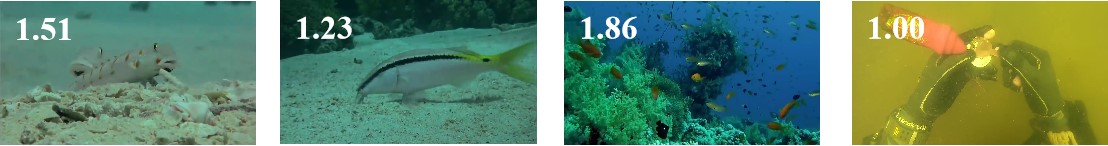} \\
				(a) input \\
				\includegraphics[width=0.9\linewidth]{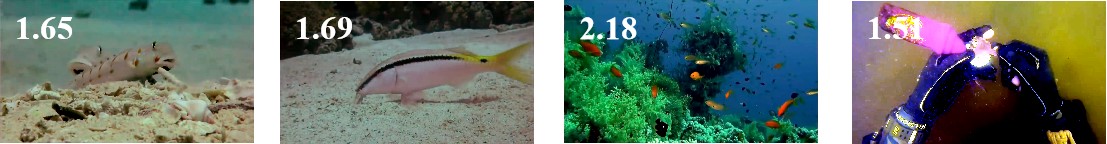}\\
				(b)  Peng \etal~\cite{Peng2017} \\
				\includegraphics[width=0.9\linewidth]{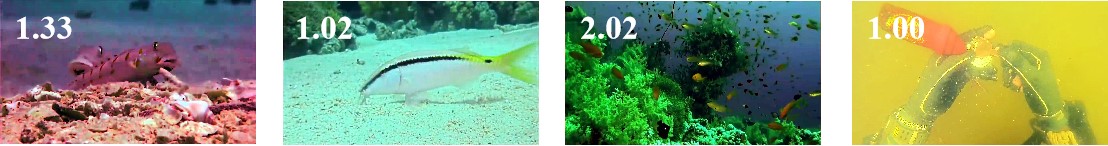}\\
				(c) GDCP \cite{Peng2018} \\
				\includegraphics[width=0.9\linewidth]{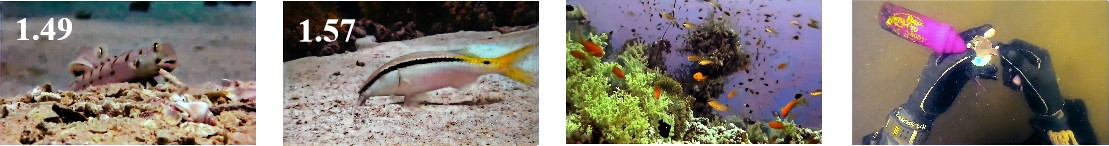}\\
				(d) Guo \etal~\cite{Guo2019} \\
				\includegraphics[width=0.9\linewidth]{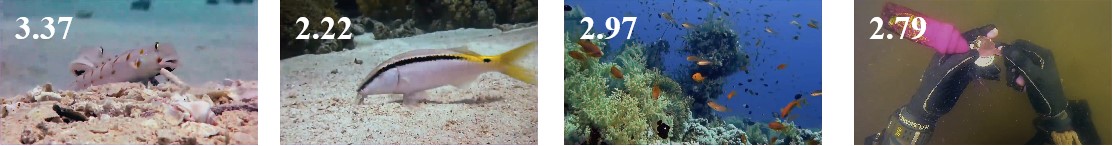}\\
				(e) WaterNet \cite{Libenchmark} \\
				\includegraphics[width=0.9\linewidth]{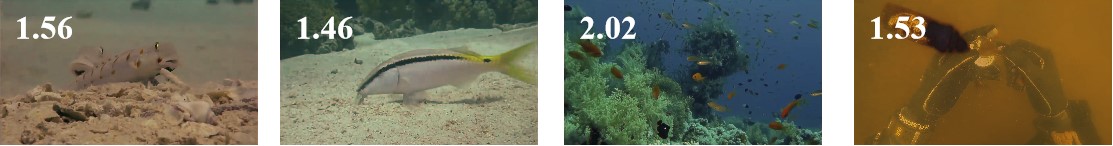}\\
				(f) UWCNN\_typeII \cite{UWCNN} \\
				\includegraphics[width=0.9\linewidth]{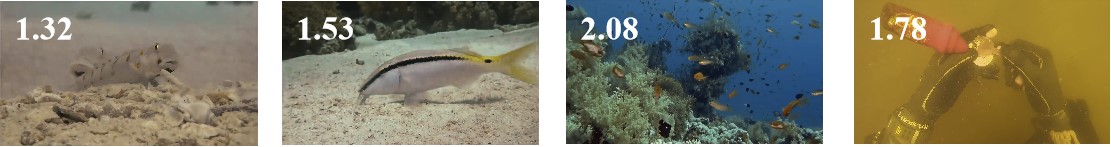}\\
				(g) UWCNN\_retrain \cite{UWCNN} \\
				\includegraphics[width=0.9\linewidth]{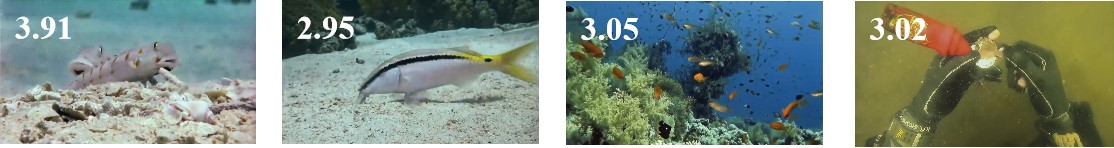}\\	
				(h) Ucolor \\		
			\end{tabular}
		\end{center}
		\caption{Visual comparisons on  challenging underwater images sampled from \textbf{Test-C60}. The number on the top-left corner of each image refers to its perceptual score (the larger, the better).}
		\label{fig:challenging}
	\end{figure*}
	
	\begin{figure*}
		\begin{center}
			\begin{tabular}{c@{ }}
				\includegraphics[height=2.4cm,width=16cm]{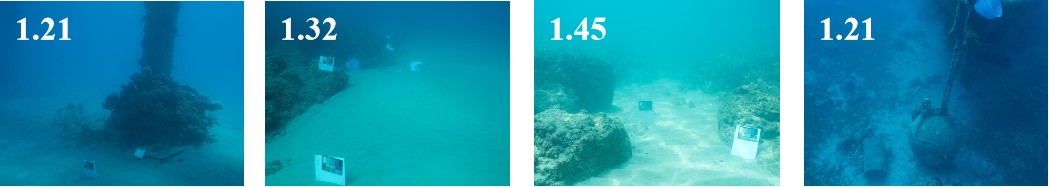} \\
				(a) input \\
				\includegraphics[height=2.4cm,width=16cm]{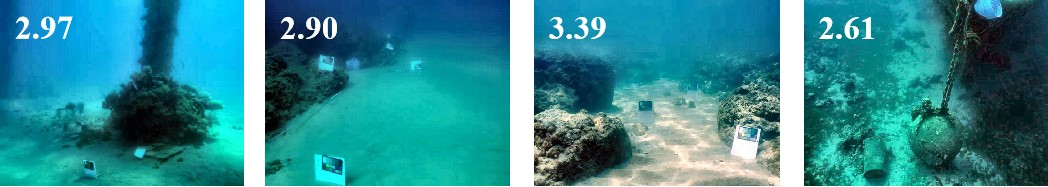} \\
				(b)  Ancuti \etal~\cite{Ancuti2012} \\
				\includegraphics[height=2.4cm,width=16cm]{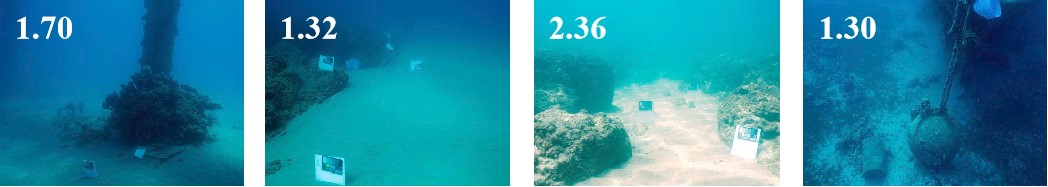}\\
				(c)  Peng \etal~\cite{Peng2017} \\
				\includegraphics[height=2.4cm,width=16cm]{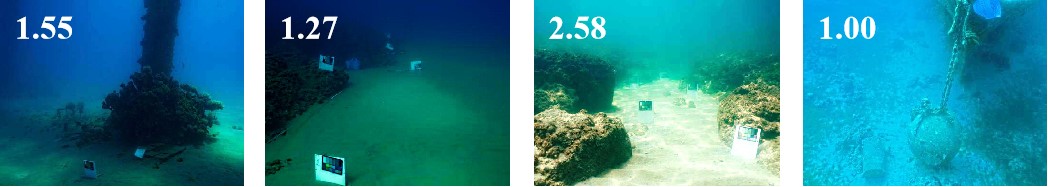}\\
				(d) GDCP \cite{Peng2018} \\
				\includegraphics[height=2.4cm,width=16cm]{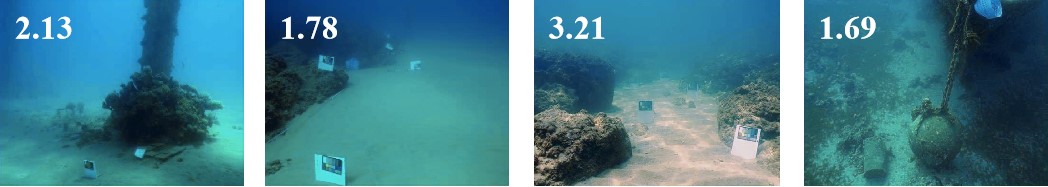}\\
				(e) Water-Net \cite{Libenchmark} \\                     
				\includegraphics[height=2.4cm,width=16cm]{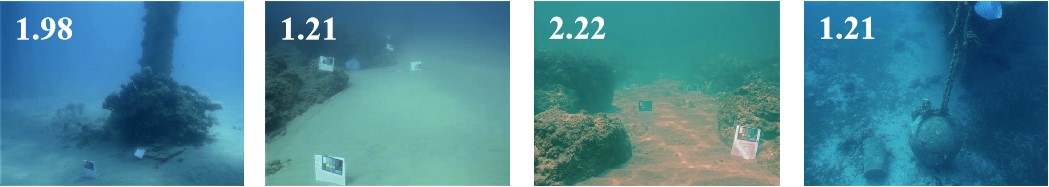}\\
				(f) UWCNN\_typeII \cite{UWCNN} \\
				\includegraphics[height=2.4cm,width=16cm]{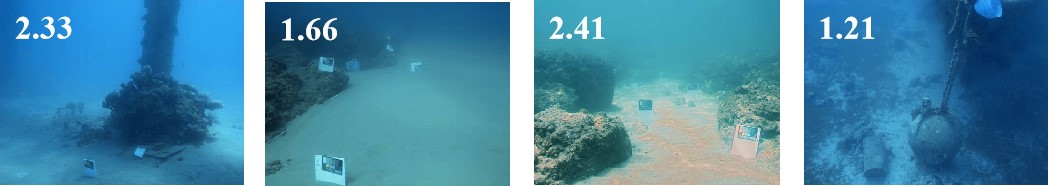}\\
				(g) UWCNN\_retrain \cite{UWCNN} \\
				\includegraphics[height=2.4cm,width=16cm]{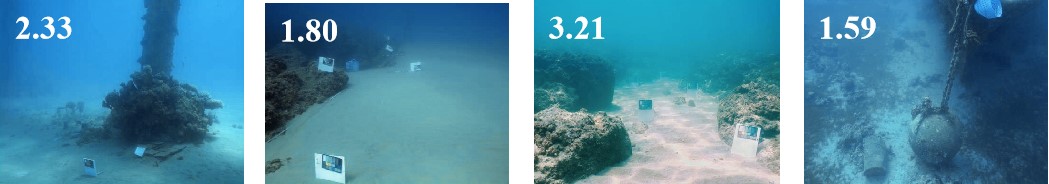}\\	
				(h) Ucolor \\		
			\end{tabular}
		\end{center}
		\caption{Visual comparisons on  challenging underwater images sampled from \textbf{SQUID}. From left to right, the images were taken from four different dive sites Katzaa, Michmoret, Nachsholim, and Satil. The number on the top-left corner of each image refers to its perceptual score  (the larger, the better).}
		\label{fig:SQUID}
	\end{figure*}
	
	\begin{figure*}[!]
		\begin{center}
			\begin{tabular}{c@{ }c@{ }c@{ }c@{ }}
				\includegraphics[height=3cm,width=4cm]{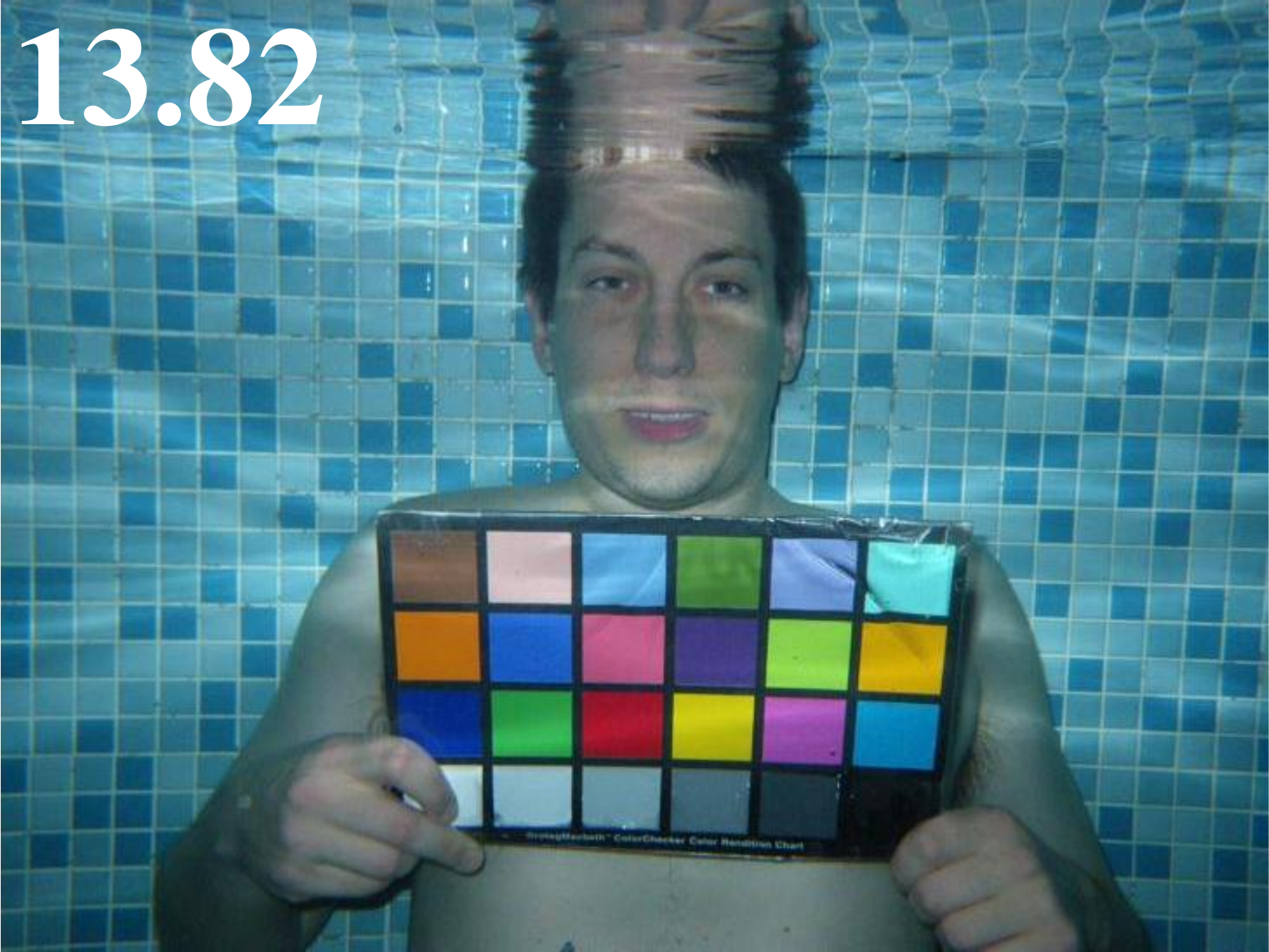} &
				\includegraphics[height=3cm,width=4cm]{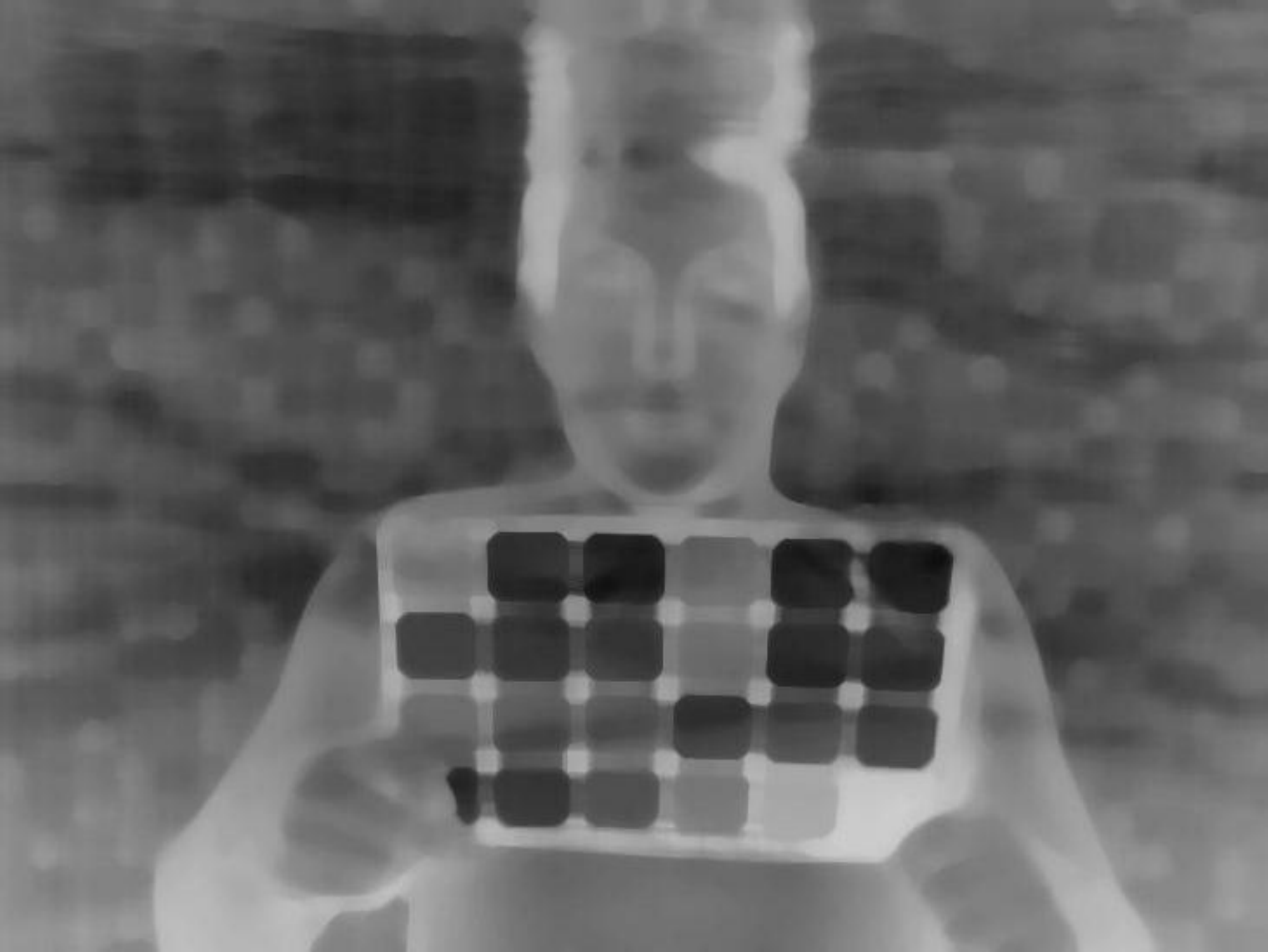} &
				\includegraphics[height=3cm,width=4cm]{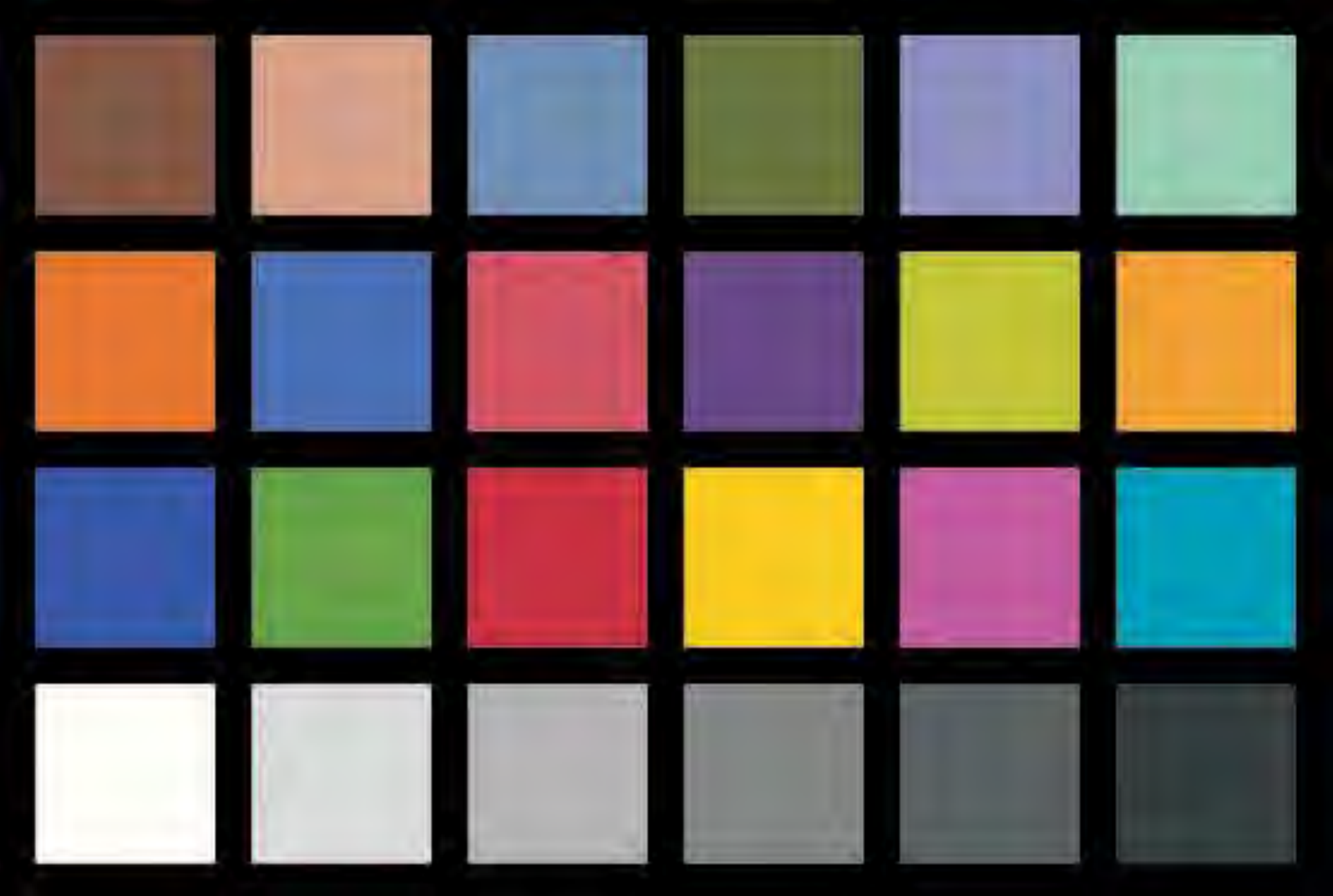}&
				\includegraphics[height=3cm,width=4cm]{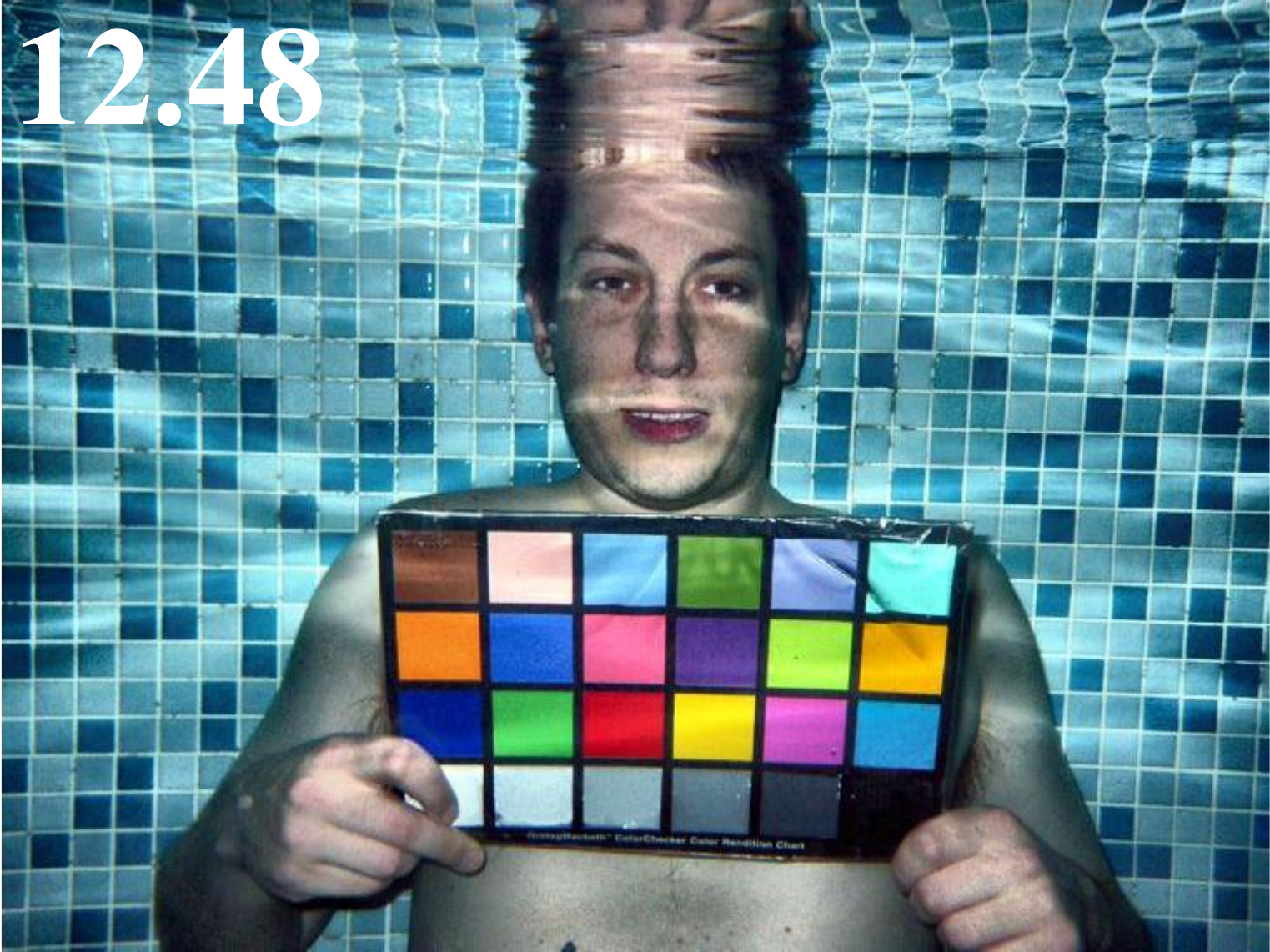}\\
				(a) input & (b) RMT map &(c) Reference & (d) Ancuti \etal~\cite{Ancuti2012} \\
				\includegraphics[height=3cm,width=4cm]{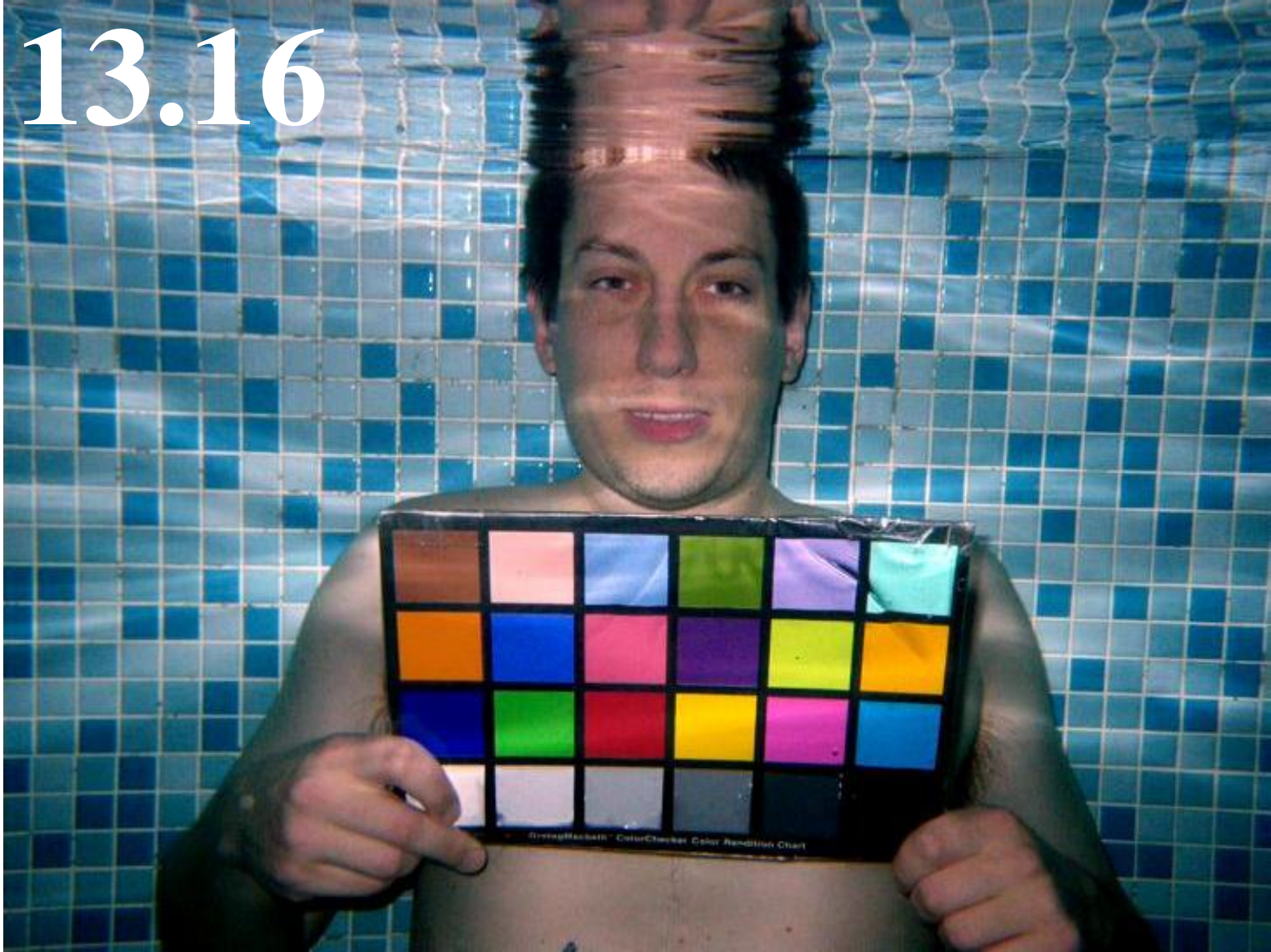}&
				\includegraphics[height=3cm,width=4cm]{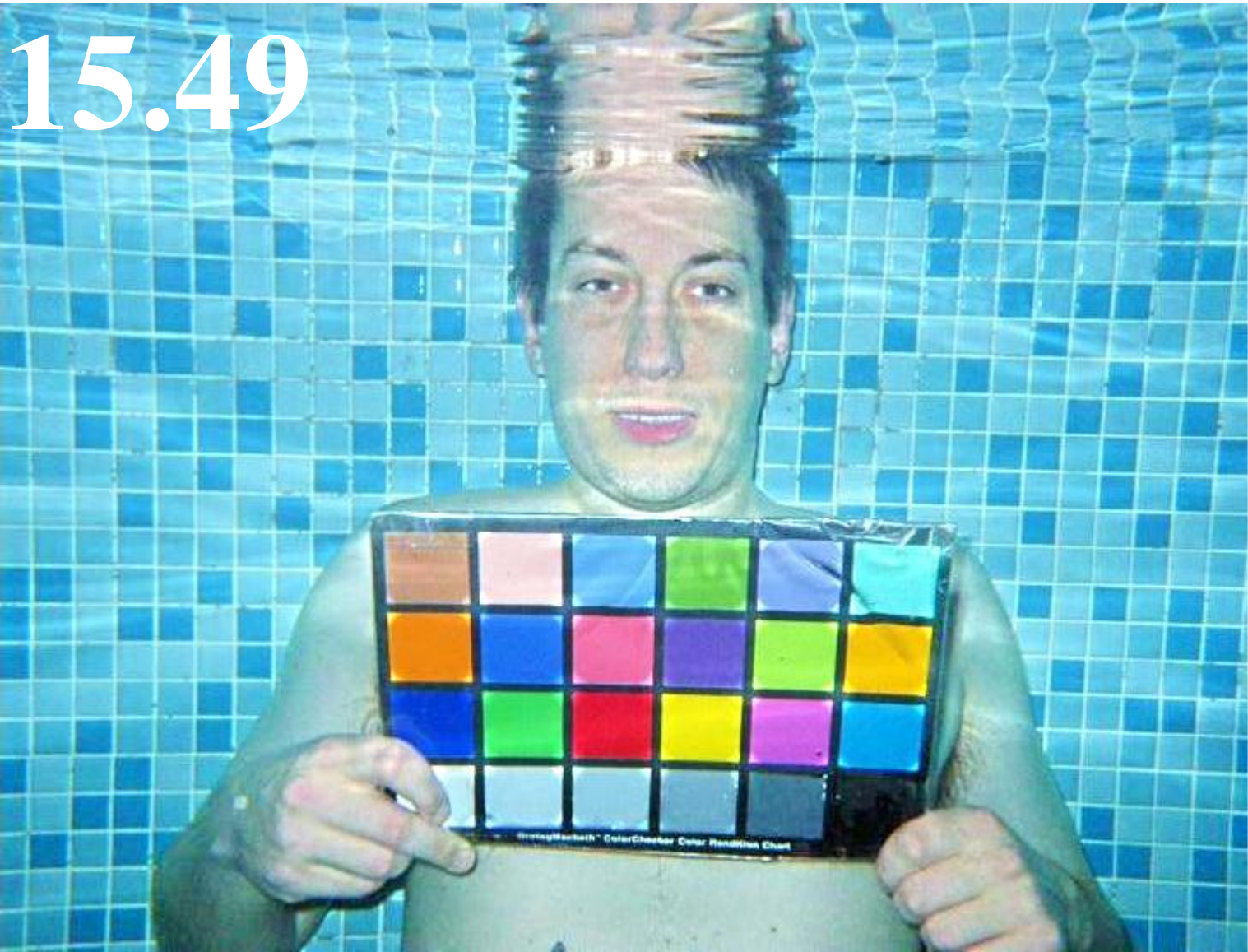}&
				\includegraphics[height=3cm,width=4cm]{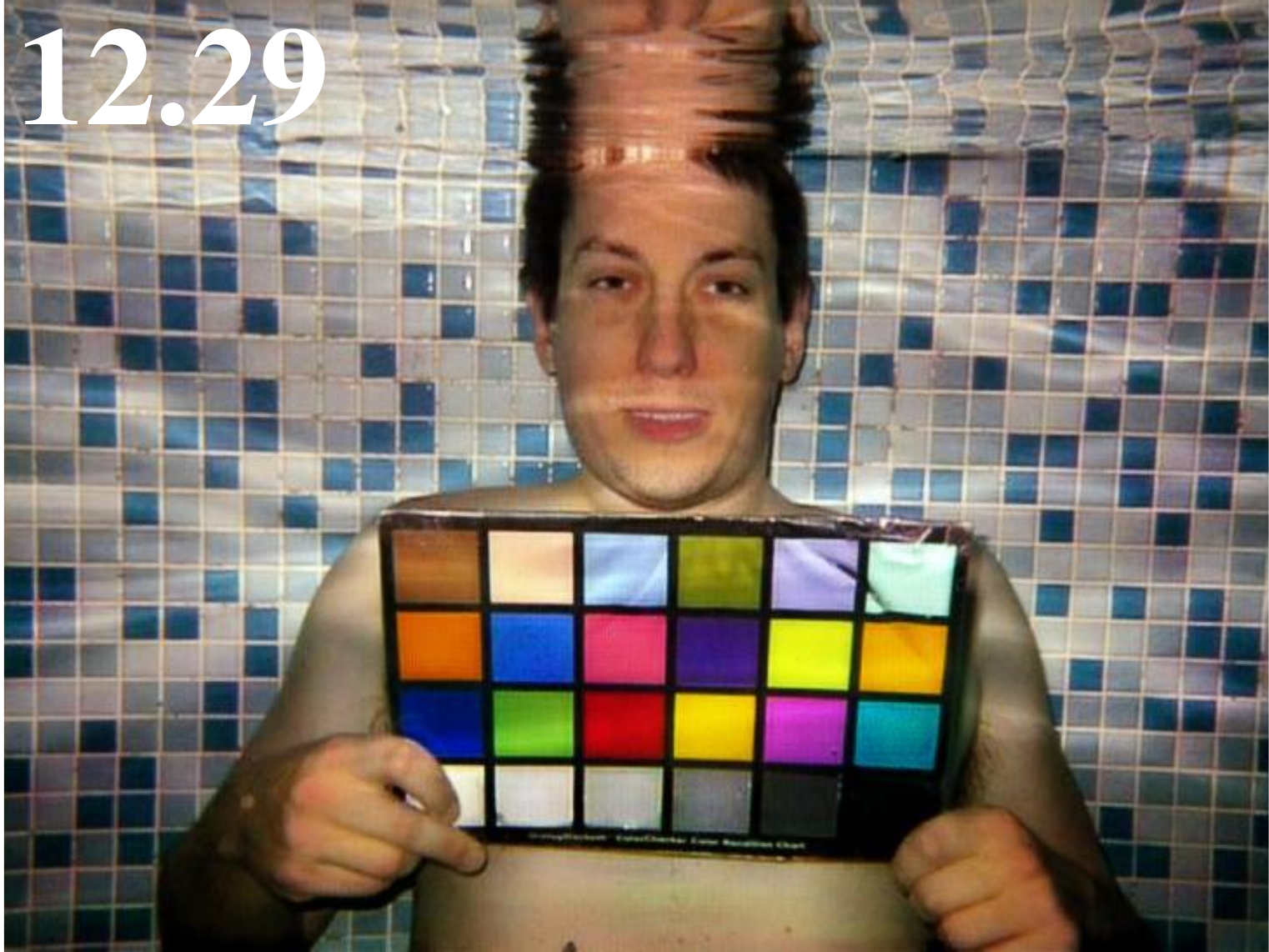}&
				\includegraphics[height=3cm,width=4cm]{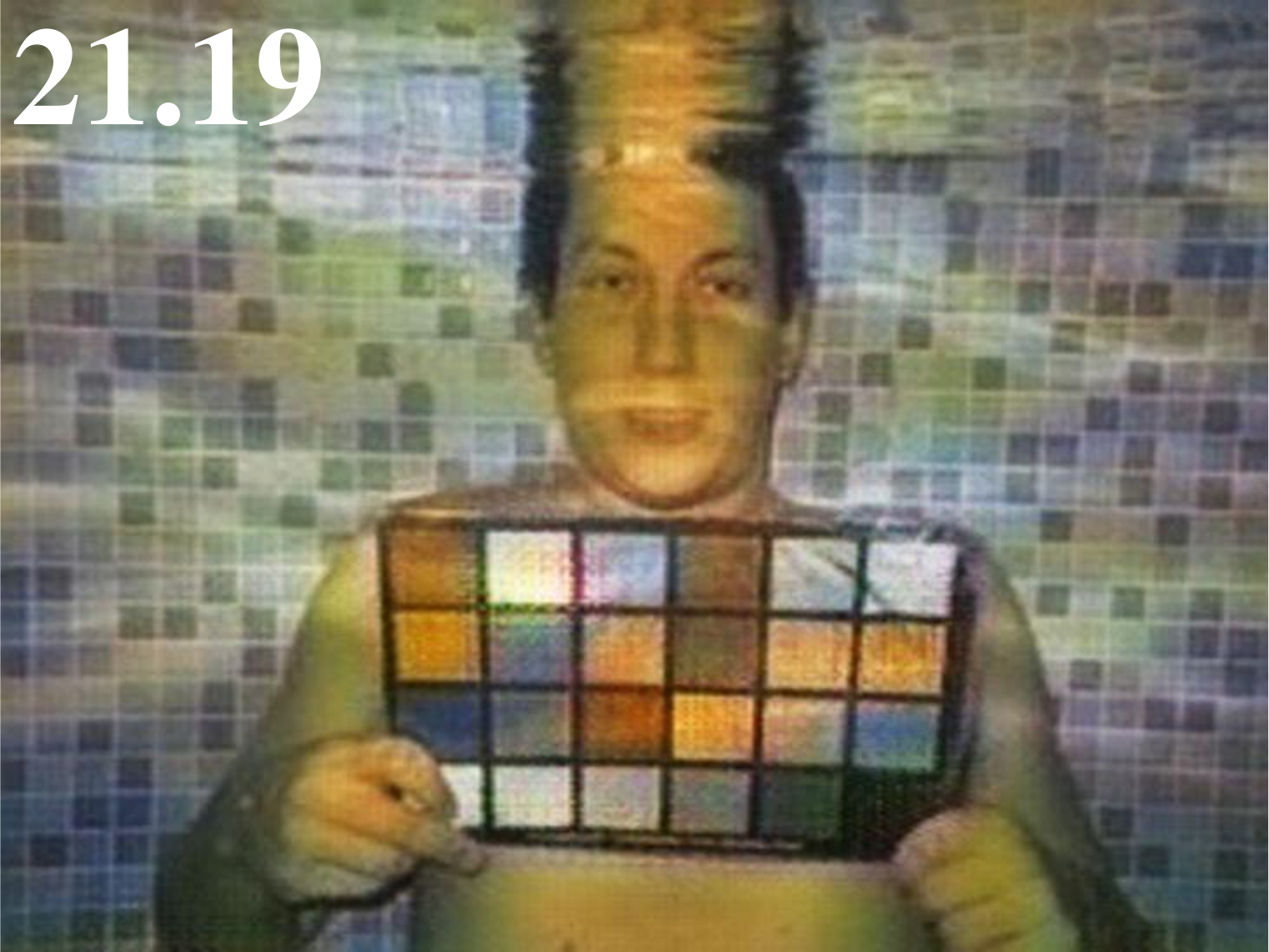}\\
				(e)  Peng \etal~\cite{Peng2017} & (f) GDCP \cite{Peng2018} &	(g) Guo \etal~\cite{Guo2019} & (h) UcycleGAN \cite{UCycleGAN} \\
				\includegraphics[height=3cm,width=4cm]{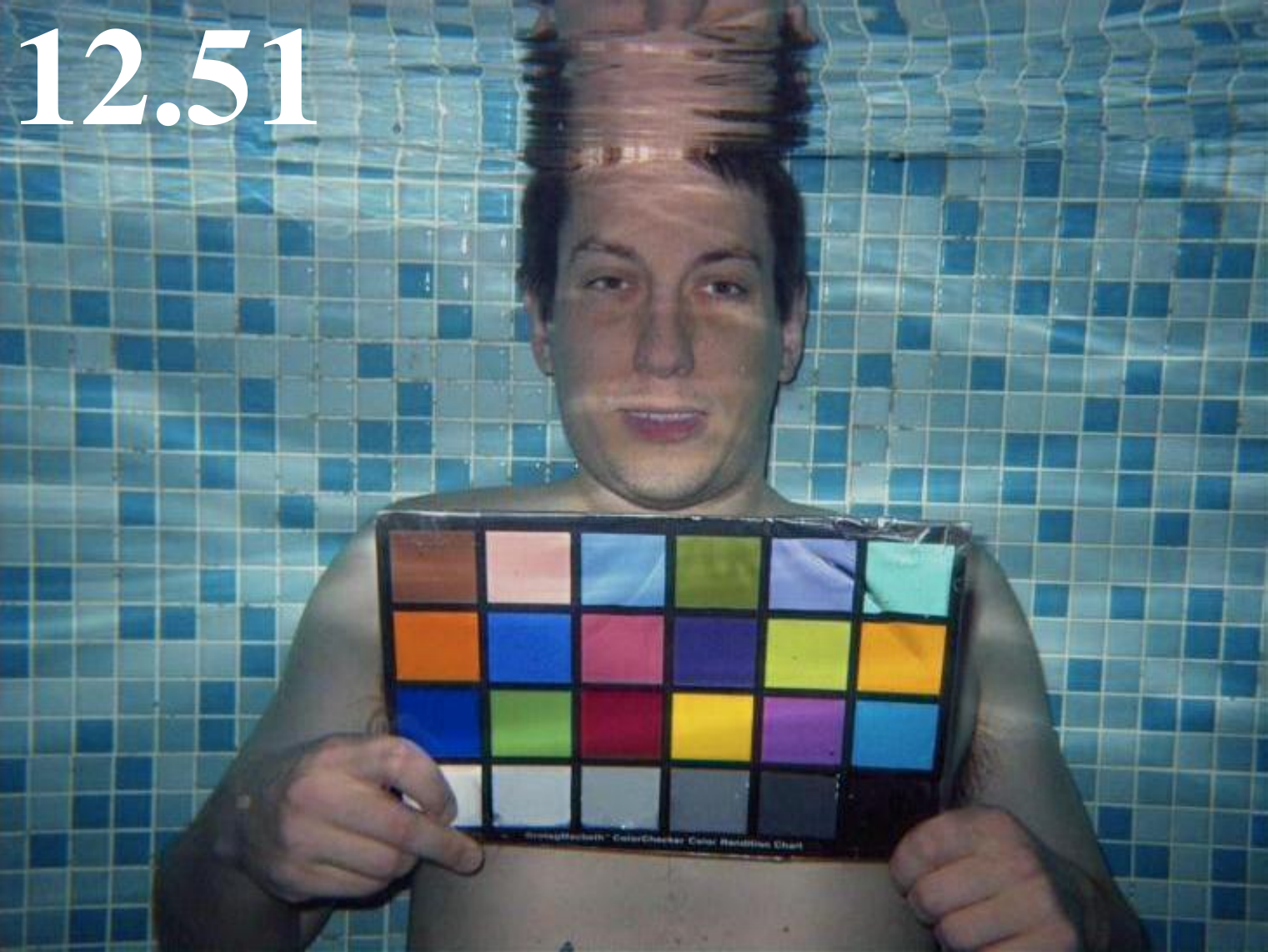}&
				\includegraphics[height=3cm,width=4cm]{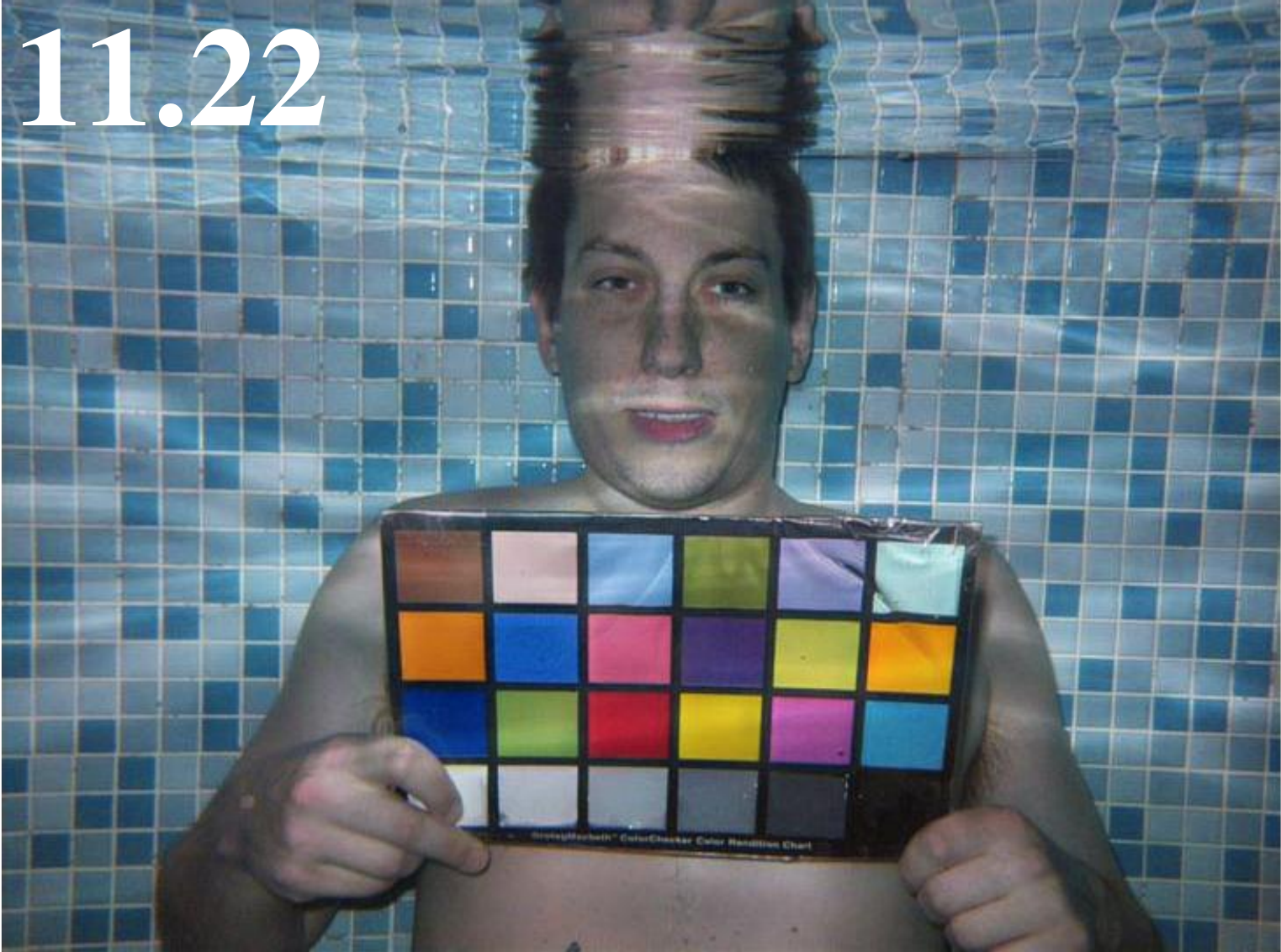}&
				\includegraphics[height=3cm,width=4cm]{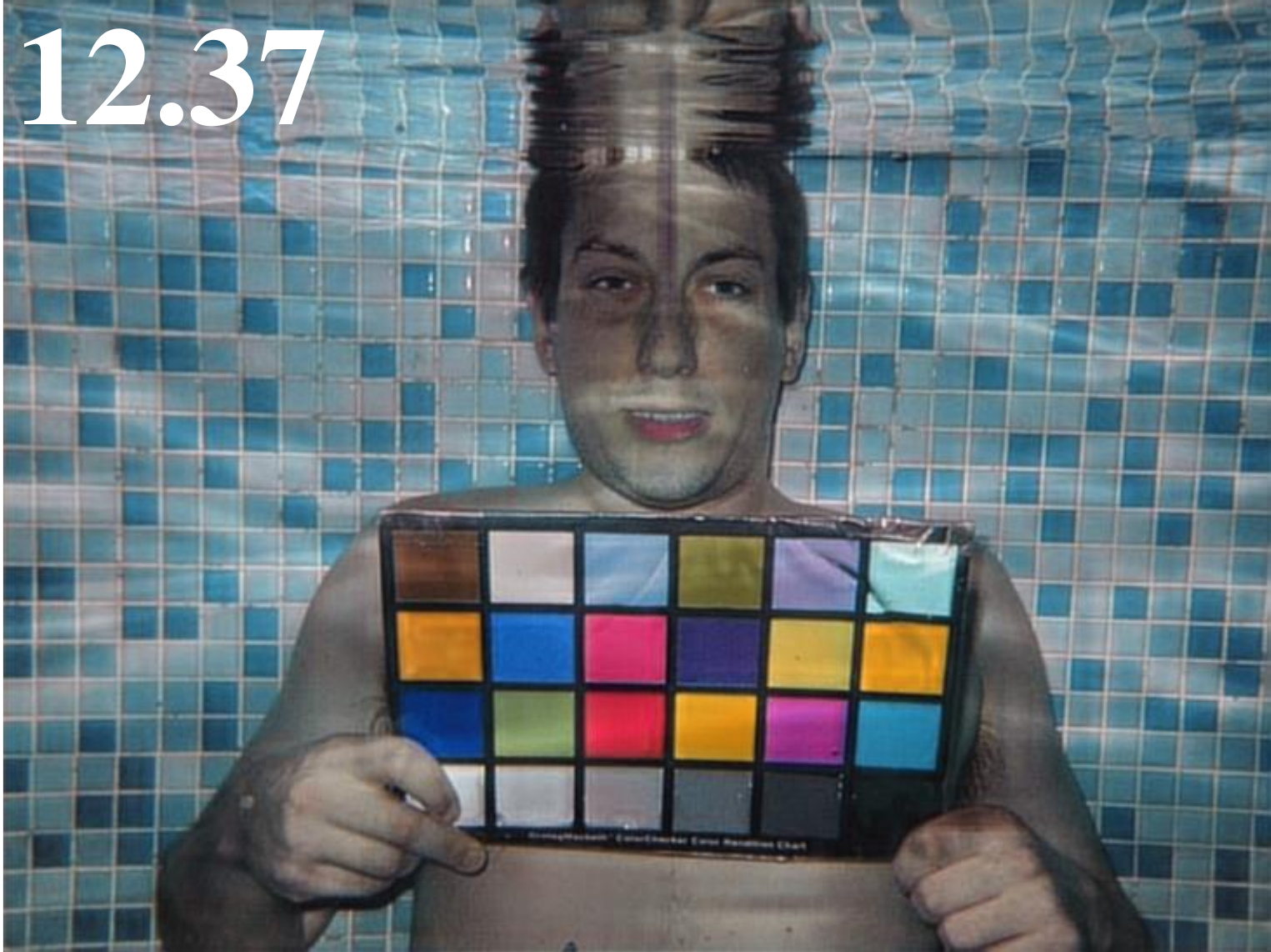}&
				\includegraphics[height=3cm,width=4cm]{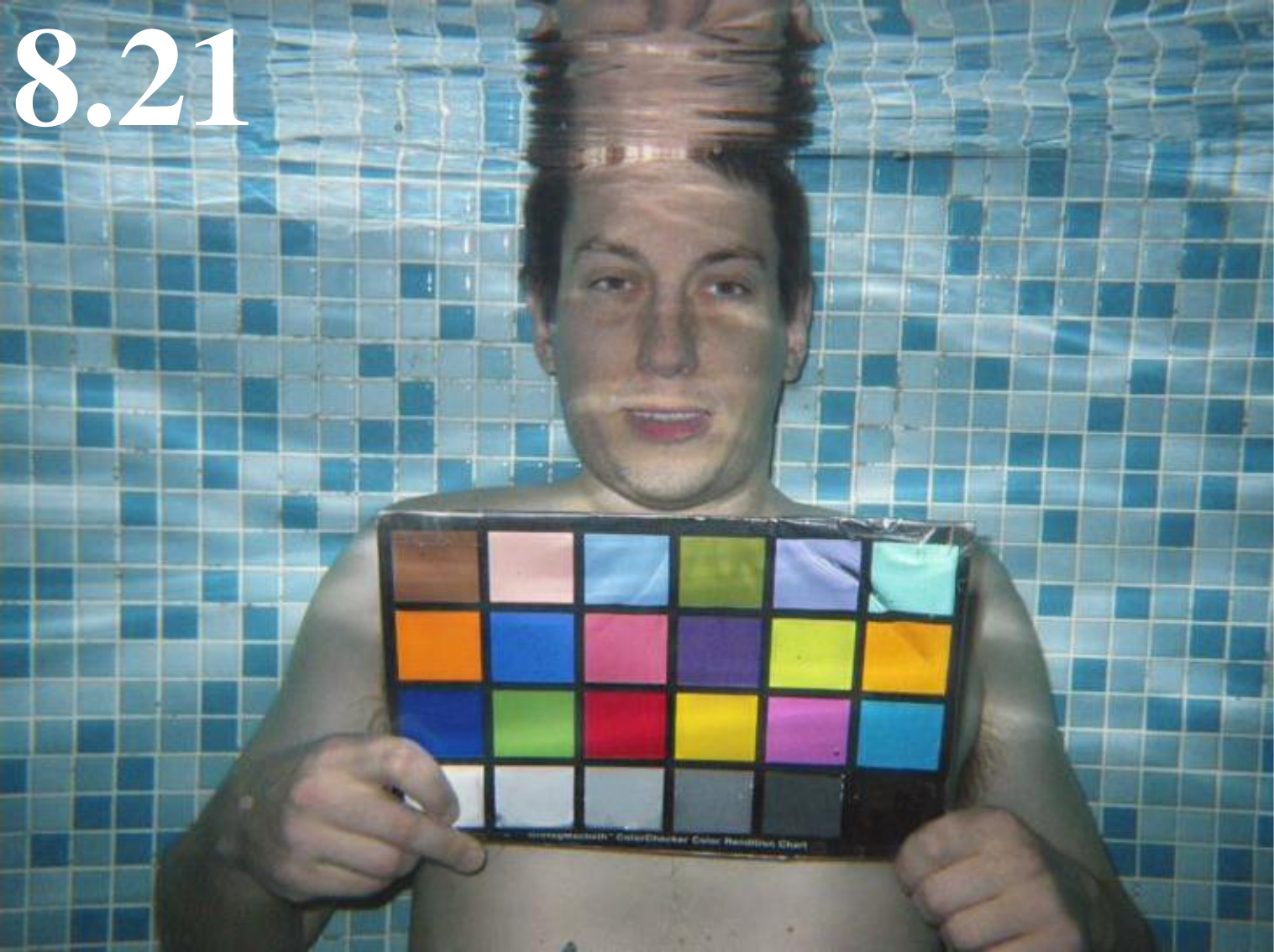}\\
				(i) Water-Net \cite{Libenchmark} & (j) Unet-U \cite{Unet} & (k)  Unet-RMT \cite{Unet} & (l) Ucolor\\
			\end{tabular}
		\end{center}
		\caption{Visual comparisons on a Color Checker image taken by a Pentax W60 camera sampled from \textbf{Color-Checker7}. The values of CIEDE2000 metric for the regions of Color Checker are reported on the top-left corner of the images (the smaller, the better).}
		\label{fig:color}
	\end{figure*}

	\noindent
	\textbf{Evaluation Metrics.}
	For Test-R90 and Test-S1000, we conducted full-reference evaluations using PSNR and MSE metrics. Following \cite{Libenchmark}, we treated the reference images of Test-R90 as the ground-truth images to compute the PSNR and MSE scores.  A higher PSNR or a lower MSE score denotes that the result is closer to the reference image in terms of image content.
	
	For Test-C60 and SQUID that do not have corresponding ground truth images, we employ the no-reference evaluation metrics UCIQE \cite{UCIQE} and UIQM \cite{UIQM} to measure the performance of different methods. A higher UCIQE or UIQM score suggests a better human visual perception. Please note that the scores of UCIQE and UIQM cannot accurately reflect the performance of underwater image enhancement methods in some cases. We refer the readers to  \cite{Libenchmark} and \cite{UnderwaterPami2020} for the discussions and visual examples. In our study, we only provide the scores of UCIQE and UIQM as the reference for the following research. In addition, we also provide the scores of NIQE \cite{NIQE} of different methods as the reference though it was not originally devised for underwater images. A lower NIQE score suggests  better image quality.
	Although SQUID provides a script to evaluate color reproduction and transmission map estimation for the underwater image, this script requires the estimated transmission map or the results having a resolution of 1827$\times$2737. However, the compared deep learning-based methods do not need to estimate the transmission maps. Moreover, our current GPU cannot process the input image with such a high resolution. Besides, the sizes of the color checker in the images of SQUID are too small to be cropped for full-reference evaluations.	Instead, we resized the images in the SQUID testing dataset to a size of 512$\times$512 and processed them by different methods. We conducted a user study to measure the perceptual quality of results on Test-C60 and SQUID.  Specifically, we invited 20 human subjects to  score the perceptual quality of the enhanced images independently. The scores of perceptual quality range from 1 to 5 (worst to best quality). These subjects were trained by observing the results from 1) whether the results introduce color deviations; 2) whether the results contain artifacts; 3) whether the results look natural; and 4) whether the results have good contrast and visibility.
	
	For Color-Check7, we measured the dissimilarity of color between the ground-truth Macbeth Color Checker and the corresponding enhanced results. To be specific, we extracted the color of 24 color patches from the ground-truth Macbeth Color Checker. Then, we respectively cropped the 24 color patches for each enhanced result and computed the average color values of each color patch. At last, we followed the previous method \cite{Ancuti2018} to employ CIEDE2000 \cite{CIEDE2000} to measure the relative perceptual differences between the corresponding color patches of ground-truth Macbeth Color Checker and those of enhanced results. The smaller the CIEDE2000 value, the better.

	\subsection{Visual Comparisons}
	In this section, we conduct visual comparisons on diverse testing datasets. We refer readers to the Google Drive link\footnote{\url{https://drive.google.com/file/d/1zrynw05ZgkVMAybGo9Nhqg2mmIYIMVOT/view?usp=sharing}} for all results of different methods (about 7.4 GB).
	
	We first show the comparisons on a synthetic image in  Fig.~\ref{fig:sythetic2}. The competing methods either fail to dehaze the input image or they introduce undesirable color artifacts. All the methods under comparison fail to recover the complete scene structure. Our result (Fig.~\ref{fig:sythetic2}(l)) is closest to the ground-truth images and obtains the best PSNR/MSE scores. Furthermore, the RMT map shown in Fig.~\ref{fig:sythetic2}(b) indicates that the high degradation regions have large pixel values. With the RMT map, our method highlights these regions and enhances them well.

	We then show the results of different methods on a real underwater image with obvious greenish color deviation in Fig.~\ref{fig:real_data}. In Fig. \ref{fig:real_data}(a), the greenish color deviation significantly hides the structural details of the underwater scene. In terms of color,  GDCP \cite{Peng2018}, UcycleGAN \cite{UCycleGAN}, UWCNN\_typeII \cite{UWCNN} and UWCNN\_retrain \cite{UWCNN} introduce extra color artifacts. All the compared methods under-enhance the image or introduce the over-saturation. In comparison, our Ucolor effectively removes the greenish tone and improves the contrast without obvious over-enhancement and over-saturation.Although UWCNN\_retrain \cite{UWCNN}, Unet-U \cite{Unet}, and Unet-RMT \cite{Unet} were trained with the same data as our Ucolor, their performance is not as good as our Ucolor, which demonstrates the advantage of our specially designed network structure for underwater image enhancement.
	
We also show  comparisons  on challenging underwater images sampled from Test-C60 in Fig.~\ref{fig:challenging}. These underwater images suffer from high backscattering and color deviations as shown in Fig.~\ref{fig:challenging}(a).
	For these images, all competing methods cannot achieve satisfactory results. Some of them even introduce artifacts, such as  GDCP \cite{Peng2018},  Guo \etal~\cite{Guo2019}, and UWCNN\_typeII \cite{UWCNN}. Additionally, some methods introduce artificial colors. For example, the red object in the fourth image becomes purple and the sand in the first two images becomes reddish.   In contrast, our Ucolor not only recovers the relatively realistic color but also enhances details, which is credited to the effective designs of multiple color spaces embedding and the introduction of medium transmission-guided decoder structure. The perceptual scores of our results also suggest the visually pleasing quality of our results.
	
	We present the results of different methods on challenging underwater images sampled from SQUID in Fig.~\ref{fig:SQUID}. As presented, the input underwater images challenge all underwater image enhancement methods. Ancuti \etal~\cite{Ancuti2012} achieves better contrast than the other methods but produces color deviations, \eg, the reddish tone in the third image and the greenish tone in the fourth image. Our Ucolor dehazes the input image, thus improving the contrast of input images. Moreover, our method does not produce obvious artificial colors on the third image. According to the quantitative scores, we can find that the artificial colors significantly affect the perceptual scores given by subjects.

	\begin{table}[!]
		\caption{The evaluations of different methods on  \textbf{Test-S1000}  and \textbf{Test-R90} in terms of average PSNR (dB) and MSE ($\times10^{3}$) values. The best result is in red under each case.}
		\begin{center}
			\begin{tabular}{c||c|c|c|c}
				\hline
				\multirow{2}{*}{} & \multicolumn{2}{c|}{\textbf{Test-S1000}} & \multicolumn{2}{c}{\textbf{Test-R90}} \\
				\cline{2-5}
				Methods  & PSNR$\uparrow$ & MSE$\downarrow$ & PSNR$\uparrow$ & MSE$\downarrow$ \\
				\hline\hline
				input &12.96  &4.60 &16.11   &2.03    \\
				Ancuti \etal~\cite{Ancuti2012} & 13.27 & 5.15 & 19.19  & 0.78  \\
				Li \etal~\cite{Li2016} & 14.29  & 3.64  &  16.73 &1.38  \\
				Peng \etal~\cite{Peng2017} &  13.04&   4.53 & 15.77 &1.72  \\
				GDCP~\cite{Peng2018}  &11.67  &5.98  &13.85   &3.40  \\
				Guo \etal~\cite{Guo2019} &{15.78}  &2.57  &18.05   &1.18   \\
				UcycleGAN~\cite{UCycleGAN} &14.73 &3.13  &  16.61   &1.65 \\
				Water-Net~\cite{Libenchmark} &15.47 &3.26 & 19.81   &1.02   \\
				UWCNN$\_$type1~\cite{UWCNN} &16.27 &2.68 &13.62 &3.52 \\
				UWCNN$\_$type3~\cite{UWCNN} &15.70 &2.87 &12.84 &4.23 \\			
				UWCNN$\_$type5~\cite{UWCNN} &14.78 &2.94 &13.26 &3.65 \\			
				UWCNN$\_$type7~\cite{UWCNN} &12.38 &4.35 &13.02 &3.67 \\			
				UWCNN$\_$type9~\cite{UWCNN} &12.83 &3.85 &12.79 &3.89 \\
				UWCNN$\_$typeI~\cite{UWCNN} &10.44 &6.42 &10.57 &6.24\\
				UWCNN$\_$typeII~\cite{UWCNN} &17.51 &2.59 & 14.75&2.57 \\
				UWCNN$\_$typeIII~\cite{UWCNN} &17.41 &2.39 &13.26 &3.40 \\
				UWCNN$\_$retrain~\cite{UWCNN} &15.87 &2.74 &16.69 &1.71 \\			
				Unet-U~\cite{Unet} &19.14  &1.22 &18.14 &1.32 \\
				Unet-RMT~\cite{Unet} &17.93 &1.43 &16.89 &1.71\\
				Ucolor &{\color{red}{23.05}}  &{\color{red}{0.50}} & {\color{red}{20.63}}   &{\color{red}{0.77}}   \\
				\hline
			\end{tabular}
		\end{center}
		\label{tab1}
	\end{table}
	
	To analyze the robustness and accuracy of color correction, we conduct the comparisons on the underwater Color Checker image in Fig.~\ref{fig:color}. As shown in Fig.~\ref{fig:color}(a), the professional underwater camera (Pentax W60) also inevitably introduces various color casts.  Both traditional and learning-based methods change the colors of input from an overall perspective. As indicated by the CIEDE2000 values on the results, our Ucolor achieves the best performance (8.21 under CIEDE2000 metric) in terms of the accuracy of color correction.
	
	All the visual comparisons demonstrate that our Ucolor not only renders visually pleasing results but also generalizes well to different underwater scenes.

	\subsection{Quantitative Comparisons}
	
	We first perform quantitative comparisons on Test-S1000 and Test-R90. The average scores of PSNR and MSE of different methods are reported in Table~\ref{tab1}. As presented in Table~\ref{tab1}, our Ucolor outperforms all competing methods on Test-S1000 and Test-R90. Compared with the second-best performer, our Ucolor achieves the percentage gain of 20$\%$/59$\%$  and 4.1$\%$/1.3$\%$ in terms of PSNR/MSE on Test-S1000 and Test-R90, respectively. There are two interesting findings from the quantitative comparisons.
	1) Although the medium transmission map used in our Ucolor is the same as the traditional GDCP \cite{Peng2018}, the performance is significantly different. Such a result suggests that the performance of underwater image enhancement can be improved by the effective combination of domain knowledge with deep neural networks. 2) The performance of Unet-U \cite{Unet} is better than Unet-RMT \cite{Unet}, which suggests that simple concatenation of input image and its reverse medium transmission map cannot improve the underwater image enhancement performance of deep model, and even decreases the performance. 3) The generalization capability of UWCNN models \cite{UWCNN} is limited because they require the images to be taken in the accurate type of water as inputs. Due to the limited space, we only present the results of the original UWCNN model that performs the best in Table~\ref{tab1} in the following experiments, \ie, UWCNN\_typeII \cite{UWCNN}.

\begin{table*}
	\caption{The average  perceptual scores (PS), UIQM \cite{UIQM} scores, UCIQE \cite{UCIQE} scorse, and NIQE \cite{NIQE} scores of different methods on \textbf{Test-C60} and \textbf{SQUID}. The best result is in red under each case. ``-'' represents the results are not available.}
	\begin{center}
		\begin{tabular}{c||c|c|c|c|c|c|c|c}
			\hline			
			\multirow{2}{*}{} & \multicolumn{4}{c|}{\textbf{Test-C60}} & \multicolumn{4}{c}{\textbf{SQUID}} \\
			\cline{2-9}
			Methods  & PS$\uparrow$ & UIQM$\uparrow$ & UCIQE$\uparrow$ & NIQE$\downarrow$& PS$\uparrow$  & UIQM$\uparrow$ & UCIQE$\uparrow$&  NIQE$\downarrow$\\
			\hline\hline
			input    &1.34 &0.84 &0.48 &7.14 &1.21&0.82& 0.42 & 4.93\\
			Ancuti \etal~\cite{Ancuti2012}     &2.11  &1.22 &0.62 &{\color{red}{4.94}} &{\color{red}{2.93}}  & 1.30 &0.62 &5.01\\
			Li \etal~\cite{Li2016}    &1.22  &{\color{red}{1.27}} &{\color{red}{0.65}} &5.32 &1.00  &{\color{red}{1.34}} &{\color{red}{0.66}} &4.81\\
			Peng \etal~\cite{Peng2017}     &2.07  &1.13 &0.58 &6.01 &2.34  &0.99 &0.50 &4.39\\
			GDCP~\cite{Peng2018}   &1.98  &1.07 &0.56 &5.92 &2.47  &1.11 &0.52 &4.48\\
			Guo \etal~\cite{Guo2019}    &2.63  &1.11 &0.60 &5.71 &-  &- &- &-\\
			UcycleGAN~\cite{UCycleGAN}    &1.01  &0.91 &0.58 &7.67 &1.16  &1.11 &0.56 &5.93\\
			Water-Net~\cite{Libenchmark}    &3.52  &0.97 &0.56 &6.04 &2.78  &1.03 &0.54 &4.72\\
			UWCNN$\_$typeII~\cite{UWCNN}    &2.19   & 0.77 &0.47 &6.76 &2.72  &0.69 &0.44 &4.60\\
			UWCNN$\_$retrain~\cite{UWCNN}    &2.91  &0.84 &0.49 &6.66 & 2.67  &0.77 &0.46 &4.38\\
			Unet-U \cite{Unet}    &3.37  &0.94 &0.50 &6.12& 2.61 &0.82 &0.50 &4.38 \\
			Unet-RMT \cite{Unet}    &3.04  &1.03 &0.52  &6.12 &2.53  & 0.82 & 0.49 &5.16\\
			Ucolor  &{\color{red}{3.74}} &0.88 &0.53 &6.21 &2.82   &0.82 &0.51 & {\color{red}{4.29}}\\
			\hline
		\end{tabular}
	\end{center}
	\label{tab_user}
\end{table*}

	Next, we conduct a user study on Test-C60 and SQUID. The average perceptual scores of the results by different methods are reported in Table \ref{tab_user}, where it can be observed that these two challenging testing datasets fail most underwater image enhancement methods in terms of perceptual quality. Some methods such as Li \etal~\cite{Li2016} and UcycleGAN~\cite{UCycleGAN} even achieve lower perceptual scores than inputs. For the Test-60 testing dataset, the deep learning-based methods achieve relatively higher perceptual scores. Among them, our Ucolor is superior to the other competing methods. For the SQUID testing dataset, the traditional fusion-based method~\cite{Ancuti2012} obtains the highest perceptual score while our Ucolor ranks the second best. Other deep learning-based methods achieve similar perceptual scores. As shown in Fig. \ref{fig:SQUID}, all deep learning-based methods cannot handle the color deviations of images in SQUID well,  while the haze can be removed. In contrast, the fusion-based method~\cite{Ancuti2012} achieves better contrast, thus obtaining a higher perceptual score. Observing the scores of non-reference image quality assessment metrics, we can see that Li \etal~\cite{Li2016} obtains the best performance in terms of UIQM and UCIQE scores. For the NIQE scores, our Ucolor achieves the lowest NIQE score on the SQUID testing set while Ancuti \etal~\cite{Ancuti2012} obtain the best performance  on the Test-C60 testing set.

	\begin{table*}
		\caption{The color dissimilarity comparisons of different methods on \textbf{Color-Check7} in terms of the CIEDE2000. The best result is in red under each case.}
		\begin{center}
			\begin{tabular}{c||c|c|c|c|c|c|c|c}
				\hline
				Methods  & Pen W60 &  Pen W80 &Can D10& Fuj Z33 & Oly T6000 & Oly T8000  & Pan TS1 & Avg \\
				\hline\hline
				input    &13.82  &17.26 &16.13  &16.37  &14.89  &23.14    &19.06 & 17.24\\
				Ancuti \etal~\cite{Ancuti2012}     &12.48  &13.30 &14.28  &11.43  &11.57  &{\color{red}{12.58}}  &{\color{red}{10.63}} &12.32\\
				Li \etal~\cite{Li2016}    &15.41 &17.56 &18.52  &25.01  &16.01  &17.12    &12.03 & 17.38\\
				Peng \etal~\cite{Peng2017}     &13.16 &16.01  &14.78  &14.09  &12.24  &14.79    &19.59 &14.95\\
				GDCP~\cite{Peng2018}   &15.49 &24.32  &16.89  &13.73  &12.76  &16.82    &12.93 &16.13\\
				Guo \etal~\cite{Guo2019}    &12.29  &15.50 &14.58  &16.65  &39.71  &15.14    &12.40 &18.04\\
				UcycleGAN~\cite{UCycleGAN}    &21.19 &21.23  &22.96  &26.28  &20.88  &23.42   &19.02 &22.14\\
				Water-Net~\cite{Libenchmark}    &12.51 &19.57 &15.44  &12.91  &17.55  &21.73    &18.84 &16.94\\
				UWCNN$\_$typeII~\cite{UWCNN}    &16.73 &20.55  &17.73  &17.20  &16.31 & 17.94   &20.97 &18.20 \\
				UWCNN$\_$retrain~\cite{UWCNN}    &13.64 &20.33  &14.91  &13.38  &14.72 &18.11    &20.19 &16.47 \\
				Unet-U \cite{Unet}    &11.22 &15.17  &13.32  &11.91  &10.87 &15.12    &17.31 &13.56 \\
				Unet-RMT \cite{Unet}    &12.37 &19.01  &15.57  &14.80  &13.26 &16.47    &19.55 &15.86 \\
				Ucolor  &{\color{red}{8.21}} &{\color{red}{10.59}}    &{\color{red}{12.27}}  &{\color{red}{8.11}}  &{\color{red}{7.22}}  & 14.42 &14.54 &{\color{red}{10.77}}\\
				\hline
			\end{tabular}
		\end{center}
		\label{tab3}
	\end{table*}

	To demonstrate the robustness to different cameras and the accuracy of color restoration, we report the average CIEDE2000 scores on
	Color-Checker7 in Table~\ref{tab3}. For the cameras of Pentax W60, Pentax W80, Cannon D10, Fuji Z33, and Olympus T6000, our Ucolor obtains the lowest color dissimilarity. Moreover, our Ucolor achieves the best average score across seven cameras. Such results demonstrate the superiority of our method for underwater color correction. It is interesting that some methods achieve worse performance in terms of the average CIEDE2000 score than the original input, which suggests that some competing methods cannot recover the real color and even break the inherent color.

	\begin{figure*}[th]
		\begin{center}
			\begin{tabular}{c@{ }c@{ }c@{ }c@{ }c}
				\includegraphics[height=2.2cm,width=3.5cm]{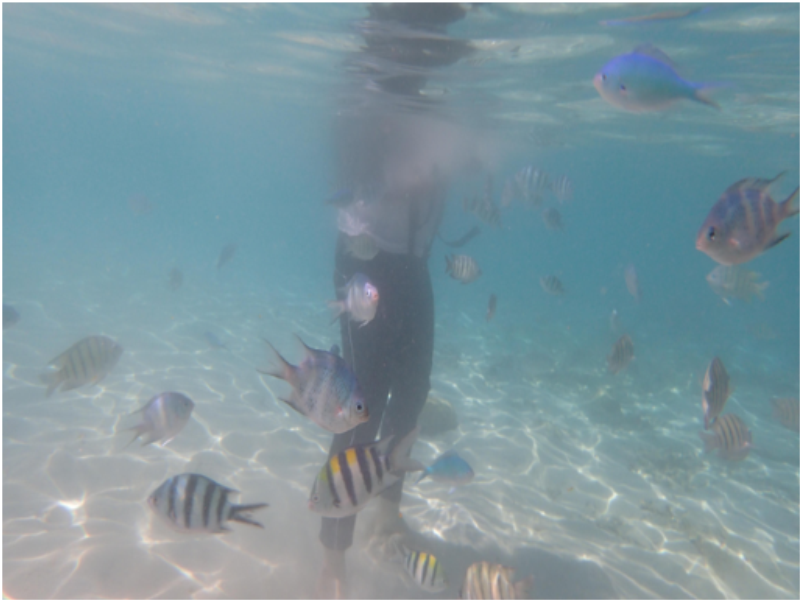} &
				\includegraphics[height=2.2cm,width=3.5cm]{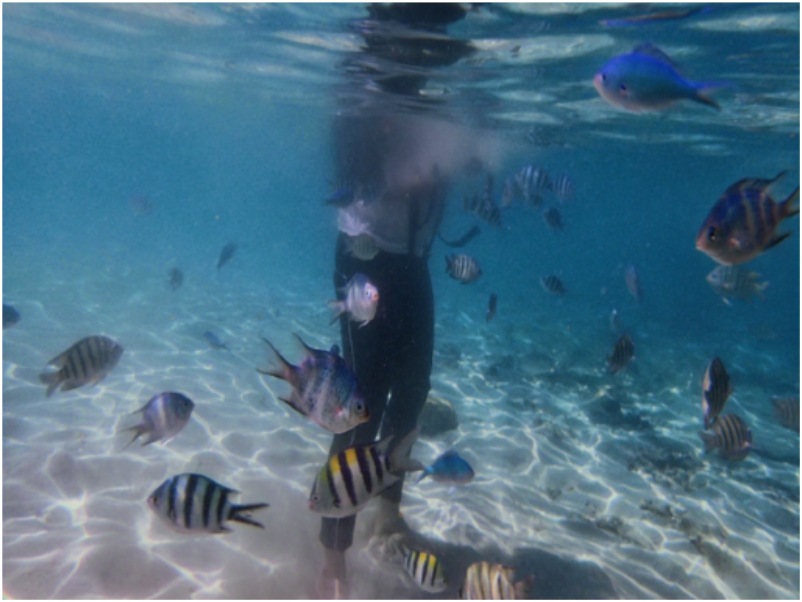} &
				\includegraphics[height=2.2cm,width=3.5cm]{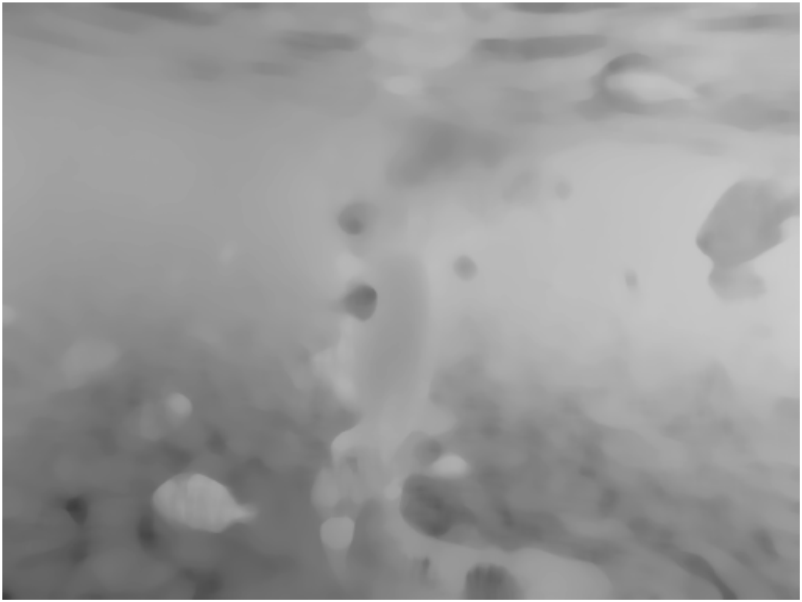}&
				\includegraphics[height=2.2cm,width=3.5cm]{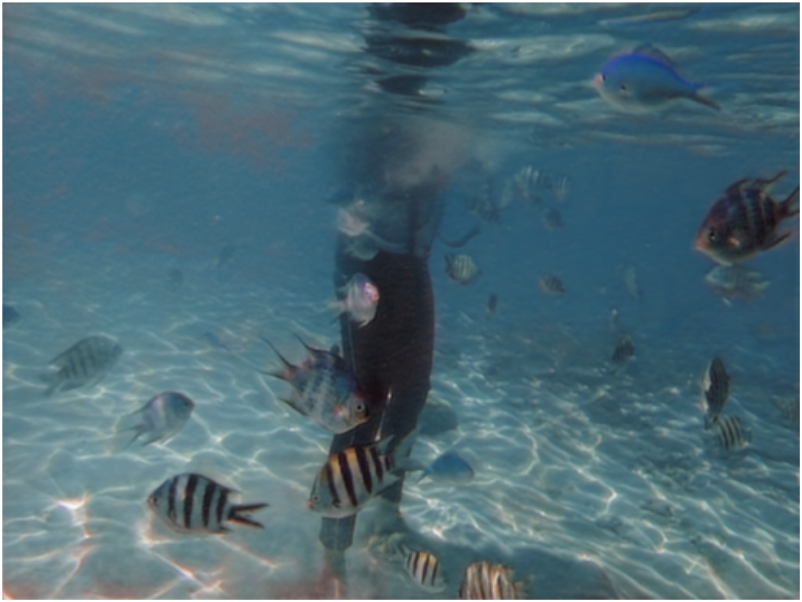}\\
				(a) input & (b) Ucolor&(c)   RMT map & (d) w/ RDCP \\
				\includegraphics[height=2.2cm,width=3.5cm]{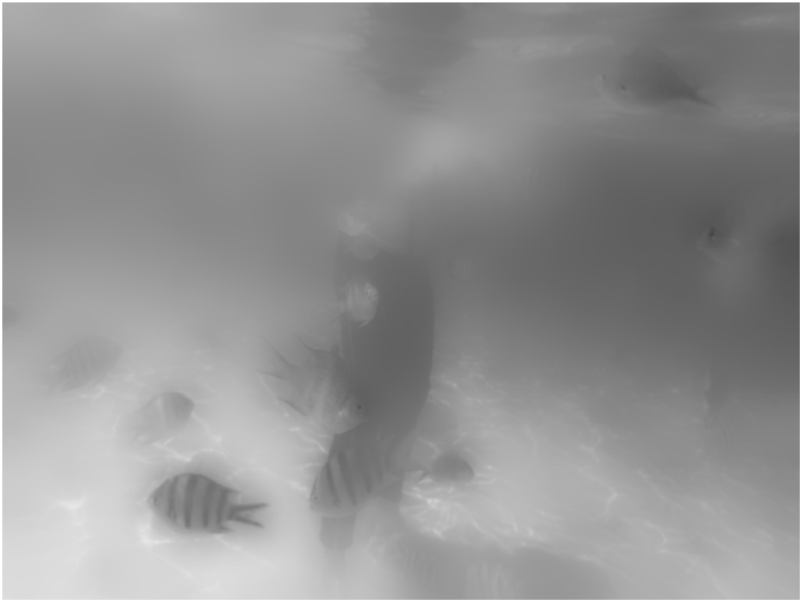}&
				\includegraphics[height=2.2cm,width=3.5cm]{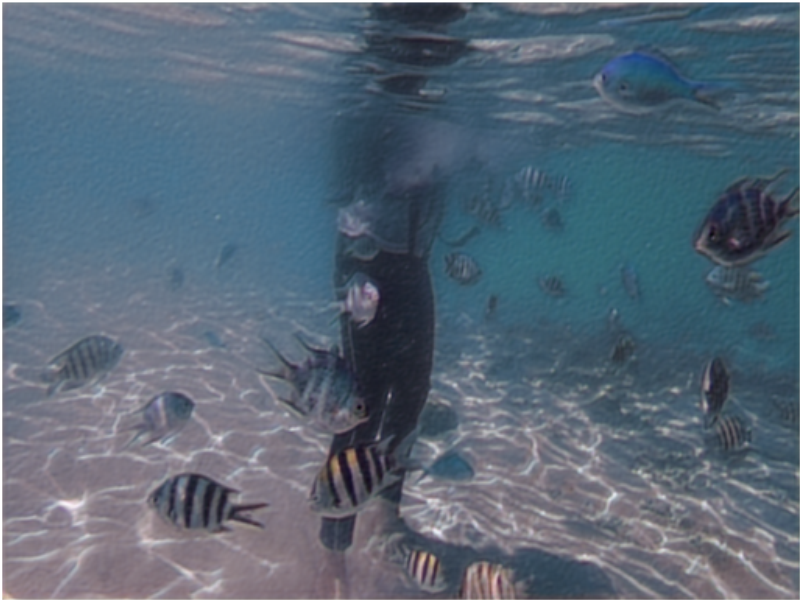}&
				\includegraphics[height=2.2cm,width=3.5cm]{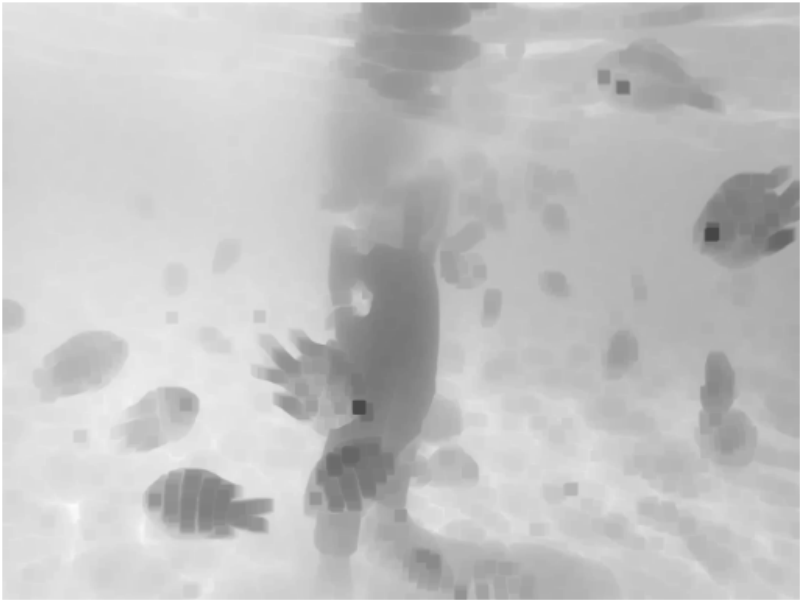} &
				\includegraphics[height=2.2cm,width=3.5cm]{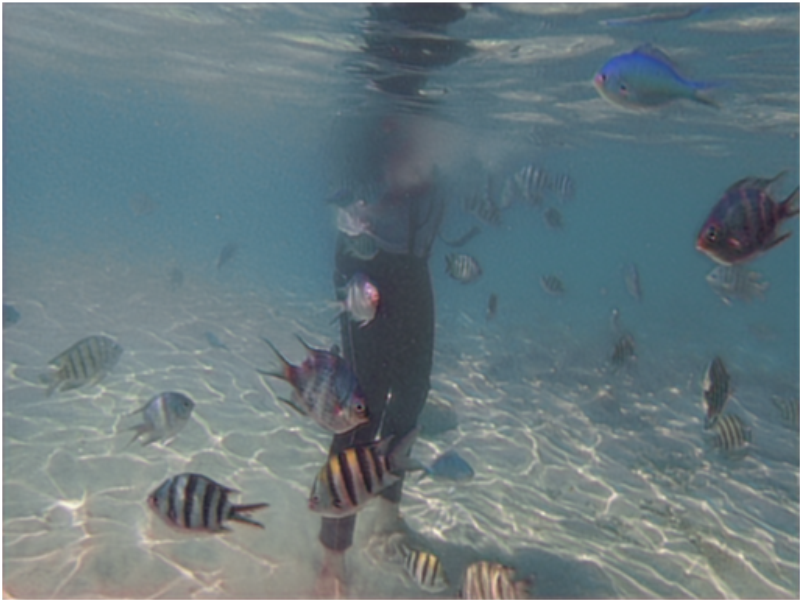} \\
				(e) RDCP map & (f) w/ RUDCP  & (g) RUDCP map & (h) w/o MTGM \\
			\end{tabular}
		\end{center}
		\caption{Ablation study of the effects of the reverse medium transmission map. RDCP map represents the reverse DCP map, where the DCP map was estimated by \cite{He2010}. RUDCP map represents the reverse UDCP map, where the UDCP map was estimated by \cite{Drews2016}. Compared with the RDCP and RUDCP maps, the RMT map can more accurately indicate the degradation of underwater image, thus leading to better enhancement performance of Ucolor.}
		\label{fig:mtgm}
	\end{figure*}
	
	\subsection{Ablation Study}
	We perform extensive ablation studies to analyze the core components of our method, including the multi-color space encoder (MCSE)), the medium transmission guidance module (MTGM), and the channel-attention module (CAM). Additionally, we analyze the combination of the $\ell_{1}$ loss and the perceptual loss. More specifically, 
	\begin{itemize}
		\item  w/o HSV, w/o Lab, and w/o HSV+Lab stand for Ucolor without the HSV, Lab, and both HSV and Lab color spaces encoder paths, respectively. 
		\item  w/ 3-RGB  means that all inputs of three encoder paths are RGB images. 
		\item  w/o MTGM refers to the Ucolor without the medium transmission guidance module. 
		\item w/ RDCP and w/ RUDCP are the models by replacing the medium transmission map estimated via \cite{Peng2018} with the algorithms in \cite{He2010} and \cite{Drews2016}, respectively. 
		\item w/o CAM stands for the Ucolor without the channel-attention module. 
		\item w/o perc loss means that the Ucolor is trained only with the constraint of $\ell_{1}$ loss. 
	\end{itemize}
	
\begin{figure}
		\begin{center}
			\begin{tabular}{c@{ }c@{ }c@{ }}
				\includegraphics[height=2.2cm,width=2.8cm]{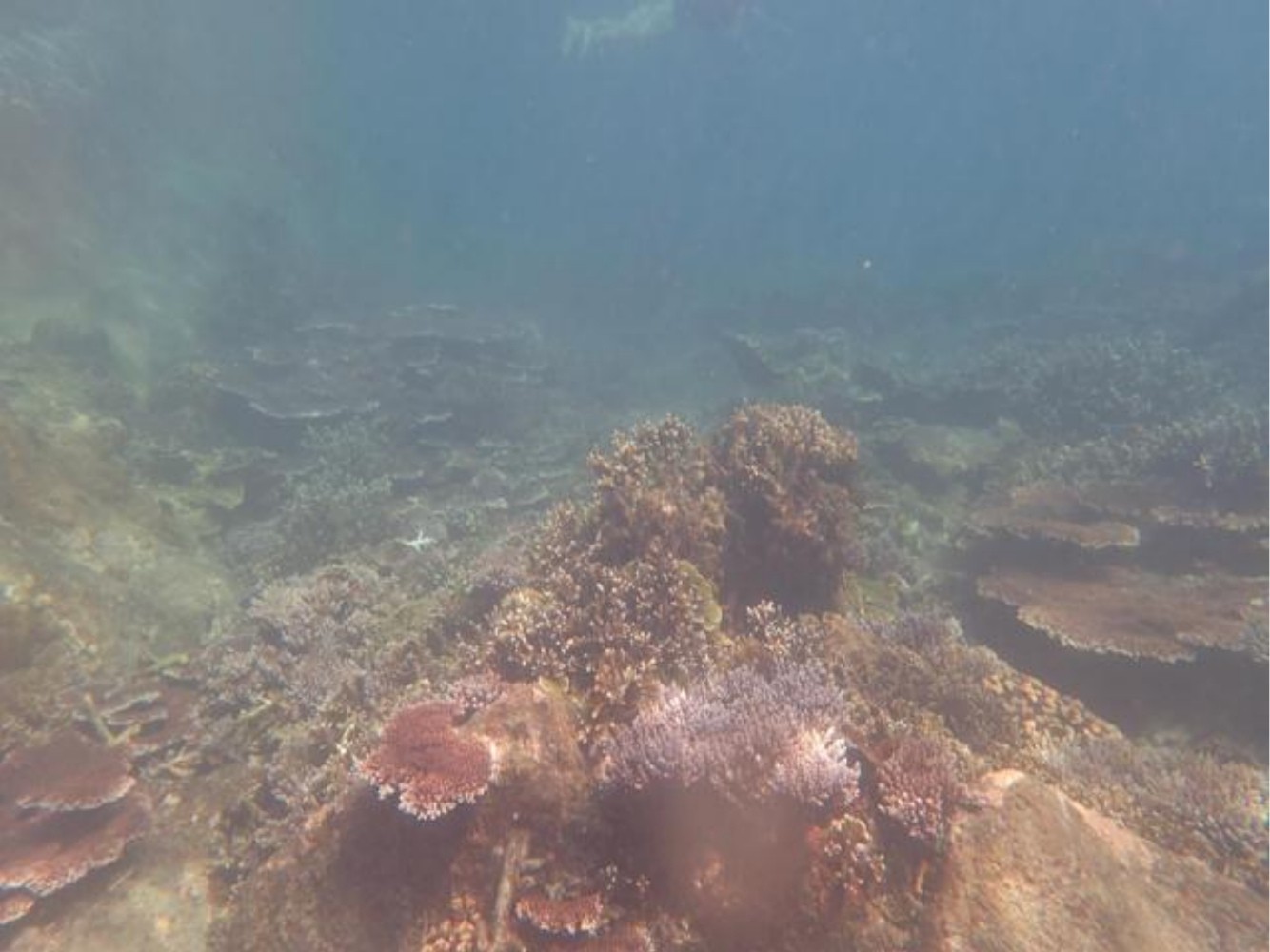} &
				\includegraphics[height=2.2cm,width=2.8cm]{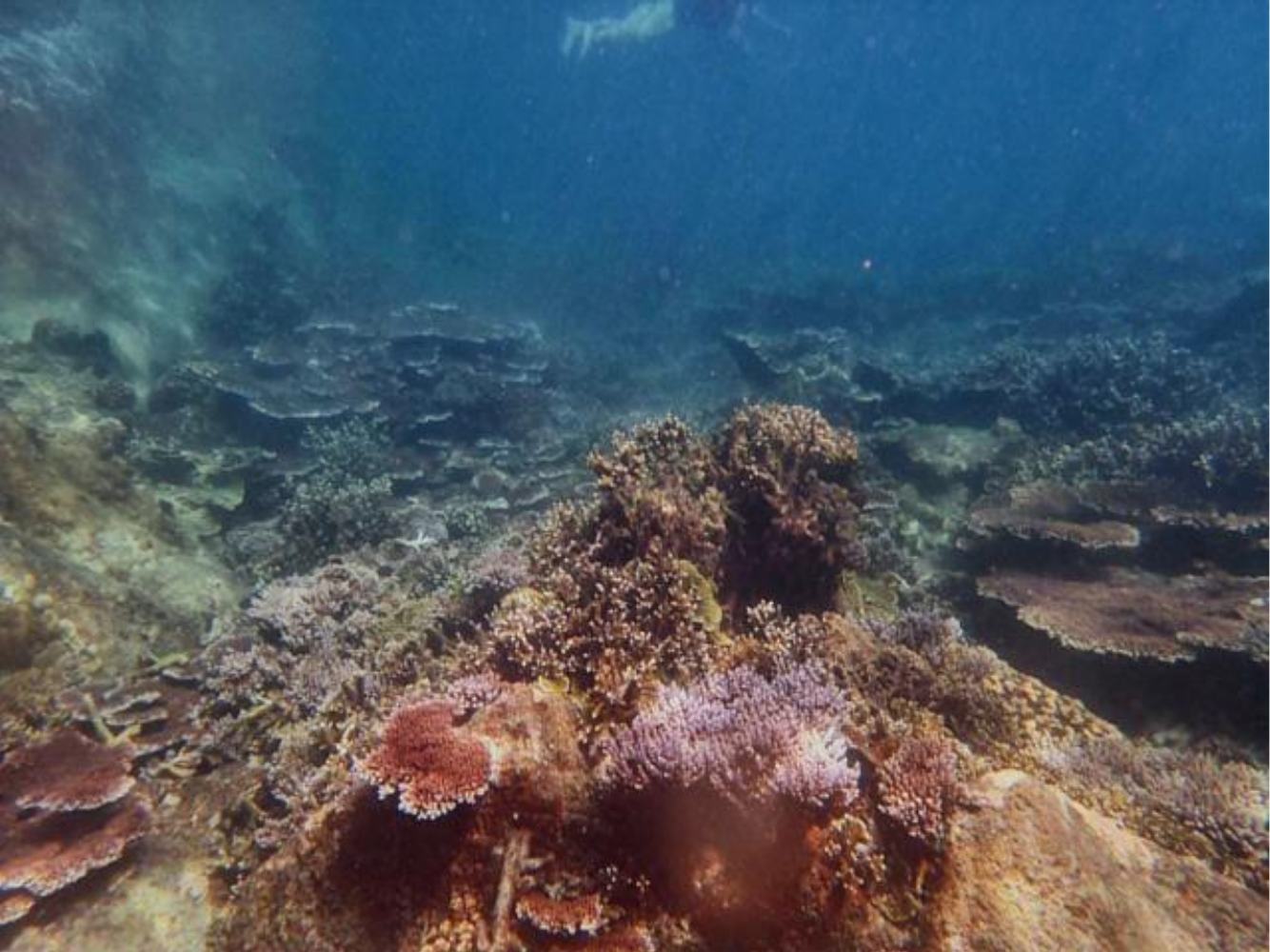} &
				\includegraphics[height=2.2cm,width=2.8cm]{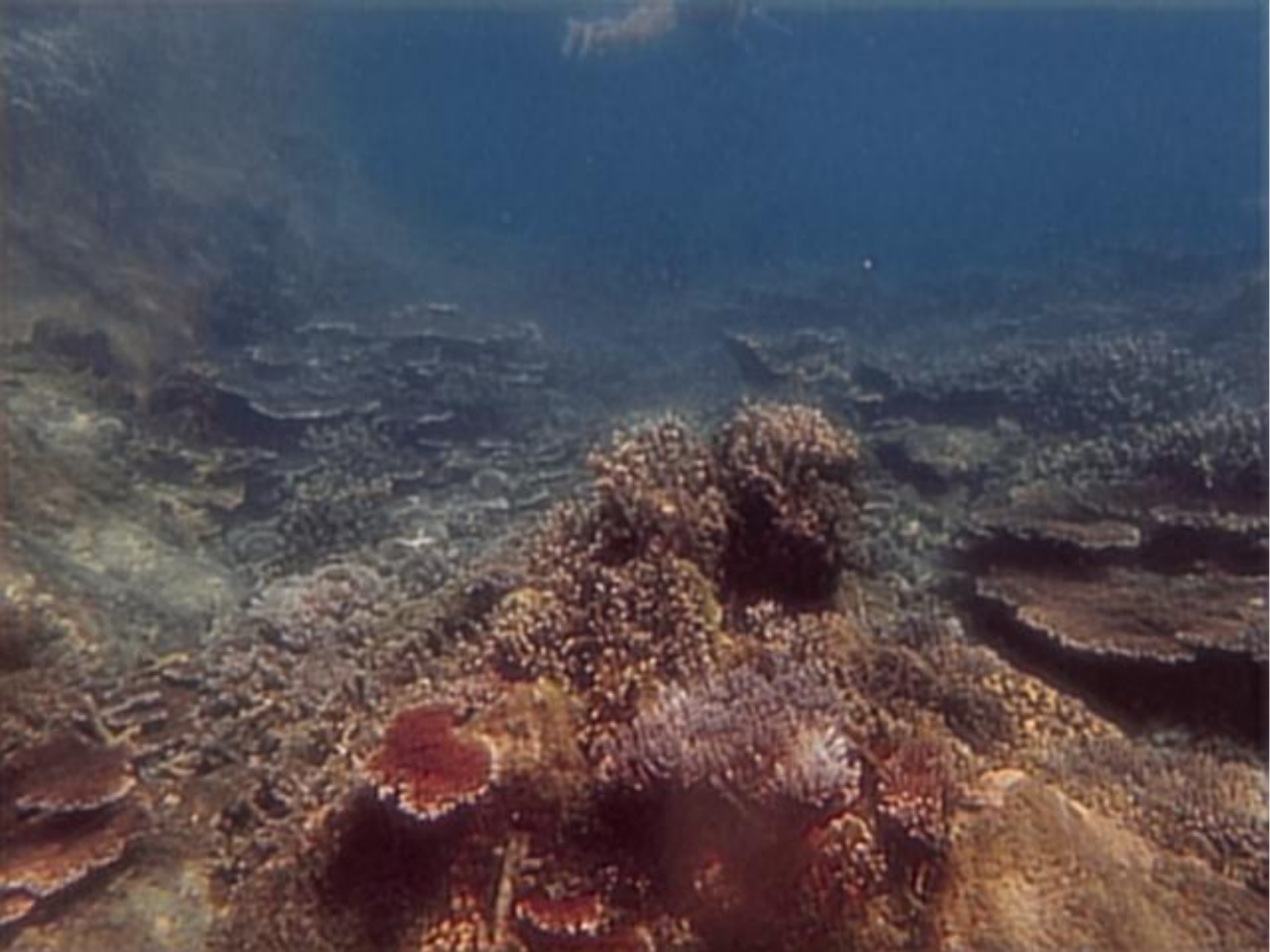}\\
				(a) input & (b) Ucolor &(c) w/o HSV \\
				\includegraphics[height=2.2cm,width=2.8cm]{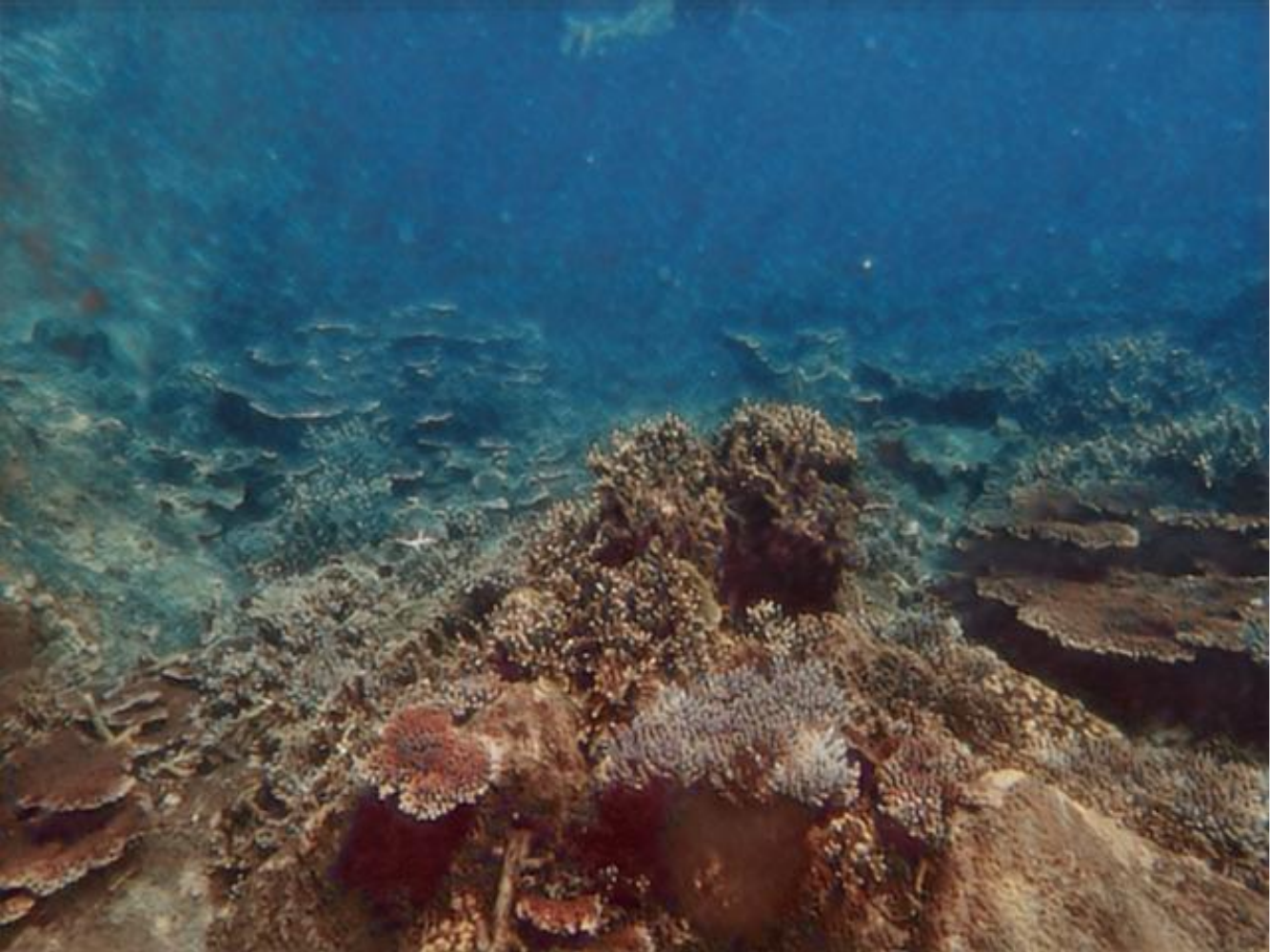}&
				\includegraphics[height=2.2cm,width=2.8cm]{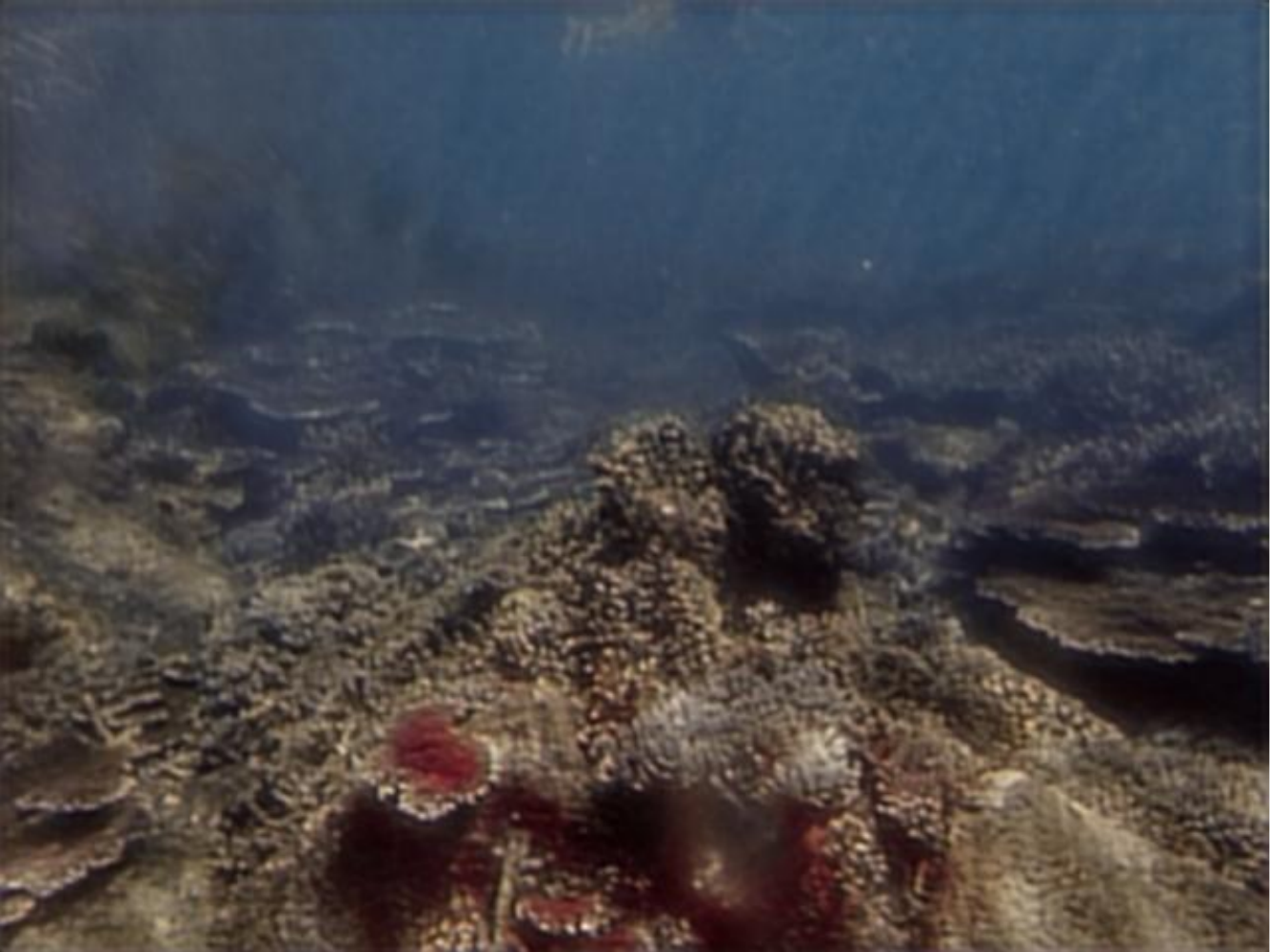}&
				\includegraphics[height=2.2cm,width=2.8cm]{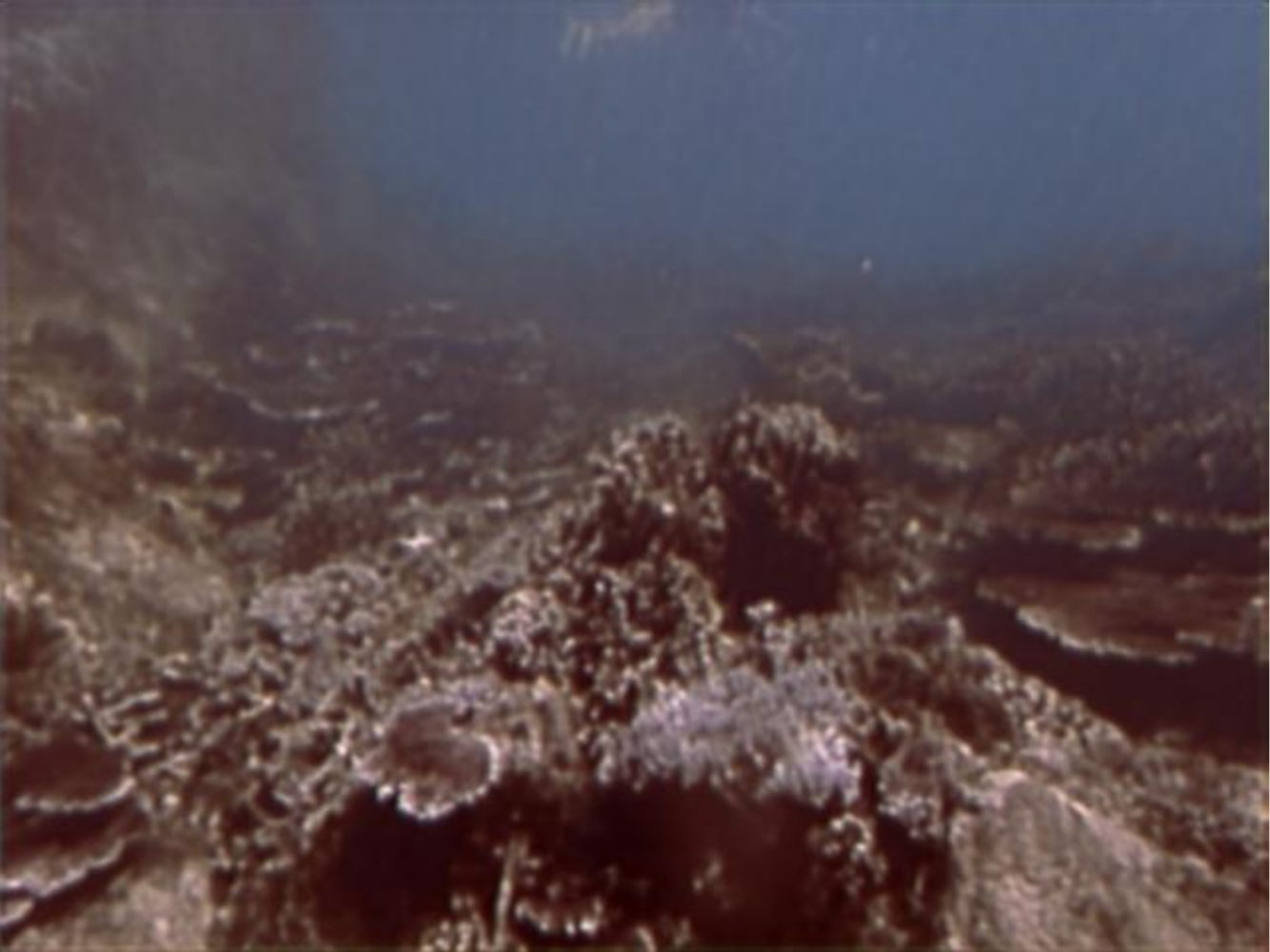}\\
				(d) w/o Lab & (e) w/o HSV+Lab & (f) w/ 3-RGB \\
			\end{tabular}
		\end{center}
		\caption{Ablation study of the contributions of each color space encoder path. Ucolor (three color spaces encoder) can achieve vivid color, high contrast, and clear details.}
		\label{fig:mcse}
	\end{figure}

\begin{figure}
		\begin{center}
			\begin{tabular}{c@{ }c@{ }c@{ }c}
				\includegraphics[height=2.2cm,width=2.8cm]{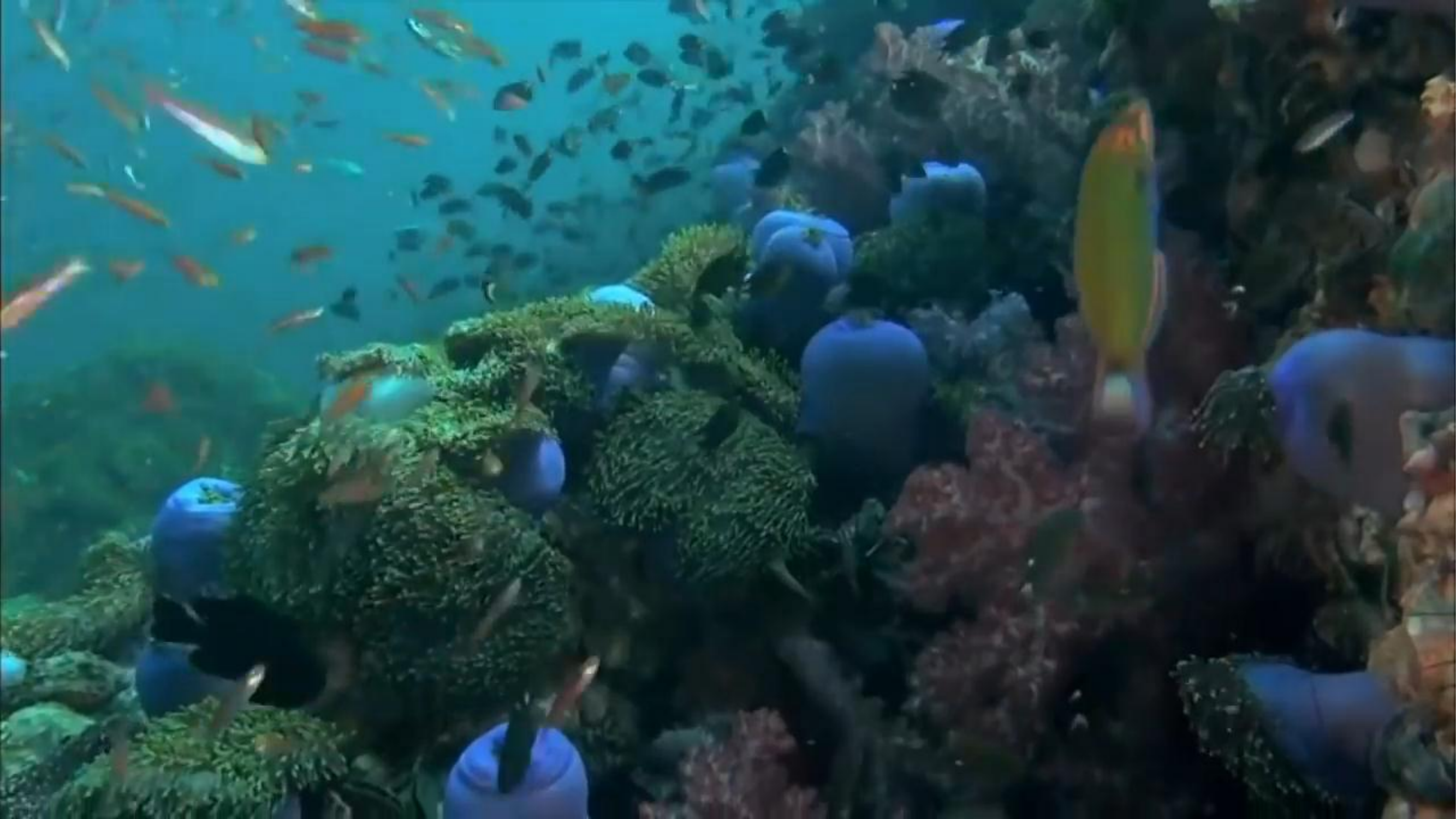} &
				\includegraphics[height=2.2cm,width=2.8cm]{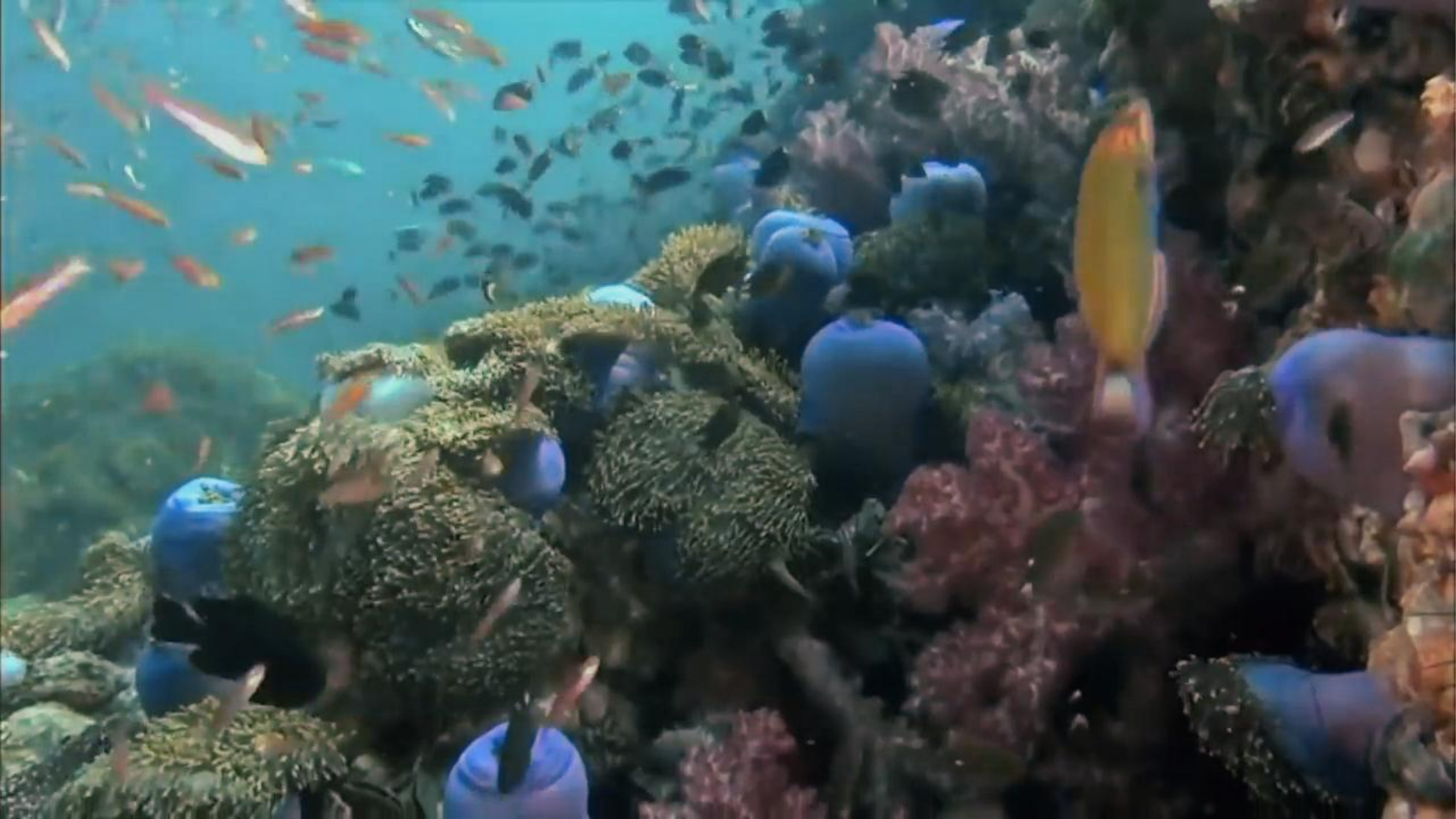} &
				\includegraphics[height=2.2cm,width=2.8cm]{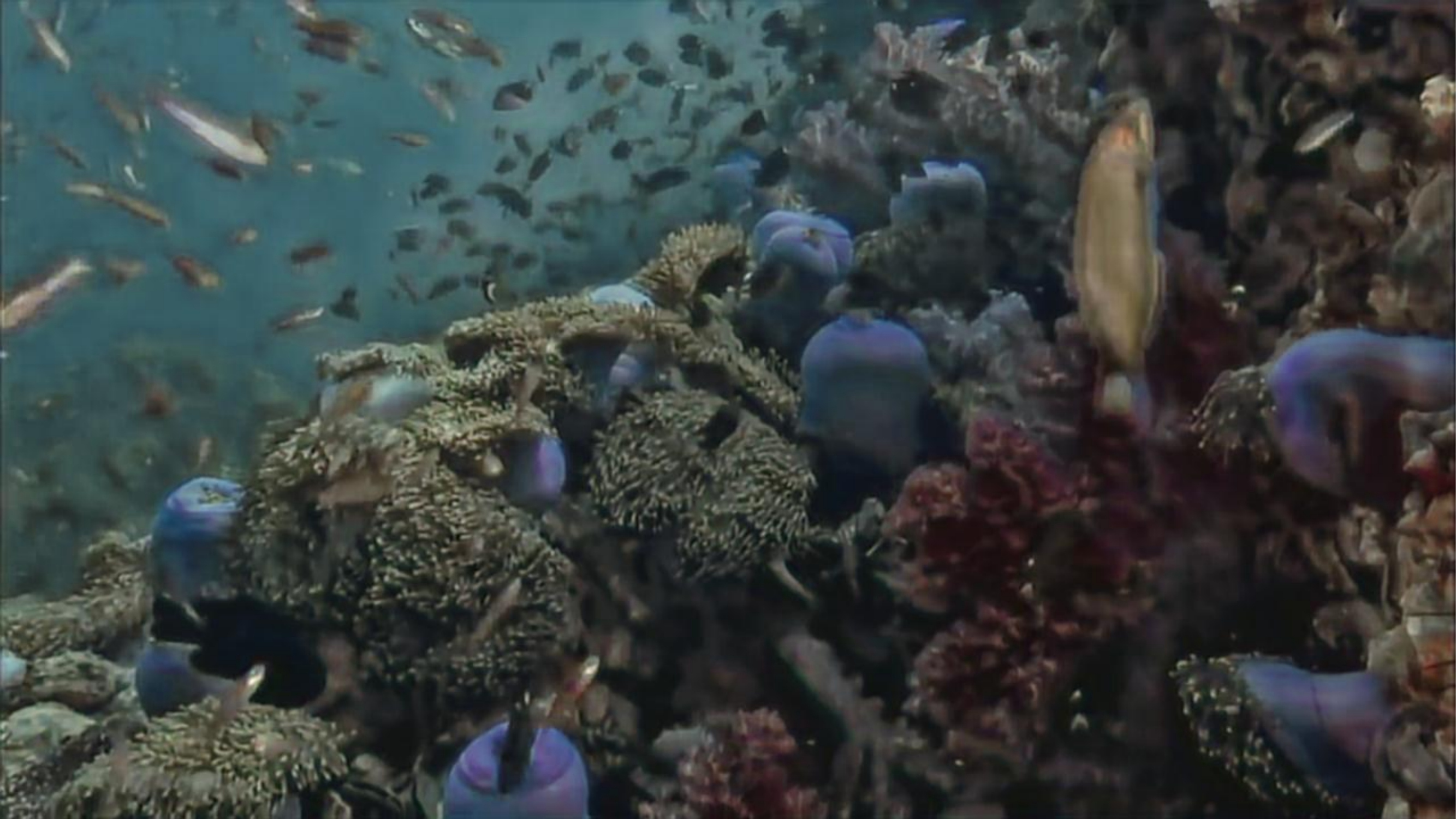}\\
				(a) input & (b) Ucolor &(c) w/o CAM \\
			\end{tabular}
		\end{center}
		\caption{Ablation study of the effectiveness of the channel-attention module for highlighting the multi-color space features. More realistic result is achieved by Ucolor (with channle-attention module) than the ablated model w/o CAM.}
		\label{fig:cam}
	\end{figure}

\begin{figure}[!t]
		\begin{center}
			\begin{tabular}{c@{ }c@{ }c@{ }c}
				\includegraphics[height=2.2cm,width=2.8cm]{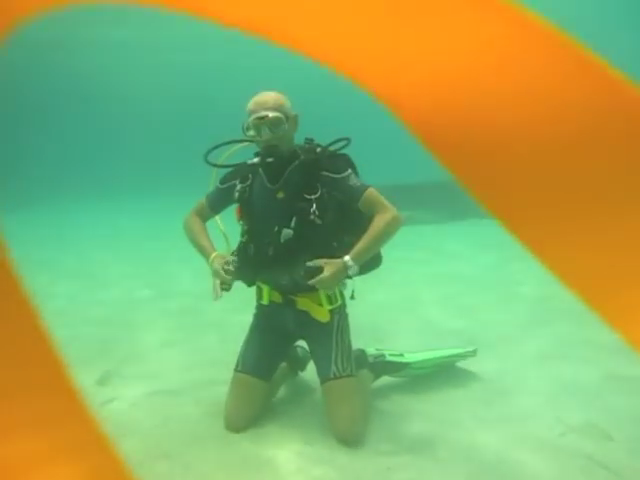} &
				\includegraphics[height=2.2cm,width=2.8cm]{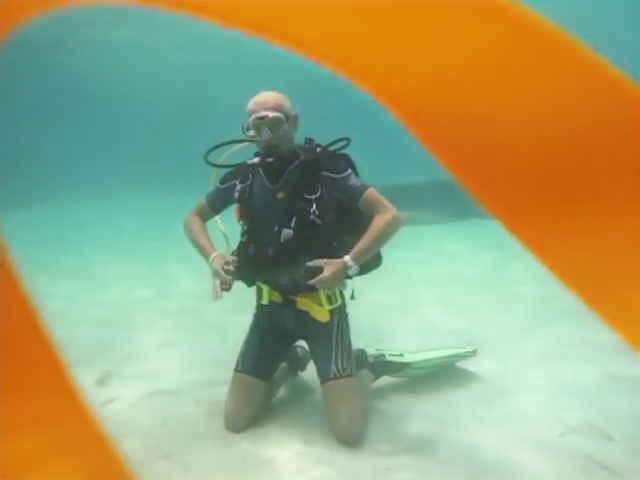} &
				\includegraphics[height=2.2cm,width=2.8cm]{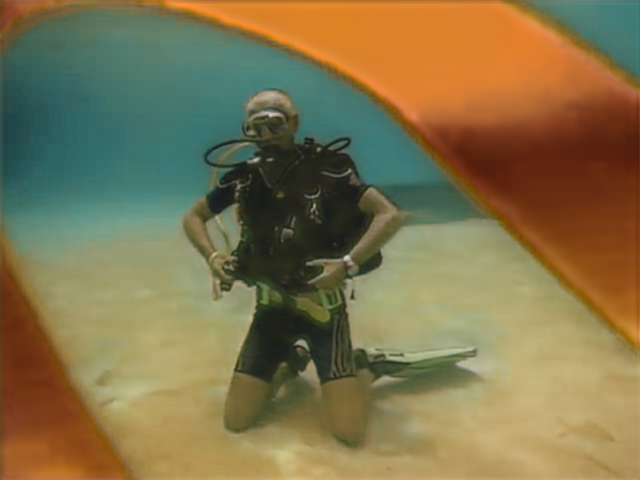}\\
				(a) input & (b) Ucolor &(c) w/o perc loss \\
			\end{tabular}
		\end{center}
		\caption{Ablation study towards the perceptual loss. By adding a perceptual loss to the $\ell_{1}$ loss, the visual quality of final result is improved.}
		\label{fig:loss}
	\end{figure}

	The quantitative PSNR (dB) and MSE ($\times10^{3}$) values on Test-S1000 and Test-R90  are presented in Table \ref{tab4}. The visual comparisons of the effects of reverse medium transmission maps, the contributions of each color space encoder path, the effectiveness of channle-attention module, and the effect of perceptual loss are shown in Figs.~\ref{fig:mtgm},~\ref{fig:mcse}, \ref{fig:cam}, and \ref{fig:loss}, respectively. 
	The conclusions drawn from the ablation studies are listed as follows. 
	
	1) As presented in Table \ref{tab4}, our full model achieves the best quantitative performance across two testing datasets when compared with the ablated models, which implies the effectiveness of the combinations of MCSE, MTGM, and CAM modules.
	
	2) Our RMT map can relatively accurately assign the high-quality degradation regions a higher weight than the RDCP map and the RUDCP map, thus achieving better visual quality, especially for high scattering regions as shown in Fig.~\ref{fig:mtgm}. Additionally, the quantitative performance of the Ucolor is better than the ablated models w/ RDCP, w/ RUDCP, and w/o MTGM, which suggests the importance of an accurate medium transmission map.
	
	3) The ablated models w/o HSV+Lab and w/ 3-RGB produce comparable performance as shown in Table \ref{tab4} and Fig.~\ref{fig:mcse}.  The results indicate that aimlessly adding more parameters or the same color encoder will not bring extra representational power to better enhance underwater images. In contrast, a well-design multi-color space embedding helps to learn more powerful representations to improve enhancement performance. In addition, removing any one of the three encoder paths (w/o HSV and w/o Lab) will decrease the performance as shown in Table \ref{tab4}.
	
	4) The ablated model w/o CAM produces an under-saturated result as shown in Fig.~\ref{fig:cam}. This may be induced by removing the CAM module that integrates and highlights the features extracted from multiple color spaces.

	5) The quantitative results in Table \ref{tab4} show that only using the $\ell_{1}$ loss can slightly improve the quantitative performance on Test-S1000 in terms of PNSR and MSE values. However, from the visual results in Fig.~\ref{fig:loss}, it can be observed that the enhanced image by Ucolor trained with the full loss function (\ie, the combination between the $\ell_{1}$ loss and the perceptual loss) is better than that by Ucolor trained without the perceptual loss. Thus, it is necessary to add the perceptual loss for improving the visual quality of final results. Note that only using the perceptual loss for training does not make sense, and thus we did not conduct such an ablation study.	
	
	\begin{table}
		\caption{Quantitative results of the ablation study in terms of average PSNR (dB) and MSE ($\times10^{3}$) values.}
		\begin{center}
			\begin{tabular}{c|c|c|c|c|c}
				\hline
				\multirow{2}{*}{Modules} & \multirow{2}{*}{Baselines} & \multicolumn{2}{c|}{\textbf{Test-S1000}} & \multicolumn{2}{c}{\textbf{Test-R90}} \\
				\cline{3-6}
				&  & PSNR$\uparrow$ & MSE $\downarrow$ &  PSNR$\uparrow$ & MSE $\downarrow$\\
				\hline\hline
				& \textbf{full model} &23.05 &0.50 &20.63  &0.77  \\
				\hline
				\multirow{5}{*}{MCSE}&w/o HSV &16.52  &1.83 &16.08  &1.97  \\
				& w/o Lab &18.33 &1.37 &17.54  &1.50  \\
				& w/o HSV+Lab &16.62 &1.88 &15.91  &2.10  \\
				& w/ 3-RGB &16.59 &2.06 &15.84  &2.11  \\
				\hline
				\multirow{3}{*}{MTGM} &w/o MTGM &17.02 &1.91 &17.37  &1.59  \\
				& w/ RDCP &18.74&0.87 &18.09  &1.01  \\
				& w/ RUDCP&18.94 &0.83 &17.56  &1.14  \\                                     
				\hline
				\multirow{1}{*}{CAM}&w/o CAM &16.36  &1.88 &16.02 &2.02   \\
				\hline
				\multirow{1}{*}{loss function}& w/o  perc loss  &23.11  &0.49 &18.29 & 0.98  \\
				\hline
			\end{tabular}
		\end{center}
		\label{tab4}
	\end{table}

	\begin{figure}[!t]
		\begin{center}
			\begin{tabular}{c@{ }c@{ }c@{ }c}
				\includegraphics[height=2.2cm,width=2.8cm]{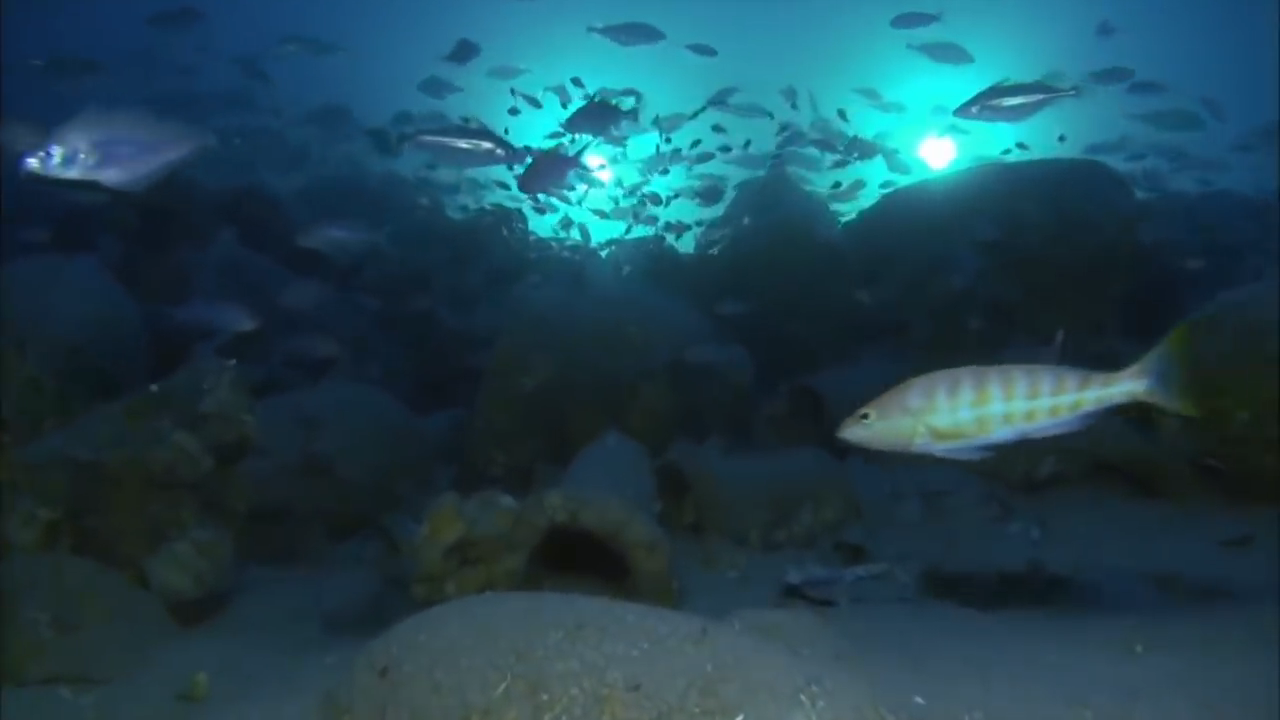} &
				\includegraphics[height=2.2cm,width=2.8cm]{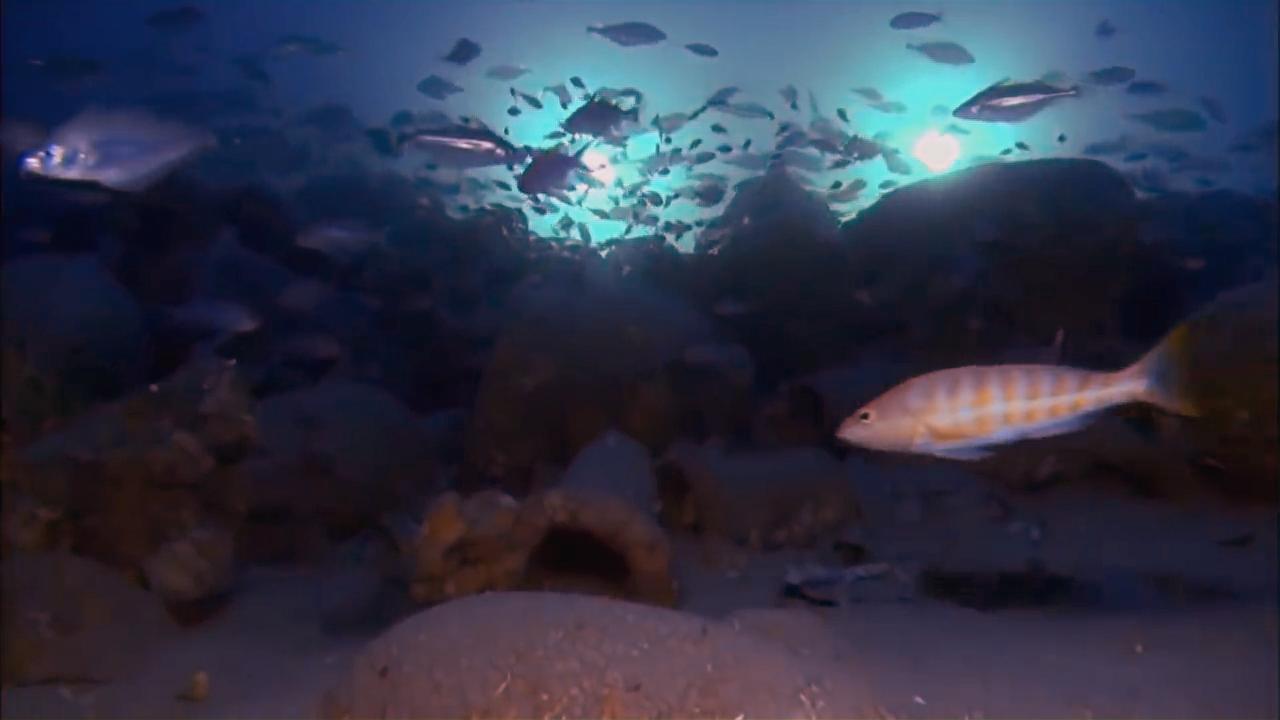} &
				\includegraphics[height=2.2cm,width=2.8cm]{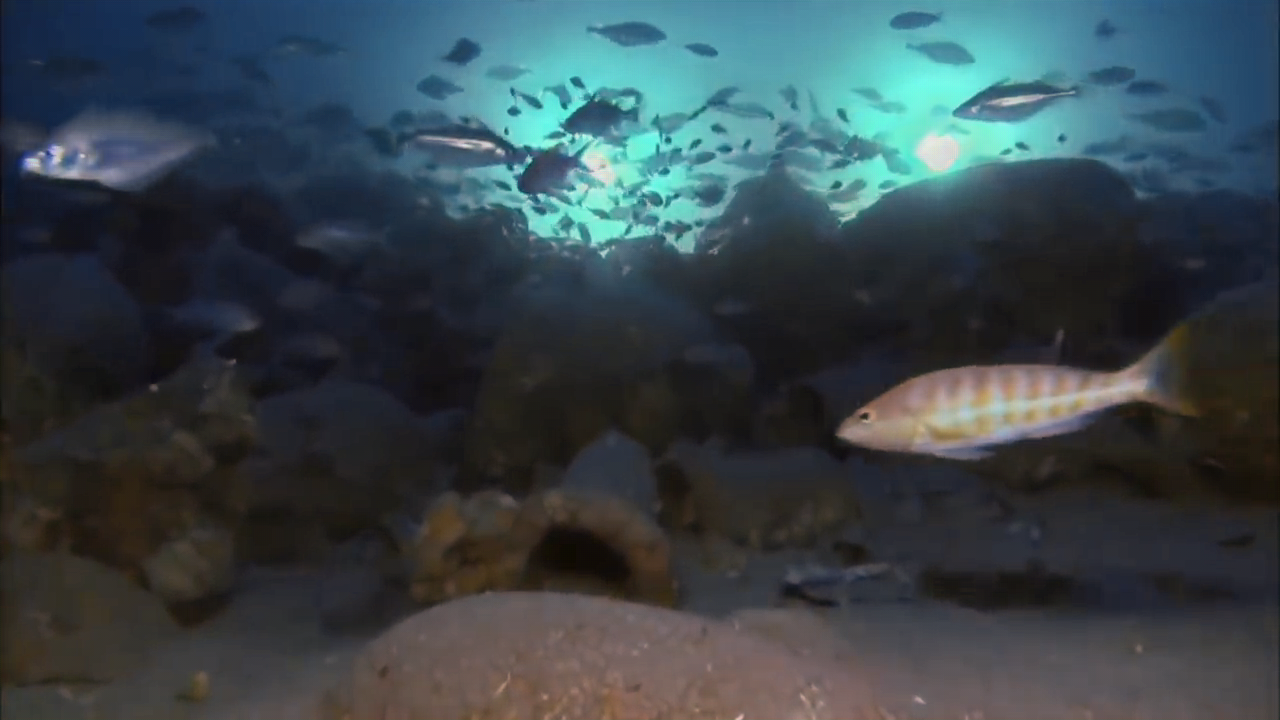}&\\
				(a) input & (b) Water-Net~\cite{Libenchmark} & (c)  Ucolor \\
			\end{tabular}
		\end{center}
		\caption{Failure case. The input underwater image has limited lighting. Although our Ucolor cannot effectively enhance this image, it does not introduce color casts like Water-Net.}
		\label{fig:failure}
	\end{figure}
	
	\subsection{Failure Case}
	When facing underwater images with limited lighting, our Ucolor, as well as other state-of-the-arts might not work well. Fig.~\ref{fig:failure} presents an example where both our Ucolor and latest deep learning-based Water-Net \cite{Libenchmark} fail to produce a visually compelling result when processing an underwater image with limited lighting. The potential reason lies in few such images in the training data sets. Thus, it is difficult for supervised learning-based networks such as Water-Net and Ucolor to handle such underwater images. The stronger ability and more diverse training data that handle such kinds of underwater images will be our future goal.

	\section{Conclusion}
	\label{Conclusion}
	We have presented a deep underwater image enhancement model. The proposed model learns the feature representations from diverse color spaces and highlights the most discriminative features by the channel-attention module. Besides, the domain knowledge is incorporated into the network by employing the reverse medium transmission map as the attention weights. Extensive experiments on diverse benchmarks have demonstrated the superiority of our solution and the effectiveness of multi-color space embedding and the reverse medium transmission guided decoder network structure. The effectiveness of the key components of our method has been verified in the ablation studies. 

	\section*{Acknowledgment}
	We thank Codruta O. Ancuti, Cosmin Ancuti, Christophe De Vleeschouwer, and Philippe Bekaert for providing the underwater Color Checker images \cite{Ancuti2018}.  We thank Yecai Guo, Hanyu Li, and Peixiain Zhuang for providing their results \cite{Guo2019}.
	\par
	
	\ifCLASSOPTIONcaptionsoff
	\newpage
	\fi
	
	{
		\bibliographystyle{IEEEtran}
		\bibliography{ref}
	}
	
\begin{IEEEbiography}[{\includegraphics[width=1in,height=1.25in,clip,keepaspectratio]{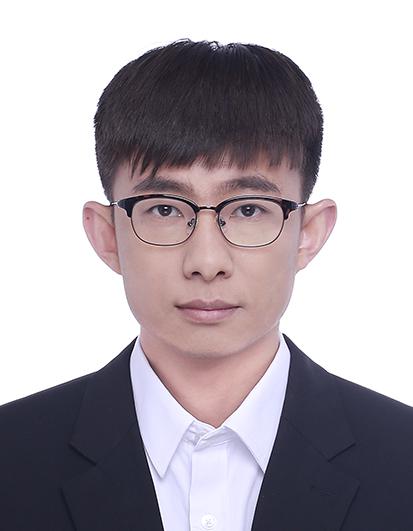}}]{Chongyi Li}  received the Ph.D. degree from the School of Electrical and Information Engineering, Tianjin University, Tianjin, China, in June 2018. From 2016 to 2017, he was a joint-training Ph.D. Student with Australian National University, Australia. He was a postdoctoral fellow with the Department of Computer Science, City University of Hong Kong, Hong Kong. He is currently a research fellow with the School of Computer Science and Engineering, Nanyang Technological University (NTU), Singapore. His current research focuses on image processing, computer vision, and deep learning, particularly in the domains of image restoration and enhancement.
\end{IEEEbiography}
\begin{IEEEbiography}[{\includegraphics[width=1in,height=1.25in,clip,keepaspectratio]{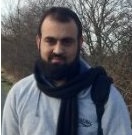}}]{Saeed Anwar} is a research Scientist at Data61, Commonwealth Scientific and Industrial Research Organization (CSIRO), Australia in Cyber-Physical Systems, adjunct Lecturer at Australian National University (ANU) and visiting fellow at University of Technology Sydney (UTS). He completed his PhD at the Computer Vision Research Group (CVRG) at Data61, CSIRO and College of Electrical \& Computer Science (CECS), Australian National University (ANU).
\end{IEEEbiography}
\begin{IEEEbiography}[{\includegraphics[width=1in,height=1.25in,clip,keepaspectratio]{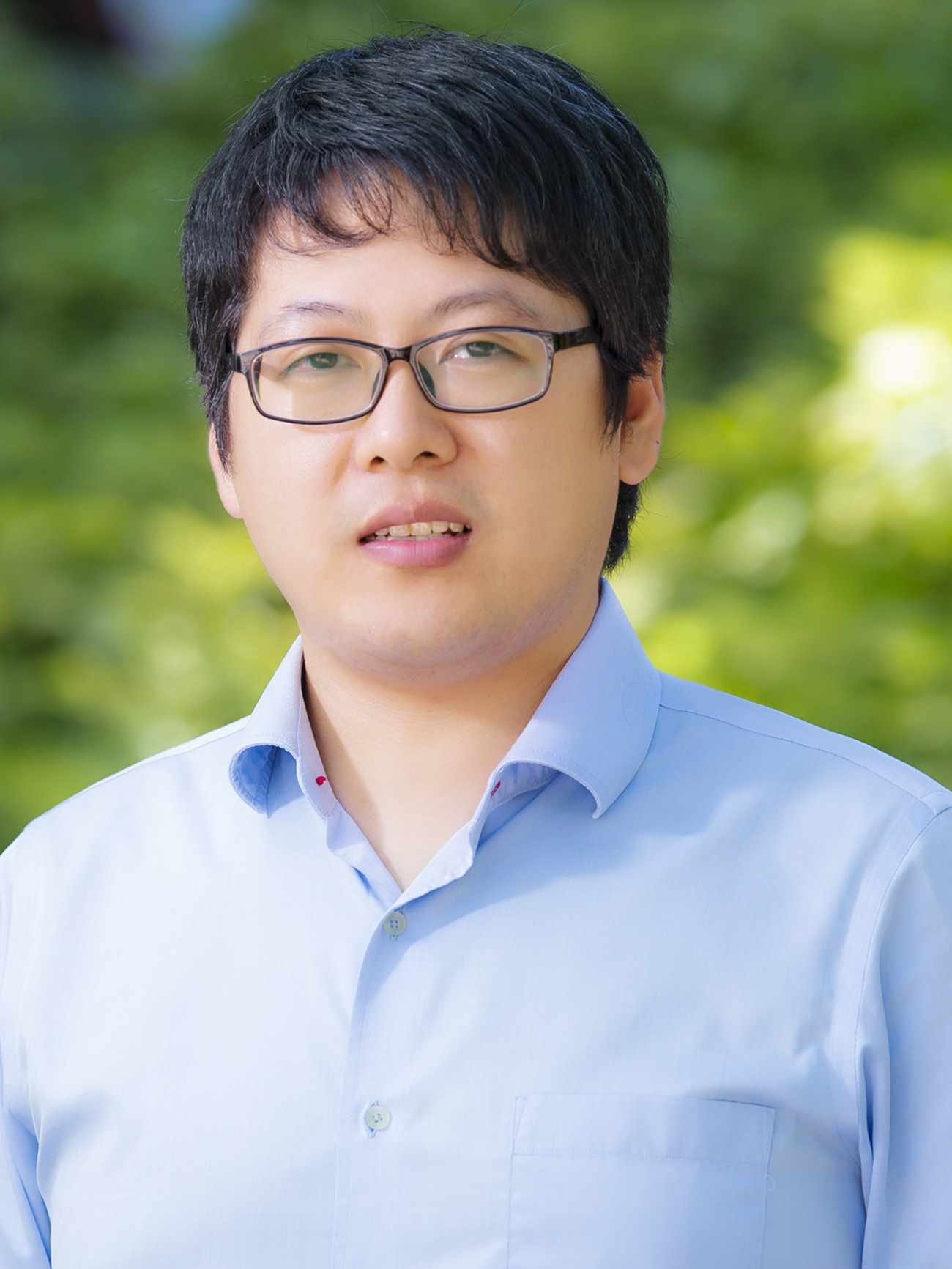}}]{Junhui Hou} (S'13-M'16-SM'20) received the B.Eng. degree in information engineering (Talented Students Program) from the South China University of Technology, Guangzhou, China, in 2009, the M.Eng. degree in signal and information processing from Northwestern Polytechnical University, Xi'an, China, in 2012, and the Ph.D. degree in electrical and electronic engineering from the School of Electrical and Electronic Engineering, Nanyang Technological University, Singapore, in 2016. 
He has been an Assistant Professor with the Department of Computer Science, City University of Hong Kong, since 2017. His research interests fall into the general areas of visual computing, such as image/video/3D geometry data representation, processing and analysis, semi/un-supervised data modeling, and data compression and adaptive transmission
Dr. Hou was the recipient of several prestigious awards, including the Chinese Government Award for Outstanding Students Study Abroad from China Scholarship Council in 2015, and the Early Career Award (3/381) from the Hong Kong Research Grants Council in 2018. He is a member of Multimedia Systems \& Applications Technical Committee (MSA-TC), IEEE CAS. He is currently serving as an Associate Editor for IEEE Transactions on Circuits and Systems for Video Technology, The Visual Computer, and Signal Processing: Image Communication, and the Guest Editor for the IEEE Journal of Selected Topics in Applied Earth Observations and Remote Sensing. He also served as an Area Chair of ACM MM 2019 and 2020, IEEE ICME 2020,
and WACV 2021. He is a senior member of IEEE.
\end{IEEEbiography}
\begin{IEEEbiography}[{\includegraphics[width=1in,height=1.25in,clip,keepaspectratio]{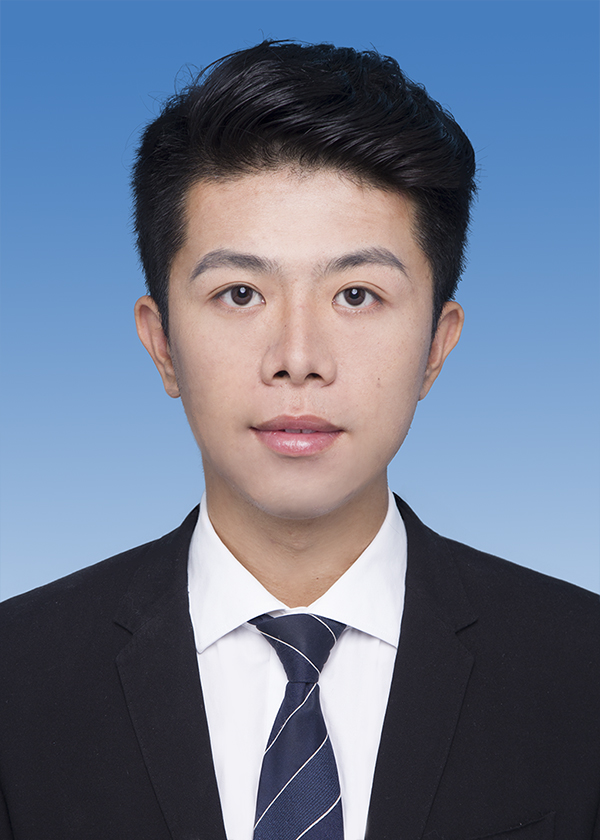}}]{Runmin Cong} Runmin Cong (M’19) received the Ph.D. degree in information and communication engineering from Tianjin University, Tianjin, China, in June 2019. He is currently an Associate Professor with the Institute of Information Science, Beijing Jiaotong University, Beijing, China. He was a visiting student/staff at Nanyang Technological University (NTU), Singapore, and City University of Hong Kong (CityU), Hong Kong. His research interests include computer vision, multimedia processing and understanding, visual attention perception and saliency computation, remote sensing image interpretation and analysis, and visual content enhancement in an open environment, etc.
\end{IEEEbiography}
\begin{IEEEbiography}[{\includegraphics[width=1in,height=1.25in,clip,keepaspectratio]{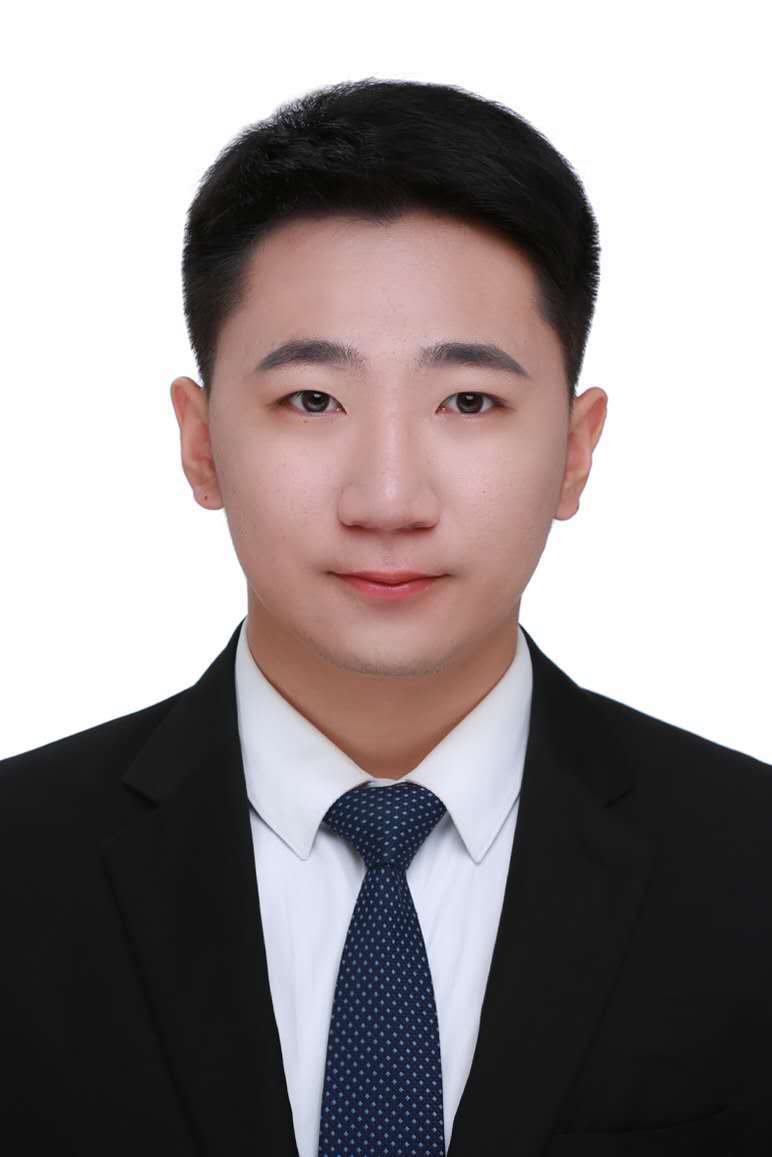}}]{Chunle Guo}  received his PhD degree from Tianjin University in China under the supervision of Prof. Jichang Guo. He conducted the Ph.D. research as a Visiting Student with the School of Electronic Engineering and Computer Science, Queen Mary University of London (QMUL), UK.  He continued his research as a Research Associate with Department of Computer Science, City University of Hong Kong (CityU), from 2018 to 2019. Now he is a postdoc research fellow working with Prof. Ming-Ming Cheng in Nankai University. His research interests lies in image processing, computer vision, and deep learning.
\end{IEEEbiography}
\begin{IEEEbiography}[{\includegraphics[width=1in,height=1.25in,clip,keepaspectratio]{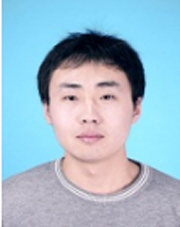}}]{Wenqi Ren}  is an Associate Professor in Institute of Information Engineering, Chinese Academy of Sciences, China. He received his Ph.D. degree from
Tianjin University, Tianjin, China, in 2017. During 2015 to 2016, he was supported by China Scholarship Council and working with Prof. Ming-Husan Yang as a joint-training Ph.D. student in the Electrical Engineering and Computer Science Department, at the University of California at Merced. He received Tencent Rhino Bird Elite Graduate Program Scholarship in 2017, MSRA Star Track Program in 2018. His research interests include image processing and related high-level vision problems.
\end{IEEEbiography}	
	
	%\flushbottom
	
\end{document}